\UseRawInputEncoding
\documentclass[12pt]{article}
\usepackage{authblk}
\usepackage[utf8]{inputenc}
\usepackage{graphicx}
\usepackage{multirow}
\usepackage{multicol}
\usepackage{mathrsfs}
\usepackage{subfigure}
\usepackage{geometry}
\usepackage{makecell}
\usepackage{color}
\usepackage{amsthm,amsmath,amssymb}
\usepackage{mathrsfs}
\usepackage{caption}
\usepackage{hyperref}
\usepackage{booktabs}
\usepackage{tikz}
\usetikzlibrary{calc}
\tikzset{
    ncbar angle/.initial=90,
    ncbar/.style={
        to path=(\tikztostart)
        -- ($(\tikztostart)!#1!\pgfkeysvalueof{/tikz/ncbar angle}:(\tikztotarget)$)
        -- ($(\tikztotarget)!($(\tikztostart)!#1!\pgfkeysvalueof{/tikz/ncbar angle}:(\tikztotarget)$)!\pgfkeysvalueof{/tikz/ncbar angle}:(\tikztostart)$)
        -- (\tikztotarget)
    },
    ncbar/.default=0.5cm,
}

\tikzset{square left brace/.style={ncbar=0.5cm}}
\tikzset{square right brace/.style={ncbar=-0.5cm}}

\tikzset{round left paren/.style={ncbar=0.5cm,out=120,in=-120}}
\tikzset{round right paren/.style={ncbar=0.5cm,out=60,in=-60}}
\hypersetup{
    colorlinks=true,
    linkcolor=cyan,
    filecolor=blue,
    urlcolor=red,
    citecolor=green
}

\newtheorem{theorem}{Theorem}[section]
\newcommand{\bbC}{\mathbb{C}}
\newcommand{\bbR}{\mathbb{R}}

\newcommand{\calB}{\mathcal{B}}
\newcommand{\calF}{\mathcal{F}}

\newcommand{\calL}{\mathcal{L}}
\newcommand{\calK}{\mathcal{K}}

\title{ButterflyNet2D: Bridging Classical Methods and Neural Network
Methods in Image Processing}
\author[1]{Gengzhi Yang}
\author[1]{Yingzhou Li}
\affil[1]{School of Mathematical Sciences, Fudan University}
\date{}

\begin{document}

\maketitle

\begin{abstract}
Both classical Fourier transform-based methods and neural network methods
are widely used in image processing tasks. The former has better
interpretability, whereas the latter often achieves better performance in
practice. This paper introduces ButterflyNet2D, a regular CNN with sparse
cross-channel connections. A Fourier initialization strategy for
ButterflyNet2D is proposed to approximate Fourier transforms. Numerical
experiments validate the accuracy of ButterflyNet2D approximating both the
Fourier and the inverse Fourier transforms. Moreover, through four image
processing tasks and image datasets, we show that training ButterflyNet2D
from Fourier initialization does achieve better performance than random
initialized neural networks.

\textbf{Keywords}: Convolutional neural network, ButterflyNet, Butterfly
Algorithm, image processing

\end{abstract}

\section{Introduction}

Image processing tasks appear widely in our daily life and are handled by
classical methods and/or neural network methods behind the screen. Image
processing tasks~\cite{bhatt2021cnn} include but are not limited to image
denoising, image deblurring, image inpainting, image recognition, image
classification, etc. A decade ago, image processing tasks were almost
always handled by classical methods, e.g., Fourier transform, wavelet
transform, partial differential equations, etc. With the rise of machine
learning and neural network, neural network based methods dominate image
processing. Among all neural network architecture, deep convolutional
neural network~(CNN) is the most popular architecture for image processing
tasks. Many variants of CNN, including LeNet~\cite{lenet},
AlexNet~\cite{alexnet}, U-Net~\cite{He_2017_ICCV, ren2019, unet}, etc.,
are proposed and successfully address image processing tasks. In this
paper, we bridge the classical method, Fourier transform, with the
convolutional neural network method for image processing via
ButterflyNet2D.

\paragraph{Fourier transform.} Fourier transform is a linear transformation
that decomposes functions into frequency components. It has a wide range
of applications in a variety of fields, including signal
processing~\cite{fractionalFT_SIGNAL,FTsignal_processing}, 
image processing~\cite{ghani2020review, uzun2005fpga,
xu2013unnatural}, etc. Besides the time-frequency
transformation, the existence of the well-known fast algorithm, fast
Fourier transform (FFT)~\cite{cooley1967historical, cooley1969fast}, makes
the transform widely adopted in practice. An FFT decomposes a dense
discrete Fourier transform matrix of size $N \times N$ into a product of
$O(\log N)$ sparse matrices, each of which is sparsity $O(N)$. Another
family of fast algorithms that could be applied to accelerate the Fourier
transform is the butterfly algorithm~\cite{candes2009fast,
li2017interpolative, li2015, li2015multiscale, li2018, ying2009sparse}.
Although butterfly algorithms were originally designed to accelerate the
computation of the Fourier integral operator, they could be applied to
approximate the discrete Fourier transform in $O(N \log N)$ operations as
well. The same scaling holds for both FFT and butterfly algorithms, while
the latter suffers from an approximation error. Hence, for Fourier
transform in classical image processing methods, FFT is applied.

\paragraph{Neural Network.} There has been a growing trend to ask for
better and faster image processing methods in the last two decades.
CNN~\cite{lenet,cnnzipcode} was initially introduced to process images
directly and has later been embedded into other deep neural network
architectures~\cite{cnnTrans} to improve image processing further. The
success of CNN and its variants in image processing have been demonstrated
in tons of works~\cite{zhang2017learning, cnnsucess}. However, unlike
Fourier transform in image processing, which has much mathematical
understanding of the methods, CNN lacks interpretability. In most cases,
researchers refer to the universal approximation theorem of
CNN~\cite{DingxuanZhou, universalapproximation, li2018butterfly} for its
great success in image processing. Recently, the Fourier transform has
been imported to be part of the neural network architecture and leads to
Fourier CNN~\cite{fcnn}, Fourier neural network~\cite{fourierOperator},
etc. In addition, the FFT structure was incorporated into neural networks
and applied to image processing tasks~\cite{Pbutterfly,learningBF}. The
connection between the Fourier transform and CNN was established via the
butterfly algorithm, and ButterflyNet~\cite{li2018butterfly,
xu2020butterfly} was proposed to address signal processing tasks and
one-dimensional PDEs.

\paragraph{Contribution.} In this work, based on the two-dimensional
butterfly algorithm, we introduce a sparsified CNN architecture named
ButterFlyNet2D. This neural network with a particular initialization can
approximate a two-dimensional discrete Fourier transform. The
approximation power is theoretically guaranteed. In summary, our
contribution can be organized as follows.
\begin{itemize}
    \item The ButterflyNet2D network is constructed, which is a CNN
    architecture with sparse channel connections;
    \item Fourier initialization is proposed for ButterflyNet2D
    approximating a two-dimensional discrete Fourier transform with
    guaranteed approximation error;
    \item ButterflyNet2D is applied to many ill-posed image processing
    tasks, i.e., denoising, deblurring, inpainting, and watermark removal,
    on practical image data sets.
\end{itemize}
Numerical experiments demonstrate that ButterflyNet2D, as a specialized
CNN, along with the Fourier initialization, outperforms its randomly
initialized counterpart and another well-known Neumann
network~\cite{gilton2019neumann}. The latter was designed for inverse
problems in image processing.

\paragraph{Organization.} The rest paper is organized as follows.
Section~\ref{sec:net_init} proposes the ButterflyNet2D architecture and
Fourier initialization strategy. In Section~\ref{sec:num_res},
ButterflyNet2Ds with Fourier initialization and random initialization are
applied to the image denoising, image deblurring, image inpainting, and
watermark removal tasks. The comparison against the Neumann network is
also presented. Finally, Section~\ref{sec:conclusion} concludes the paper.

\section{ButterflyNet2D and Fourier Initialization}
\label{sec:net_init}

This section constructs ButterflyNet2D and initializes it as an
approximated two-dimensional discrete Fourier transform. We first pave the
path to review the two-dimensional butterfly algorithm in
Section~\ref{sec:preliminary}. In Section~\ref{sec:butterflyalgo}, the
two-dimensional butterfly algorithm is detailed, with the kernel function
being Fourier transform. Section~\ref{sec:butterflynet} introduces the
architecture of ButterflyNet2D, and its Fourier initialization is proposed
in Section~\ref{sec:fourierinit}.
    
\subsection{Preliminary}
\label{sec:preliminary}

\paragraph{Chebyshev Interpolation.}
An important numerical tool in approximating the Fourier transform is the
Lagrange polynomial on Chebyshev points, which is known as the Chebyshev
interpolation. The Chebyshev points of order $r$ on $[-1/2, 1/2]$ are
defined as
\begin{equation*}
    \left\{ z_i = \frac{1}{2} \cos \frac{(2i-1)\pi}{2r} \right\}_{i
    \in \{1, \dots, r\}}.
\end{equation*}
The associated Lagrange polynomial at $z_k$ admits
\begin{equation*}
    \calL_k(x) = \prod_{p\neq k}\frac{x-z_p}{z_k-z_p}.
\end{equation*}
In two dimension, $r^2$ Chebyshev points in $[-1/2, 1/2] \times [-1/2,
1/2]$ are the tensor grid of one-dimensional Chebyshev points,
\begin{equation*}
    \left\{ z_{i,j} = \left( \frac{1}{2} \cos\frac{(2i-1)\pi}{2r},
    \frac{1}{2}\cos\frac{(2j-1)\pi}{2r} \right) \right\}_{i,j \in
    \{1, \dots, r\}},
\end{equation*}
The two-dimensional Chebyshev interpolation then admits,
\begin{equation} \label{eq:chebinterp}
    \calL_{(i,j)} = \calL_{i} \calL_{j}.
\end{equation}

\paragraph{Domain Decomposition.}

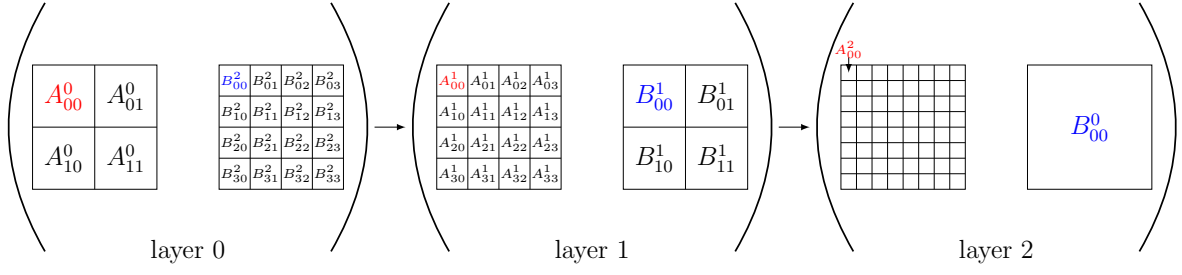
\begin{figure}[htb]
    \centering
    \resizebox{\textwidth}{!}{
    \begin{tikzpicture}
    \draw [black, thick] (0.2,-1) to [round left paren ] (0.2,3);
    \draw [black, thick] (4.8,-1) to [round right paren ] (4.8,3);
    \node[font=\fontsize{12}{12}\selectfont, color=black]  at(2.5,-1) {layer 0};
    
    \draw [black, thick] (6.7,-1) to [round left paren ] (6.7,3);
    \draw [black, thick] (11.3,-1) to [round right paren ] (11.3,3);
    \node[font=\fontsize{12}{12}\selectfont, color=black]  at(9,-1) {layer 1};
    
    \draw [black, thick] (13.2,-1) to [round left paren ] (13.2,3);
    \draw [black, thick] (17.8,-1) to [round right paren ] (17.8,3);
    \node[font=\fontsize{12}{12}\selectfont, color=black]  at(15.5,-1) {layer 2};
    
    \draw (0,0) rectangle (2,2);
    \draw (3,0) rectangle (5,2);
    
    \draw[-latex] (5.5,1) -- (6,1);
    \draw (6.5,0) rectangle (8.5,2);
    \draw (9.5,0) rectangle (11.5,2);
    
    \draw[-latex] (12,1) -- (12.5,1);
    \draw (13,0) rectangle (15,2);
    \draw (16,0) rectangle (18,2);
    
    \draw (0,1) -- (2,1);
    \draw (1,0) -- (1,2);
    
    \draw (3,1) -- (5,1);
    \draw (4,0) -- (4,2);
    \draw (3,0.5) -- (5,0.5);
    \draw (3.5,0) -- (3.5,2);
    \draw (3,1.5) -- (5,1.5);
    \draw (4.5,0) -- (4.5,2);
    
    \draw (6.5,1) -- (8.5,1);
    \draw (7.5,0) -- (7.5,2);
    \draw (6.5,0.5) -- (8.5,0.5);
    \draw (7,0) -- (7,2);
    \draw (6.5,1.5) -- (8.5,1.5);
    \draw (8,0) -- (8,2);
    
    \draw (13,1) -- (15,1);
    \draw (14,0) -- (14,2);
    \draw (13,0.5) -- (15,0.5);
    \draw (13.5,0) -- (13.5,2);
    \draw (13,1.5) -- (15,1.5);
    \draw (14.5,0) -- (14.5,2);
    \draw (13,0.25) -- (15,0.25);
    \draw (13.75,0) -- (13.75,2);
    \draw (13,0.75) -- (15,0.75);
    \draw (13.25,0) -- (13.25,2);
    \draw (14.25,0) -- (14.25,2);
    \draw (13,1.25) -- (15,1.25);
    \draw (13,1.75) -- (15,1.75);
    \draw (14.75,0) -- (14.75,2);
    
    \draw (9.5,1) -- (11.5,1);
    \draw (10.5,0) -- (10.5,2);
    
    \node[color=red]  at(0.5,1.5) {$A_{00}^0$};
    \node  at(1.5,1.5) {$A_{01}^0$};
    \node  at(0.5,0.5) {$A_{10}^0$};
    \node  at(1.5,0.5) {$A_{11}^0$};
    
    \node[font=\fontsize{6}{6}\selectfont, color=blue]  at(3.25,1.75) {$B_{00}^2$};
    \node[font=\fontsize{6}{6}\selectfont]  at(3.75,1.75) {$B_{01}^2$};
    \node[font=\fontsize{6}{6}\selectfont]  at(4.25,1.75) {$B_{02}^2$};
    \node[font=\fontsize{6}{6}\selectfont]  at(4.75,1.75) {$B_{03}^2$};
    \node[font=\fontsize{6}{6}\selectfont]  at(3.25,1.25) {$B_{10}^2$};
    \node[font=\fontsize{6}{6}\selectfont]  at(3.75,1.25) {$B_{11}^2$};
    \node[font=\fontsize{6}{6}\selectfont]  at(4.25,1.25) {$B_{12}^2$};
    \node[font=\fontsize{6}{6}\selectfont]  at(4.75,1.25) {$B_{13}^2$};
    \node[font=\fontsize{6}{6}\selectfont]  at(3.25,0.75) {$B_{20}^2$};
    \node[font=\fontsize{6}{6}\selectfont]  at(3.75,0.75) {$B_{21}^2$};
    \node[font=\fontsize{6}{6}\selectfont]  at(4.25,0.75) {$B_{22}^2$};
    \node[font=\fontsize{6}{6}\selectfont]  at(4.75,0.75) {$B_{23}^2$};
    \node[font=\fontsize{6}{6}\selectfont]  at(3.25,0.25) {$B_{30}^2$};
    \node[font=\fontsize{6}{6}\selectfont]  at(3.75,0.25) {$B_{31}^2$};
    \node[font=\fontsize{6}{6}\selectfont]  at(4.25,0.25) {$B_{32}^2$};
    \node[font=\fontsize{6}{6}\selectfont]  at(4.75,0.25) {$B_{33}^2$};
    
    \node[font=\fontsize{6}{6}\selectfont, color=red]  at(6.75,1.75) {$A_{00}^1$};
    \node[font=\fontsize{6}{6}\selectfont]  at(7.25,1.75) {$A_{01}^1$};
    \node[font=\fontsize{6}{6}\selectfont]  at(7.75,1.75) {$A_{02}^1$};
    \node[font=\fontsize{6}{6}\selectfont]  at(8.25,1.75) {$A_{03}^1$};
    \node[font=\fontsize{6}{6}\selectfont]  at(6.75,1.25) {$A_{10}^1$};
    \node[font=\fontsize{6}{6}\selectfont]  at(7.25,1.25) {$A_{11}^1$};
    \node[font=\fontsize{6}{6}\selectfont]  at(7.75,1.25) {$A_{12}^1$};
    \node[font=\fontsize{6}{6}\selectfont]  at(8.25,1.25) {$A_{13}^1$};
    \node[font=\fontsize{6}{6}\selectfont]  at(6.75,0.75) {$A_{20}^1$};
    \node[font=\fontsize{6}{6}\selectfont]  at(7.25,0.75) {$A_{21}^1$};
    \node[font=\fontsize{6}{6}\selectfont]  at(7.75,0.75) {$A_{22}^1$};
    \node[font=\fontsize{6}{6}\selectfont]  at(8.25,0.75) {$A_{23}^1$};
    \node[font=\fontsize{6}{6}\selectfont]  at(6.75,0.25) {$A_{30}^1$};
    \node[font=\fontsize{6}{6}\selectfont]  at(7.25,0.25) {$A_{31}^1$};
    \node[font=\fontsize{6}{6}\selectfont]  at(7.75,0.25) {$A_{32}^1$};
    \node[font=\fontsize{6}{6}\selectfont]  at(8.25,0.25) {$A_{33}^1$};
    
    \node[color=blue]  at(10,1.5) {$B_{00}^1$};
    \node  at(11,1.5) {$B_{01}^1$};
    \node  at(10,0.5) {$B_{10}^1$};
    \node  at(11,0.5) {$B_{11}^1$};
    
    \draw[-latex] (13.125,2.125) -- (13.125,1.875);
    \node[font=\fontsize{6}{6}\selectfont, color=red]  at(13.125,2.25) {$A_{00}^2$};
    
    \node[color=blue]  at(17,1) {$B_{00}^0$};
    \end{tikzpicture}
    }
    \caption{Recursive domain decomposition for $A, B$.}
    \label{fig:bfexample}
\end{figure}

The butterfly algorithm essentially relies on multiscale domain
decomposition. Here we introduce the domain decomposition for square
domain pairs. Given two domains $A = [0, K) \times [0, K)$ and $B = [0, 1)
\times [0, 1)$ for $K$ being the frequency range, we conduct 4-partition
recursively to both $A$ and $B$. The resulting decomposed domains are
denoted as $A_{i_x, i_y}^\ell$ and $B_{j_x, j_y}^{L-\ell}$, where $(i_x, i_y)$
and $(j_x,j_y)$ are indices, and $\ell$ is recursion layer index. More
precisely, the explicit expressions of $A_{i_x, i_y}^{\ell}$ and $B_{j_x,
j_y}^{L-\ell}$ are
\begin{equation*}
    \begin{split}
        A_{i_x, i_y}^{\ell} & = \Bigg[ \frac{i_x K}{2^{\ell+1}},
        \frac{(i_x+1) K}{2^{\ell+1}} \Bigg) \times
        \Bigg[\frac{i_y K}{2^{\ell+1}},
        \frac{(i_y+1) K}{2^{\ell+1}} \Bigg),\\
        B_{j_x, j_y}^{L-\ell} & = \Bigg[\frac{j_x}{2^{L-\ell-1}},
        \frac{(j_x+1)}{2^{L-\ell-1}} \Bigg)
        \times \Bigg[\frac{j_y}{2^{L-\ell-1}},
        \frac{(j_y+1)}{2^{L-\ell-1}} \Bigg).
    \end{split}
\end{equation*}
Figure~\ref{fig:bfexample} gives a 3-layer recursive domain decomposition
example.

We further adopt the notation $\prec$ to denote the relationship among
recursive partitions, i.e., $(\tilde{i}_x, \tilde{i}_y)
\prec (i_x, i_y)$ or $A_{\tilde{i}_x,\tilde{i}_y}^{\ell+1} 
\prec A_{i_x, i_y}^{\ell}$ means that
\begin{equation*}
    A_{\tilde{i}_x,\tilde{i}_y}^{\ell+1} \subset A_{i_x, i_y}^{\ell}
\end{equation*}
for some $\ell$. Similarly, $(j_x, j_y) \prec (\tilde{j}_x, \tilde{j}_y)$,
or $B_{j_x, j_y}^{L-\ell} \prec B_{\tilde{j}_x,\tilde{j}_y}^{L-\ell-1}$,
means
\begin{equation*}
    B_{j_x, j_y}^{L-\ell} \subset B_{\tilde{j}_x,\tilde{j}_y}^{L-\ell-1}
\end{equation*}
for some $\ell$. The layer index $\ell$ in all cases can be referred from
the text around. As in Figure~\ref{fig:bfexample}, we have 
\begin{equation*}
\begin{split}
    B_{00}^2 \prec B_{00}^1 \prec B_{00}^0,\\
    A_{00}^2 \prec A_{00}^1 \prec A_{00}^0.
\end{split}
\end{equation*}

\paragraph{Low-Rank Approximation of Fourier Kernel.} The Fourier kernel,
\begin{equation*}
    \calK(\xi, t) = e^{-2 \pi \imath \xi \cdot t},
    \quad \xi \in [0, K) \times[0, K) \text{ and }
    t \in [0, 1) \times [0, 1),
\end{equation*}
is a Fourier integral operator. Different from general multi-dimensional
Fourier integral operator, the Fourier kernel does not have a singularity
around the origin. Hence the low-rank approximation theorem for
one-dimensional Fourier integral operator could be extended to the
two-dimensional Fourier kernel.

\begin{theorem} \label{thm:low-rank}
    Let $A \subset [0, K)^2$ and $B \subset [0, 1)^2$ be a domain pair
    such that $\gamma = \frac{e \pi \omega(A) \omega(B)}{r^2} <
    1$~\footnote{$\omega(\cdot)$ is the sidelength function of a domain.},
    where $r^2$ is the number of Chebyshev points. Then the Fourier kernel
    restricted to the domain pair admits both low-rank approximations,
    \begin{equation*}
        \begin{split}
            \sup_{\xi \in A, t \in B} \left| e^{-2\pi \imath \xi \cdot t}
            - \sum_{k_x = 1}^r \sum_{k_y = 1}^r e^{-2 \pi \imath \xi \cdot 
            t_{k_x, k_y}} e^{-2 \pi \imath \xi_0 \cdot (t - t_{k_x, k_y })}
            \calL_{k_x, k_y}(t) \right| \leq & C
            \frac{\gamma^{r^2}}{1 - \gamma}, \text{ and} \\
            \sup_{\xi \in A, t \in B} \left| e^{-2 \pi \imath \xi \cdot t}
            - \sum_{k_x = 1}^r \sum_{k_y = 1}^r \calL_{k_x, k_y}(\xi)
            e^{-2 \pi \imath (\xi - \xi_k) \cdot t_0} e^{-2 \pi \imath
            \xi_{k_x, k_y} \cdot t} \right| \leq & C
            \frac{\gamma^{r^2}}{1 - \gamma},
        \end{split}
    \end{equation*}
    where $\xi_0$ and $t_0$ are centers of $A$ and $B$ respectively, $C$
    is a constant independent of $t$ and $\xi$, $\xi_{k_x,k_y}$ and 
    $t_{k_x,k_y}$ are Chebyshev points on $A$ and $B$.
\end{theorem}
A sketched proof of Theorem~\ref{thm:low-rank} can be found in
Appendix~\ref{app:proof}. From Theorem~\ref{thm:low-rank}, when restricted
to the desired domain pairs, the Fourier kernel can be well-approximated
by a low-rank factorization. An underlying matrix-vector multiplication
can be approximated as,
\begin{equation} \label{eq:low-rank-app}
    \begin{split}
        u^{B}(\xi) = \sum_{t\in B} K(\xi,t) x(t)
        \approx & \sum_{k_x = 1}^r \sum_{k_y = 1}^r
        \alpha_{k_x, k_y}(\xi) \left( \sum_{t\in B}
        \beta_{k_x, k_y}(t) x(t) \right)\\
        = & \sum_{k_x = 1}^r \sum_{k_y = 1}^r
        \alpha_{k_x, k_y}(\xi) \lambda_{k_x, k_y}^{AB},
    \end{split}
\end{equation}
where
\begin{equation*}
    \begin{split}
        \alpha_{k_x, k_y}(\xi) = & e^{-2 \pi \imath \xi \cdot t_{k_x,k_y}}, \\
        \beta_{k_x, k_y}(t) = & e^{-2 \pi \imath \xi_0 \cdot
        (t - t_{k_x, k_y})} \calL_{k_x, k_y}(t), \text{ and} \\
        \lambda_{k_x, k_y}^{AB} = & \sum_{t \in B} \beta_{k_x, k_y}(t)x(t).
    \end{split}
\end{equation*}
Hence, the Fourier transform of the vector $x(t)$ is turned into computing
$\lambda_{k_x, k_y}^{AB}$. The butterfly algorithm employs
\eqref{eq:low-rank-app} recursively to reduce the quadratic computational
cost down to quasi-linear.

\subsection{2D Butterfly Algorithm Revisit}
\label{sec:butterflyalgo}

We revisit the butterfly algorithm applying to the two-dimensional Fourier
kernel. We first conduct a $L$-layer recursive domain decomposition for
both $A^0_{00} = [0, K)^2$ and $B^0_{00} = [0, 1)^2$. The butterfly
algorithm comprises three major steps: interpolation at $\ell=0$,
recursion, and kernel application at $\ell = L$. The interpolation at
$\ell = 0$ interpolates function on a uniform grid in $B^0_{00}$ to
Chebyshev grids on $B^L_{j_x, j_y}$ for all $(j_x, j_y)$. The recursion
step then recursively interpolates the function on four Chebyshev grids at
the finer domain layer to the Chebyshev grid at the coarse domain layer.
Finally, the kernel application step applies the Fourier kernel. The 2D
butterfly algorithm applying to the Fourier kernel is detailed as follows.

\begin{enumerate}
    \item \textbf{Interpolation} ($\ell = 0$). For each domain pair
    $(A_{i_x, i_y}^0, B_{j_x, j_y}^{L})$ at layer $\ell = 0$, $i_x, i_y
    \in [2], j_x, j_y \in [2^{L-1}]$~\footnote{Notation $[n]$ is the set
    of integers, i.e., $[n] = \{0, 1, \dots, n-1\}$.}, we conduct a
    coefficient transfer from the uniform grid in $B_{j_x,j_y}^{L}$ to
    Chebyshev points in the same domain. The constructed expansion
    coefficients admit,
    \begin{equation} \label{eq:bflayer1}
        \lambda_{t_{(k_x, k_y)}}^{A_{i_x, i_y}^0 B_{j_x, j_y}^{L}} =
        \sum_{t_{\text{uni}} \in B_{j_x, j_y}^L} e^{-2 \pi \imath
        \xi_0^{i_x,i_y} \cdot (t_{\text{uni}} - t_{(k_x, k_y)})}
        \calL_{(k_x, k_y)}(t_{\text{uni}}) x(t_{\text{uni}}),
    \end{equation}
    for $t_{(k_x, k_y)} \in B_{j_x, j_y}^L$ being the Chebyshev points
    therein and $\xi_0^{i_x,i_y}$ being the center of $A_{i_x, i_y}^0$.
    Throughout the paper, we add a subscript ``uni'' to indicate the
    uniform grid point in the domain, e.g., $t_{\text{uni}} \in B^L_{j_x,
    j_y}$ denotes the uniform grid points in $B^L_{j_x, j_y}$.

    \item \textbf{Recursion}($\ell = 1, \dots, L-1$). For each domain pair
    ($A_{i_x,i_y}^{\ell-1}$, $B_{j_x,j_y}^{L-\ell+1}$), $i_x, i_y \in
    [2^{\ell}]$, $j_x, j_y \in [2^{L-\ell}]$, we conduct coefficients
    transfer from the Chebyshev points in $B_{j_x,j_y}^{L-\ell+1}$ to the
    Chebyshev points in $B_{\tilde{j}_x,\tilde{j}_y}^{L-\ell}$. The domain
    $B_{\tilde{j}_x, \tilde{j}_y}^{L-\ell}$ is a parent domain of 
    $B_{j_x,j_y}^{L-\ell+1}$, i.e., $B_{j_x,j_y}^{L-\ell+1} \prec
    B_{\tilde{j}_x,\tilde{j}_y}^{L-\ell}$. Similarly, the other domain
    satisfies $A_{\tilde{i}_x,\tilde{i}_y}^{\ell} \prec
    A_{i_x,i_y}^{\ell-1}$. The coefficients transfer admits,
    \begin{equation} \label{eq:bflayer2}
        \lambda_{t_{(k_x, k_y)}}^{A_{\tilde{i}_x,\tilde{i}_y}^{\ell}
        B_{\tilde{j}_x,\tilde{j}_y}^{L-\ell}} =
        \sum_{(j_x,j_y) \prec (\tilde{j}_x,\tilde{j}_y)}
        \sum_{t \in B_{j_x,j_y}^{L-\ell+1}}
        e^{-2 \pi \imath \xi_0 \cdot (t - t_{(k_x, k_y)})}
        \calL_{(k_x, k_y)}(t) \lambda_t^{A_{i_x,i_y}^{\ell-1}
        B_{j_x, j_y}^{L-\ell+1}},
    \end{equation}
    for $t_{(k_x, k_y)} \in B_{j_x, j_y}^{L-\ell}$ and $(\tilde{i}_x,
    \tilde{i}_y) \prec (i_x, i_y)$, where $t$ and $t_{(k_x, k_y)}$ are
    Chebyshev points in $B_{j_x, j_y}^{L-\ell+1}$ and $B_{\tilde{j}_x,
    \tilde{j}_y}^{L-\ell}$ respectively, $\xi_0$ is the center of 
    $A_{\tilde{i}_x, \tilde{i}_y}^{l}$.

    \item \textbf{Kernel Application}($\ell = L$). In the last step, the
    domain pairs are ($A_{i_x,i_y}^L$, $B_{0,0}^{0}$) for $i_x, i_y \in
    [2^{L}]$. All previous layers are transferring coefficients via
    interpolation. This step applies the Fourier kernel to the transferred
    coefficients, and approximate $u(\xi)$ as
    \begin{equation} \label{eq:bflayer3}
        u(\xi_{\text{uni}}) \approx \sum_{t_{(k_x, k_y)} \in B_{0, 0}^{0}}
        e^{-2 \pi \imath \xi_{\text{uni}} \cdot t_{(k_x, k_y)}}
        \lambda_{t_{(k_x,k_y)}}^{A_{i_x,i_y}^{L} B_{0, 0}^{0}},
    \end{equation}
    for $\xi_{\text{uni}}$ being uniform grids in $A_{i_x,i_y}^{L}$. This
    gives us the desired approximation of $u(\xi)$.
\end{enumerate}

Butterfly algorithm, in general, carries a complicated procedure. In this
revisit, we omit most intuition behind operations and review the algorithm
flow. For more details, readers are referred to \cite{li2018butterfly}.

\subsection{ButterflyNet2D}
\label{sec:butterflynet}

This section introduces the network architecture of ButterflyNet2D, which
is a CNN architecture with sparsely connected channels. In the revisit of
the butterfly algorithm, we make a crucial observation that the summation
kernels in \eqref{eq:bflayer1} and \eqref{eq:bflayer2} are independent of
the subdomain $B^\ell_{j_x,j_y}$.
More precisely, the summation kernels in \eqref{eq:bflayer1} and
\eqref{eq:bflayer2},
\begin{equation*}
    e^{-2 \pi \imath
    \xi_0^{i_x,i_y} \cdot (t_{\text{uni}} - t_{(k_x, k_y)})}
    \calL_{(k_x, k_y)}(t_{\text{uni}})
    \quad \text{and} \quad
    e^{-2 \pi \imath \xi_0 \cdot (t - t_{(k_x, k_y)})}
    \calL_{(k_x, k_y)}(t).
\end{equation*}
depend only on the relative distance of either $t_{\text{uni}}$ and
$t_{(k_x, k_y)}$ or $t$ and $t_{(k_x, k_y)}$. Therefore, these two
summation kernels are the same for all $B^{\ell}_{j_x,j_y}$ at the same
$\ell$-layer. The summations as in \eqref{eq:bflayer1} and
\eqref{eq:bflayer2} are convolutions and could be represented by a CNN. In
the following, we construct the ButterflyNet2D architecture, which mimics
the structure of the butterfly algorithm and replace the summation
operations by convolutions.

\paragraph{Network Architecture.}

We introduce the construction of a ButterflyNet2D with $L$ layers. The
input is denoted as $Z^{(0)}$, whose size is $\omega_x2^{L-1} \times
\omega_y2^{L-1}$ and channel size is 1. The generalization to inputs of other
sizes is straightforward. We adopt a channel-first index rule, i.e., the
input is indexed as $Z^{0}(c, i, j)$, where $c = 0$ is the channel index,
$i$ and $j$ range from $0$ to $\omega_x2^{L-1}-1$ and $\omega_y2^{L-1}-1$
respectively are two spacial indices. The desired output is assumed to be
of size $m_x2^{L} \times m_y2^{L}$. The network architecture could then be
constructed as follows.

\begin{enumerate}
    \item \textbf{Interpolation}($\ell = 0$).
    Given $c \in [4r^2]$, the first layer is described as
    \begin{equation} \label{eq:net-interp}
        Z^{(1)}(c, i, j) = \sigma \left\{ B^{(0)}(c) +
        \sum_{\substack{s_x \in [\omega_x]\\ s_y \in [\omega_y]}}
        W^{(0)}(0,c,s_x,s_y)Z^{(0)}(0,\omega_x i+s_x,\omega_y j+s_y)
        \right\},
    \end{equation}
    for $i$, $j = [2^{L-1}]$, where $\sigma\{\cdot\}$ is the nonlinear
    ReLU activation function, $B^{(0)}$ is the bias, $W^{(0)}$ is the
    convolution kernel. The output function at the current layer has
    $4r^2$ channels. This layer is a regular 2D convolution layer with
    input sizes $\omega_x 2^{L-1} \times \omega_y 2^{L-1}$, input channel
    size 1, output channel size $4r^2$, convolution kernel size $\omega_x
    \times \omega_y$, stride $(\omega_x, \omega_y)$.

    \item \textbf{Recursion}($\ell = 1, \dots, L-1$). The input at the
    $\ell$ layer is of size $2^{L-\ell} \times 2^{L-\ell}$ and channel
    size $4^{\ell}r^2$. The convolution operation at the current
    layer admits,
    \begin{equation} \label{eq:net-recursion}
        Z^{(\ell+1)}(c_{\text{o}}, i, j) = \sigma \left\{
        B^{(\ell)}(c_{\text{o}}) + \sum_{\substack{kr^2 \leq c_{\text{i}}
        < (k+1)r^2\\ s_x, s_y \in [2]}}
        W^{(\ell)}(c_{\text{i}}, c_{\text{o}}, s_x, s_y)
        Z^{(\ell)}(c_{\text{i}}, 2i + s_x, 2j + s_y) \right\}, 
    \end{equation}
    where $i, j = [2^{L-\ell-1}]$, $c_{\text{o}} \in [4^{\ell+1}r^2]$, and
    $k = c_{\text{o}}/(4r^2)$. We notice the relation $(\tilde{i}_x,
    \tilde{i}_y)\prec (i_x, i_y)$ in \eqref{eq:bflayer2}. Hence in our network architecture, different from
    the regular fully connected channel 2D convolutional layer, the input
    channels and output channels in \eqref{eq:net-recursion} are sparsely
    connected. We could also view \eqref{eq:net-recursion} as a sequence
    of $4^{\ell}$ regular 2D convolutional layers with input size
    $2^{L-\ell}\times2^{L-\ell}$, output size
    $2^{L-\ell-1}\times2^{L-\ell-1}$, input channel size $r^2$, and output
    channel size $4r^2$, acting to each set of input channels and then
    concatenate the output along the channel direction.
    
    \item \textbf{Kernel Application}($\ell = L$). There are two ways to
    view the kernel application layer. From the convolutional layer point
    of view, we apply 2D convolutional layers with sparse channel
    connection, where the input size is $1 \times 1$. From a dense layer
    point of view, we apply a dense layer to each set of channels in the
    input. From either point of view, we could represent the operation as,
    \begin{equation} \label{eq:net-application}
        Z^{(L+1)}(c_{\text{o}}, 0, 0) = \sigma \left\{
        B^{(L)}(c_{\text{o}})
        + \sum_{kr^2 \leq c_{\text{i}} < (k+1)r^2}
        W^{(L)}(c_{\text{i}}, c_{\text{o}}, 0, 0)
        Z^{(L)}(c_{\text{i}}, 0, 0)\right\},
    \end{equation}
    where $c_{\text{o}} \in [m_x m_y 4^{L}]$.
\end{enumerate}

\begin{figure}[htb]
\begin{center}
    \resizebox{\textwidth}{!}{
    \subfigure{
    \begin{minipage}{0.33\linewidth}
    \includegraphics[width=1.1\linewidth]{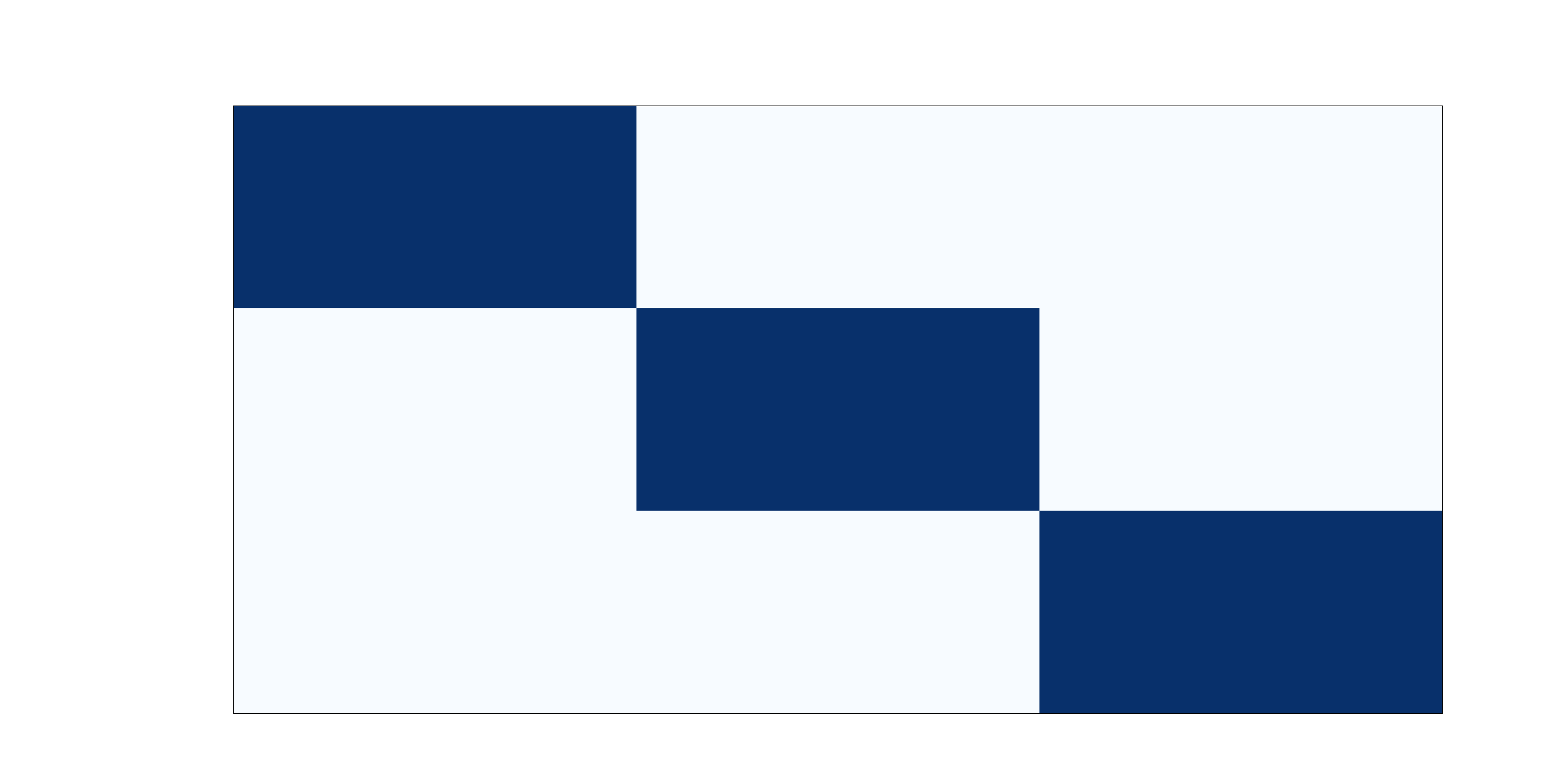}
    \end{minipage}
    \begin{minipage}{0.33\linewidth}
    \includegraphics[width=1.1\linewidth]{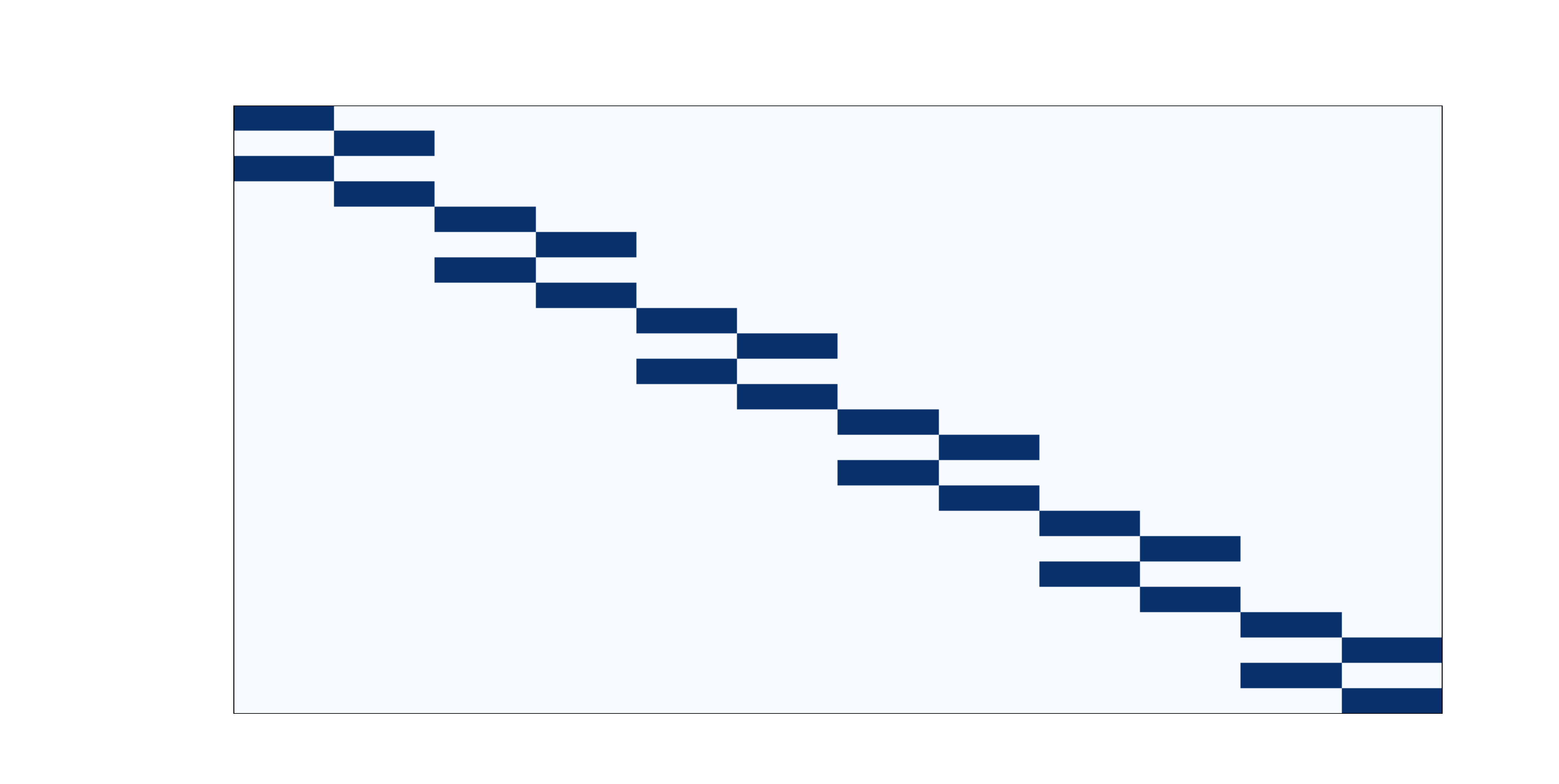}
    \end{minipage}
    \begin{minipage}{0.33\linewidth}
    \includegraphics[width=1.1\linewidth]{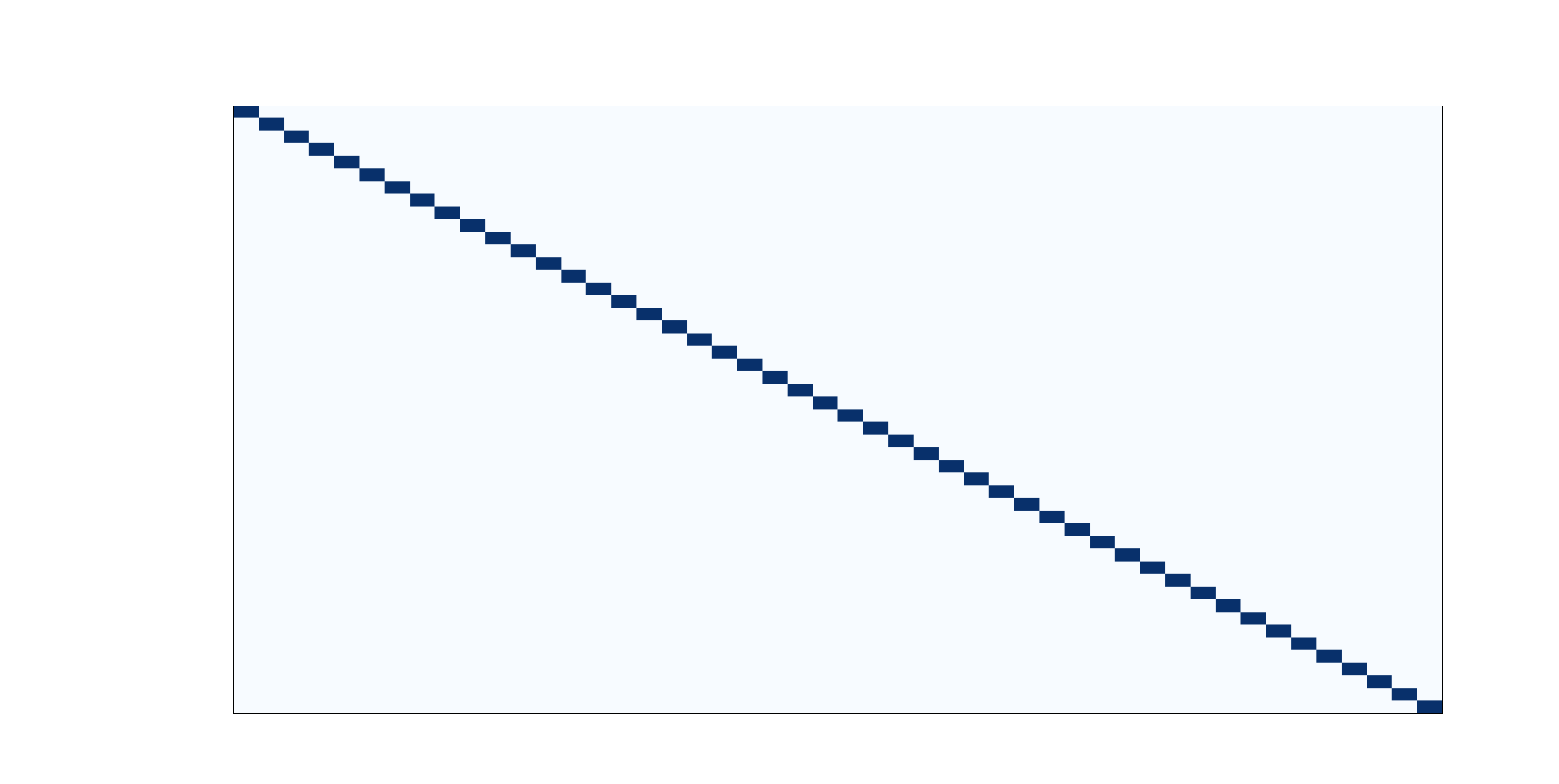}
    \end{minipage}
    } }
    \caption{Input and output channel connection in ButterflyNet2D.}
    \label{fig:sparseChannelConnection}
\end{center}
\end{figure}

An important difference between the ButterflyNet2D and regular 2D CNN is
the connectivity between input and output channels. ButterflyNet2D has a
sparse channel connection, whereas a regular 2D CNN has a dense connection
between input and output channels.
Figure~\ref{fig:sparseChannelConnection} offers an illustration for the
channel connections in ButterflyNet2D. 

\paragraph{Parameter Counting.}

For a ButterflyNet2D with $L$ layers and $r^2$ Chebyshev grid points, the
input and output are assumed to be of size $\omega_x2^{L-1} \times
\omega_y 2^{L-1}$ and $m_x 2^{L} \times m_y 2^{L}$ respectively. We
calculate the number of convolutional kernel weights and bias weights. As
been explained in Appendix~\ref{app:complexNN}, all weights and bias in
\eqref{eq:net-interp}, \eqref{eq:net-recursion}, and
\eqref{eq:net-application}, are $4 \times 4$ real matrices, which would
then be initialized to approximate the complex numbers for Fourier
transform. The calculation for the number of weights is as follows.

\begin{enumerate}
    \item \textbf{Interpolation}($\ell = 0$).
    The number of nonzeros in the convolution kernel $W^{(0)}$ is
    \begin{equation*}
        \mathrm{nnz}(W^{(0)}) = 4 \times (4 \times 4 \times r^2) \times
        (\omega_x \times \omega_y) = 4^3 r^2 \omega_x \omega_y,
    \end{equation*}
    whereas that in the bias $B^{(0)}$ is
    \begin{equation*}
        \mathrm{nnz}(B^{(0)}) = 4 \times (4 \times r^2) = 4^2 r^2.
    \end{equation*}

    \item \textbf{Recursion}($\ell = 1, \dots, L-1$).
    The number of nonzeros in the convolution kernel $W^{(\ell)}$ is
    \begin{equation*}
        \mathrm{nnz}(W^{(\ell)}) = 
        4^{\ell}\times (4\times r^2) \times 
        (4\times 4\times r^2) \times 
        (2\times2)= 4^{\ell+4}r^4,
    \end{equation*}
    whereas that in the bias $B^{(\ell)}$ is
    \begin{equation*}
        \mathrm{nnz}(B^{(\ell)}) = 4^{\ell} \times (4 
        \times 4 \times r^2) = 4^{\ell+2}r^2.
    \end{equation*}
    Hence summing over all layers, the total number of weights in
    convolution kernel and bias in the recursion step admits,
    \begin{equation*}
        \sum_{\ell = 1}^{L-1} \mathrm{nnz}(W^{(\ell)})
        = \frac{4^{L+4}-4^5}{3}r^4, \quad
        \sum_{\ell = 1}^{L-1} \mathrm{nnz}(B^{(\ell)})
        = \frac{4^{L+2}-4^3}{3}r^2.
    \end{equation*}

    \item \textbf{Kernel Application}($\ell = L$).
    The number of nonzeros in the convolution kernel $W^{(L)}$ is
    \begin{equation*}
        \mathrm{nnz}(W^{(\ell)}) =
        4^{L}\times
        (4 \times r^2)\times(4\times m_x\times m_y) \times 
        (1 \times 1) =
        4^{L+2} r^2 m_x m_y,
    \end{equation*}
    whereas that in the bias $B^{(L)}$ is
    \begin{equation*}
        \mathrm{nnz}(B^{(L)}) = 4^{L}\times (4\times m_x \times m_y)
        = 4^{L+1} m_x m_y.
    \end{equation*}
\end{enumerate}

With an input that is of the size $N\times N$, the number of layers in the
network could be $L = \log N$. The overall number of weights is
\begin{equation*}
    4^2r^2(1+4\omega_x\omega_y) + 
    \frac{4^{L+2}-4^3}{3}r^2(1+4^2r^2) + 
    4^{L+1}m_x m_y (1+4r^2) = O(N).
\end{equation*}
When a similar CNN is considered and channels are fully connected, the
number of weights would be 
\begin{equation*}
    4^2r^2(1+4\omega_x\omega_y) + \frac{4^{L+2}-4^3}{3} + 
    \frac{4^{2L+8}-4^6}{15} + 4^{L+1}m_x m_y (1+4r^2) = O(N^2+N). 
\end{equation*}
Hence, ButterflyNet2D, compared to the regular CNN, reduces the number of
weights by a factor of $O(N)$.

\subsection{Fourier Initialization}
\label{sec:fourierinit}

The ButterflyNet2D is constructed in a way mimicking the butterfly
algorithm. Extra bias weights and ReLU activation functions are added. We
now propose an initialization strategy, which transfers coefficients in
the butterfly algorithm to convolution kernels in ButterflyNet2D. Given
the same input and output sizes, and the same number of Chebyshev points,
when the initialization strategy is applied, the ButterflyNet2D is
identical to the butterfly algorithm. We adopt a re-indexing of the
channel,
\begin{equation*}
    c = (i_x, i_y, k_x, k_y)
\end{equation*}
for subdomain $A_{i_x,i_y}^{\ell}$ and the Chebyshev node index $(k_x,
k_y)$. In the initialization strategy, named Fourier initialization, all
bias weights are initialized to be zero. The convolution kernel weights
are initialized as follows, where complex coefficients are converted to $4
\times 4$ matrixes as in Appendix~\ref{app:complexNN}.

\begin{enumerate}
    \item \textbf{Interpolation}($\ell = 0$).
     Since the summation in
    \eqref{eq:bflayer1} is a convolution, the dependence on $B^{(L)}_{j_x,
    j_y}$ could be ignored. We initialize the convolution kernel $W^{(0)}$
    as,
    \begin{equation*}
        W^{(0)}(0, c, s_x, s_y) = e^{-2 \pi \imath \xi_0^{i_x, i_y} \cdot
        (u_{(s_x, s_y)} - t_{(k_x, k_y)})} \calL_{(k_x, k_y)}(u_{(s_x, s_y)}),
    \end{equation*}
    where $u_{(s_x,s_y)}$ and $t_{(k_x, k_y)}$ are the uniform grid and
    Chebyshev grid in $B_{0, 0}^{L}$ respectively.

    \item \textbf{Recursion}($\ell = 1, \dots, L-1$). The summation in
    \eqref{eq:bflayer2} is a convolution with a sparse channel connection,
    the dependence on $B^{(L-\ell+1)}_{j_x, j_y}$ could be ignored. We
    initialize the convolution kernel $W^{(\ell)}$ as
    \begin{equation*}
        W^{(\ell)}(c_{\text{i}}, c_{\text{o}}, s_x, s_y)
        = e^{-2 \pi \imath \xi_0 \cdot (u_{(s_x, s_y)} -
        t_{(k_x, k_y)})} \calL_{(k_x, k_y)}(u_{(s_x, s_y)}),
    \end{equation*}
    where $u_{(s_x, s_y)}$ is the Chebyshev node in $B_{j_x,
    j_y}^{L-\ell+1}$, where $j_x \in [2], j_y \in [2]$. And 
    $t_{(k_x, k_y)}$ is the Chebyshev node in $B_{0,
    0}^{L-\ell}$. Importantly, the input and output channel indices
    $c_{\text{i}}$ and $c_{\text{o}}$ are re-indexed as
    $(i_x,i_y,s_x,s_y)$ and $(\tilde{i}_x,\tilde{i}_y,k_x,k_y)$, such that
    $A_{\tilde{i}_x, \tilde{i}_y}^{\ell+1} \prec A_{i_x, i_y}^{\ell}$.

    \item \textbf{Kernel Application}($\ell = L$). For the domain pair
    $(A_{i_x,i_y}^{L}, B_{0,0}^{0})$, the kernel weights connects
    \eqref{eq:bflayer3} and \eqref{eq:net-application} as,
    \begin{equation*}
        W^{(L+1)}(c_{\text{i}}, c_{\text{o}}, 0, 0) =
        e^{ -2 \pi \imath \xi_{(k_x, k_y)} \cdot (t_{(s_x, s_y)})},
    \end{equation*}
    where $t_{(s_x, s_y)}$ is the Chebyshev node in $B_{0,0}^{0}$ and
    $\xi_{(k_x,k_y)}$ is the uniform grid in $A_{i_x,i_y}^{L}$.
\end{enumerate}

Both the Fourier kernel and the inverse Fourier kernel admit the low-rank
approximation property in Theorem~\ref{thm:low-rank}. Hence, we could also
initialize the ButterflyNet2D to approximate the inverse Fourier
transform. The initialization detail is omitted. Instead, we numerically
demonstrate the performance in the next section.

\section{Numerical Experiments}
\label{sec:num_res}

We implemented ButterflyNet2D together with random and Fourier
initialization in Python, with PyTorch (1.11.0). The code is available at
\url{https://github.com/Genz17/ButterFlyNet2D}. In
Section~\ref{sec:num_fourier}, we first apply the ButterflyNet2D to
approximate the Fourier transform and inverse Fourier transform to verify
the approximation accuracy of the Fourier initialization. Then in
Section~\ref{sec:num_image}, several image processing tasks on practical
image datasets are addressed by ButterflyNet2D.

\subsection{Approximation of Fourier Transform}
\label{sec:num_fourier}

We explore the approximation power of Fourier initialization for the
ButterflyNet2D before and after training. The approximation power is
measured by the relative matrix norm, 
\begin{equation*}
    \epsilon_p := \frac{\|\calB-\calF\|_p}{\|\calF\|_p},
\end{equation*}
where $\calB$, $\calF$ denote the matrix representation of ButterflyNet2D
and discrete (inverse) Fourier transform matrix, respectively.

\paragraph{Approximation Before Training.}
We apply ButterflyNet2D with Fourier initialization to both the Fourier
transform and inverse Fourier transform with $N = 64 \times 64$. The
numerical results are included in Table~\ref{tab:ftapp} and
Table~\ref{tab:iftapp}.

\begin{table}[htb]
\centering
\small
\begin{tabular}{ccccccc}
    \toprule
    & \multicolumn{3}{c}{layer $L$ (with $r = 6$)} &
    \multicolumn{3}{c}{Cheb $r^2$ (with $L = 6$)} \\
    \cmidrule(lr){2-4}
    \cmidrule(lr){5-7}
    & 4 & 5 & 6 & $4^2$ & $5^2$ & $6^2$ \\
    \toprule
    $\epsilon_1$&
    $5.27\times10^{-1}$&
    $3.64\times10^{-2}$&
    $1.72\times10^{-3}$&
    $5.30\times10^{-2}$&
    $8.18\times10^{-3}$&
    $1.72\times10^{-3}$\\
    $\epsilon_2$& 
    $7.71\times10^{-1}$&
    $6.05\times10^{-1}$&
    $1.84\times10^{-3}$&
    $8.20\times10^{-2}$&
    $1.20\times10^{-2}$&
    $1.84\times10^{-3}$\\
    $\epsilon_{\infty}$&
    $8.07\times10^{0}$ &
    $3.73\times10^{-2}$&
    $1.12\times10^{-3}$&
    $6.65\times10^{-2}$&
    $8.16\times10^{-3}$&
    $1.12\times10^{-3}$ \\ 
    \bottomrule
\end{tabular}
\caption{ButterflyNet2D with Fourier initialization approximating Fourier
transform with $N = 64\times 64$ before training.}
\label{tab:ftapp}
\end{table}

\begin{table}[htb]
\centering
\small
\begin{tabular}{ccccccc}
    \toprule
    & \multicolumn{3}{c}{layer $L$ (with $r = 6$)} &
    \multicolumn{3}{c}{Cheb $r^2$ (with $L = 6$)} \\
    \cmidrule(lr){2-4}
    \cmidrule(lr){5-7}
    & 4 & 5 & 6 & $4^2$ & $5^2$ & $6^2$ \\
    \toprule
    $\epsilon_1$&
    $9.04\times10^{-1}$&
    $6.80\times10^{-2}$&
    $3.07\times10^{-3}$&
    $1.07\times10^{-1}$&
    $1.89\times10^{-2}$&
    $3.07\times10^{-3}$ \\
    $\epsilon_2$&
    $1.16\times10^{0}$&
    $7.87\times10^{-2}$&
    $3.10\times10^{-3}$&
    $1.09\times10^{-1}$&
    $1.89\times10^{-2}$&
    $3.10\times10^{-3}$ \\ 
    $\epsilon_{\infty}$&
    $4.19\times10^{0}$&
    $1.76\times10^{-1}$&
    $4.83\times10^{-3}$&
    $1.79\times10^{-1}$&
    $3.03\times10^{-2}$&
    $4.83\times10^{-3}$ \\
    \bottomrule
\end{tabular}
\caption{ButterflyNet2D with Fourier initialization approximating
inverse Fourier transform with $N = 64\times 64$ before training.}
\label{tab:iftapp}
\end{table}

From both Table~\ref{tab:ftapp} and Table~\ref{tab:iftapp}, the relative
error decays exponentially with respect to both the layer number $L$ and
the number of Chebyshev points $r^2$.

\paragraph{Approximation After Training.}
We further train ButterflyNet2D with Fourier initialization in
approximating the Fourier transform and inverse Fourier transform with $N
= 64 \times 64$. The training data is generated by the exact Fourier and
inverse Fourier transforms with uniform random input vectors. The loss
function is the $\ell_2$ relative error. For training, we adopt Adam
optimizer with a learning rate $0.001$ and batch size of 20. The numerical
results after 200 epochs are included in Table~\ref{tab:ftapp-train} and
Table~\ref{tab:iftapp-train}.

\begin{table}[htb]
\centering
\small
\begin{tabular}{ccccc}
    \toprule
    & \multicolumn{2}{c}{layer $L$ (with $r = 6$)} &
    \multicolumn{2}{c}{Cheb $r^2$ (with $L = 6$)} \\ 
    \cmidrule(lr){2-3}
    \cmidrule(lr){4-5}
    & 3 & 4 & $2^2$ & $3^2$ \\ 
    \toprule
    $\epsilon_1$&
    $9.51\times10^{-1}$ & 
    $2.78\times10^{-1}$ & 
    $3.66\times10^{-1}$ & 
    $6.49\times10^{-2}$ \\ 
    $\epsilon_2$&
    $4.96\times10^{-1}$ & 
    $1.64\times10^{-1}$ & 
    $1.97\times10^{-1}$ & 
    $3.74\times10^{-2}$ \\ 
    $\epsilon_{\infty}$&
    $2.58\times10^{-2}$ & 
    $2.08\times10^{-2}$ & 
    $1.93\times10^{-2}$ & 
    $8.54\times10^{-3}$ \\
    \bottomrule
\end{tabular}
\caption{ButterflyNet2D with Fourier initialization approximating Fourier
transform with $N = 64\times 64$ after training.}
\label{tab:ftapp-train}
\end{table}

\begin{table}[htb]
\centering
\small
\begin{tabular}{ccccc}
    \toprule
    & \multicolumn{2}{c}{layer $L$ (with $r = 6$)} &
    \multicolumn{2}{c}{Cheb $r^2$ (with $L = 6$)} \\ 
    \cmidrule(lr){2-3}
    \cmidrule(lr){4-5}
    & 3 & 4 & $2^2$ & $3^2$ \\
    \toprule
    $\epsilon_1$&
    $5.00\times10^{-1}$ & 
    $1.39\times10^{-1}$ & 
    $2.42\times10^{-1}$ & 
    $3.32\times10^{-2}$ \\ 
    $\epsilon_2$&
    $5.02\times10^{-1}$ & 
    $1.60\times10^{-1}$ & 
    $2.62\times10^{-1}$ & 
    $3.60\times10^{-2}$ \\
    $\epsilon_{\infty}$&
    $7.04\times10^{-1}$ & 
    $5.04\times10^{-1}$ & 
    $5.45\times10^{-1}$ & 
    $1.03\times10^{-1}$ \\
    \bottomrule
\end{tabular}
\caption{ButterflyNet2D with Fourier initialization approximating inverse
Fourier transforms with $N = 64\times 64$ after training.}
\label{tab:iftapp-train}
\end{table}

Comparing Table~\ref{tab:ftapp} and Table~\ref{tab:ftapp-train}, we find
that training a Fourier initialized ButterflyNet2D could further improve
the approximation accuracy. ButterflyNet2D with smaller $r^2$ after
training achieves better accuracy than the network with larger $r^2$
without training. A similar conclusion holds for inverse Fourier transform
if we compare Table~\ref{tab:iftapp} and Table~\ref{tab:iftapp-train}. We
further include the training result for ButterflyNet2D with Kaiming random
initialization in Appendix~\ref{app:ftapp-rand-train}. If we compare
the training result for Fourier initialization and Kaiming random
initialization, we find that ButterFlyNet2D with Fourier initialization
achieves better accuracy in approximating Fourier and inverse Fourier
transform. 

\subsection{Image Processing Tasks}
\label{sec:num_image}

We apply ButterflyNet2D to four major image processing tasks: inpainting,
deblurring, denoising, and watermark removal. For the inpainting task, we
adopt a $10\times 10$ mask for $32\times 32$ pictures, and when the
picture size doubles, the mask size doubles. For example, a mask of the
size $80 \times 80$ is applied to input pictures of size $256 \times 256$.
For the deblurring task, the blurring kernel is a $5\times5$ Gaussian
kernel with a standard deviation of $2.5$. For the denoising task, we add
Gaussian noise with mean zero and standard deviation $0.1$. We add $8$
horizontal and vertical black lines to the original input picture for the
watermark removal task. The line width increases with the picture size.
Three picture datasets used are CIFAR10, STL10, and CelebA. They contain
30000, 5000, 30000 pictures, respectively. For testing purposes, we resize
pictures into different sizes. We crop the pictures into $2\times 2$ or
$4\times 4$ parts to make the neural network more efficient.

The neural network is not the vanilla ButterflyNet2D. We adopt the idea of
special-frequence transformation property and construct the neural network
architecture to be a ButterflyNet2D applied after another ButterflyNet2D,
named ButterflyNet2D\textsuperscript{2}.
Both ButterflyNet2Ds have $\log_2(\text{Input Size})$ layers and
$2^2$ Chebyshev points. The first ButterflyNet2D is initialized to
approximate the Fourier transform, whereas the second one is initialized
to approximate the inverse Fourier transform. Hence the neural network is
initialized as an approximation of the identity mapping.

Relative vector 2-norm error is used as the loss function,
\begin{equation*}
    \calL = \sum_{i = 1}^N \frac{\|\calB(x_i) - x_i\|_2}{\|x_i\|_2},
\end{equation*}
where $\calB$ denotes the neural network and $x_i$ is the $i$-th training
data out of $N$ pictures.
The PSNR with a batch size of 256 is used as the measurement of testing
results,
\begin{equation}
    \text{PSNR} = \frac{\sum_{x \in \text{Batch}} -10 \log_{10} \left(
    \|\calB(x)-x\|_2^2/(3 S_x S_y) \right)}{\text{Batch Size}},
\end{equation}
where $\calB(x)$ and $x$ are RGB-colored pictures whose sizes 
are $S_x\times S_y$. Adam optimizer with an
initial learning rate of $2\times 10^{-3}$ is adopted as the minimizer.
Further, ``ReduceLROnPlateau'' learning rate decay strategy is applied
with a factor of 0.98 and patience of 100. All the details of epochs and
batch sizes are shown in Table~\ref{tab:traindetails}.

\begin{table}[htb]
    \centering
    \small
    \begin{tabular}{ccccc}
    \toprule
        Task & Dataset & Input Size & Training Epochs & Batch Size  \\
        \cmidrule(lr){1-5}
        \multirow{5}*{\parbox[c]{2cm}{Inpainting\\ Denoising\\ Deblurring}} 
        & CelebA($64\times64$)& $32\times32$ & 12 & 20 \\ 
        & CelebA($128\times128$)& $64\times64$ & 12 & 20 \\ 
        & CelebA($256\times256$) & $64\times64$ & 3 & 5 \\ 
        & CIFAR10($32\times32$) & $32\times32$ & 12 & 20 \\ 
        & STL10($64\times64$) & $32\times32$ & 72 & 20\\
        \cmidrule(lr){1-5}
        \multirow{5}*{Watermark Removal}
        & CelebA($64\times64$)& $16\times16$ & 3 & 5 \\
        & CelebA($128\times128$)& $32\times32$ & 3 & 5 \\
        & CelebA($256\times256$)& $32\times32$ & 3 & 5 \\
        & CIFAR10($32\times32$)& $16\times16$ & 12 & 20 \\
        & STL10($64\times64$)& $16\times16$ & 12 & 5\\
        \bottomrule
    \end{tabular}
    \caption{Batch sizes and epochs for various tasks on various datasets.}
    \label{tab:traindetails}
\end{table}

All the pictures in the datasets are RGB-colored pictures. We turn them
into grayscale for training. Then the trained
ButterflyNet2D\textsuperscript{2} is applied to each of the three color
channels of testing pictures. The three outputs are concatenated along the
channel direction and form an RGB-colored picture.

\begin{table}[htb]
\centering
\begin{tabular}{cccccc}
    \toprule
    & & \multicolumn{3}{c}{Initialization} \\
    \cmidrule(lr){3-5}
    Task & Dataset & Fourier & Uniform & Normal \\
    \toprule
    \multirow{5}*{Inpaint} 
    & CelebA($64\times64$)
    & $30.25$ & $16.51$ & $17.39$ \\
    & CelebA($128\times128$) 
    & $30.55$ & $17.91$ & $18.39$ \\
    &CelebA($256\times256$) 
    & $30.83$ & $17.63$ & $18.49$ \\
    & CIFAR10($32\times32$) 
    & $28.73$ & $15.97$ & $17.23$ \\
    & STL10($64\times64$) 
    & $25.77$ & $16.03$ & $16.87$ \\
    \toprule
    \multirow{5}*{Deblur}
    & CelebA($64\times64$) 
    & $36.27$ & $16.16$ & $17.27$ \\
    & CelebA($128\times128$) 
    & $38.30$ & $17.93$ & $18.02$ \\
    & CelebA($256\times256$) 
    & $43.30$ & $17.60$ & $17.49$ \\
    & CIFAR10($32\times32$) 
    & $40.39$ & $16.35$ & $17.14$ \\
    & STL10($64\times64$) 
    & $33.88$ & $16.16$ & $16.99$ \\
    \toprule
    \multirow{5}*{Denoise}
    & CelebA($64\times64$) 
    & $26.71$ & $16.18$ & $17.55$ \\
    & CelebA($128\times128$) 
    & $29.07$ & $17.30$ & $18.15$ \\
    & CelebA($256\times256$) 
    & $32.02$ & $18.17$ & $18.59$ \\
    & CIFAR10($32\times32$) 
    & $26.33$ & $16.38$ & $16.57$ \\
    & STL10($64\times64$) 
    & $26.57$ & $15.40$ & $17.10$ \\
    \toprule
    \multirow{5}*{Watermark Removal}
    & CelebA($64\times64$)
    & $31.63$ & $18.25$ & $18.02$ \\
    & CelebA($128\times128$) 
    & $35.14$ & $16.02$ & $17.18$ \\
    & CelebA($256\times256$) 
    & $41.69$ & $16.03$ & $17.31$ \\
    & CIFAR10($32\times32$) 
    & $31.13$ & $17.00$ & $16.08$ \\
    & STL10($64\times64$) 
    & $32.29$ & $17.48$ & $17.54$ \\
    \bottomrule
\end{tabular}
\caption{The numerical results of ButterflyNet2D\textsuperscript{2}.}
\label{tab:resBF}
\end{table}

\begin{figure}[htb]
    \centering
    \subfigure{
    \begin{minipage}{\linewidth}
    \centering
    \includegraphics[width=0.1\linewidth]{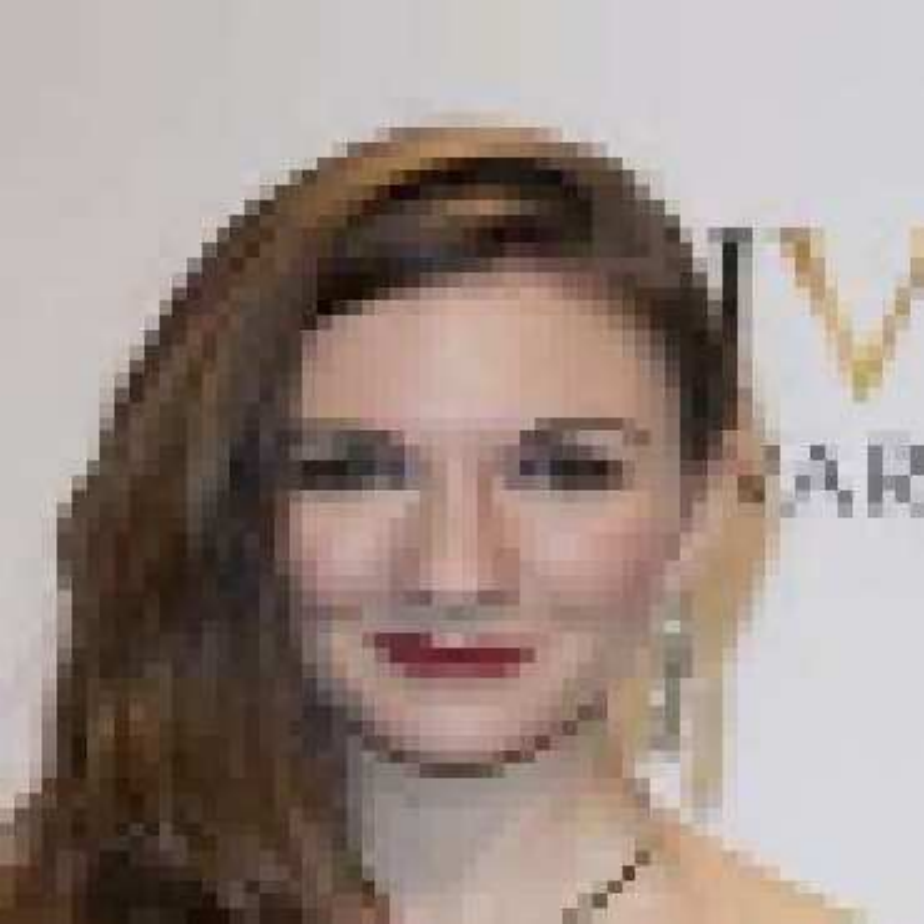}
    \includegraphics[width=0.1\linewidth]{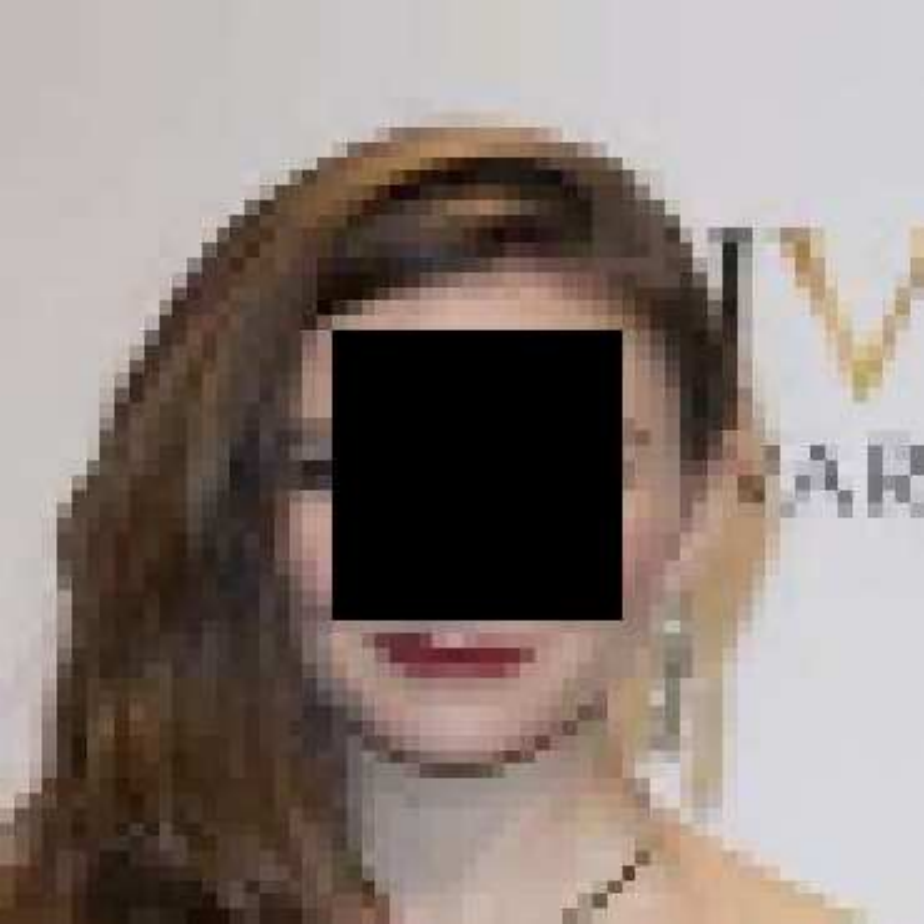}
    \includegraphics[width=0.1\linewidth]{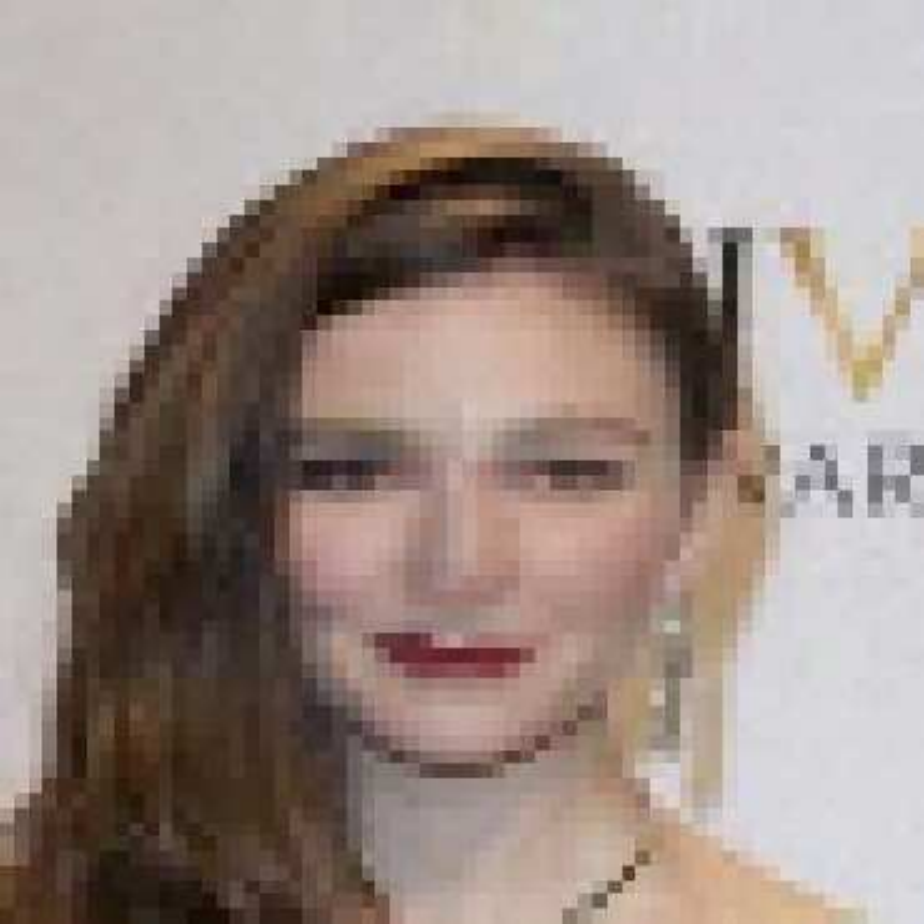}
    \includegraphics[width=0.1\linewidth]{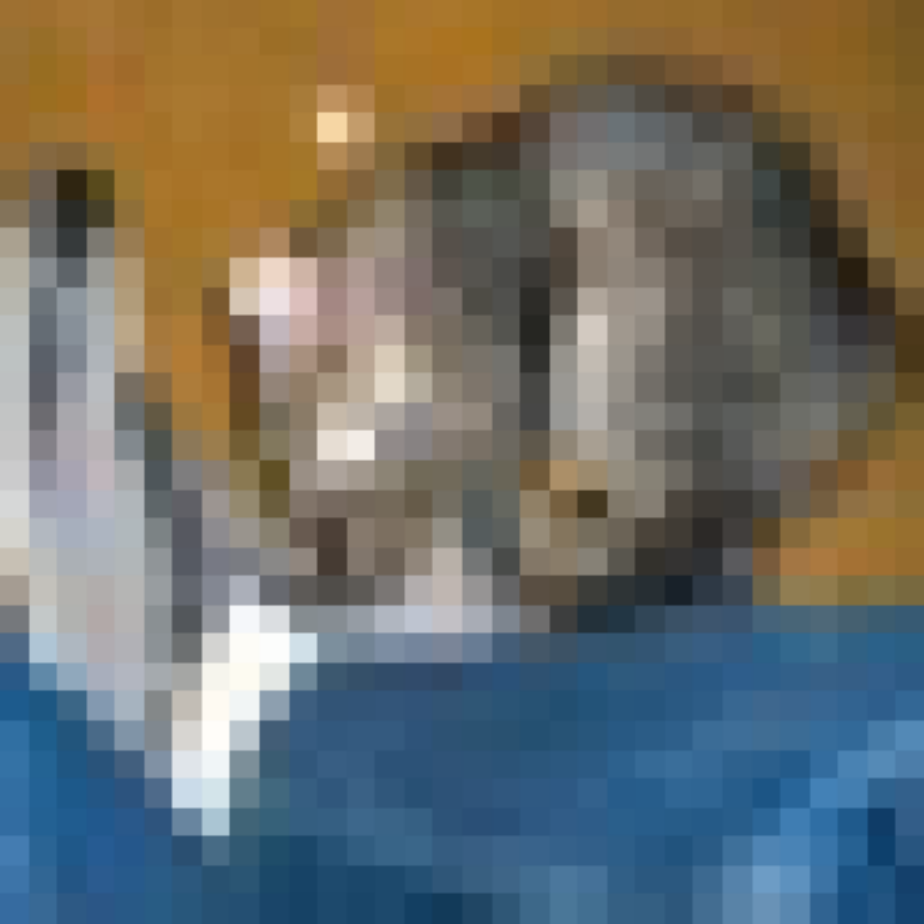}
    \includegraphics[width=0.1\linewidth]{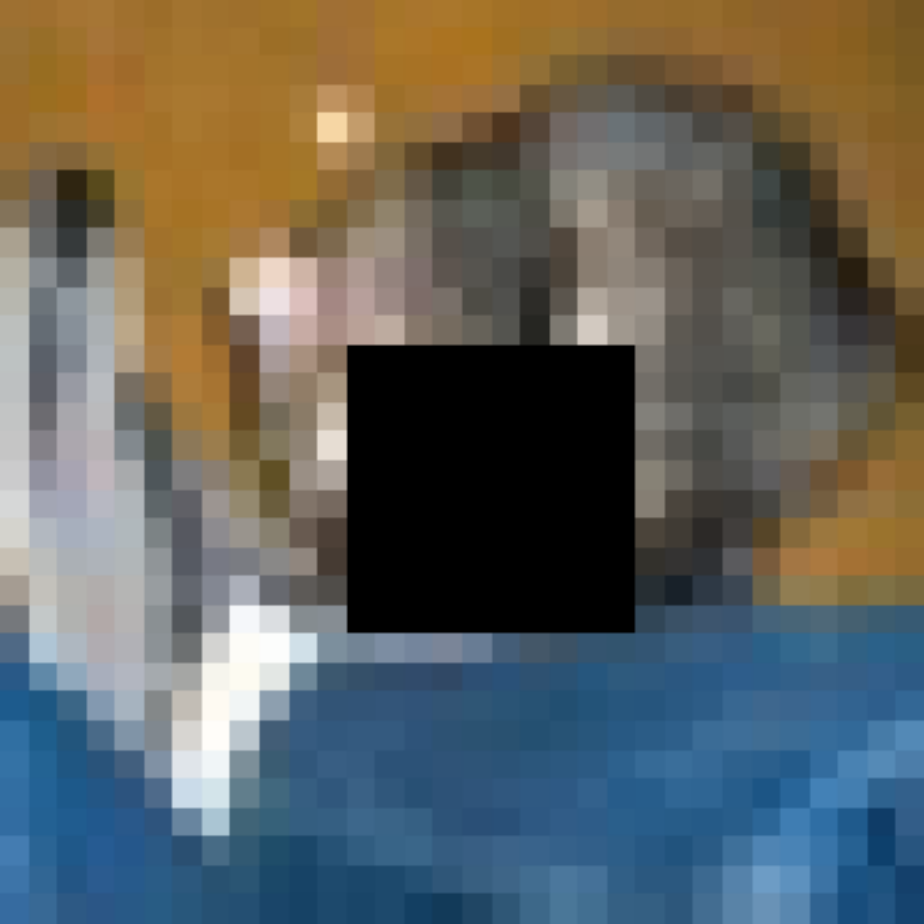}
    \includegraphics[width=0.1\linewidth]{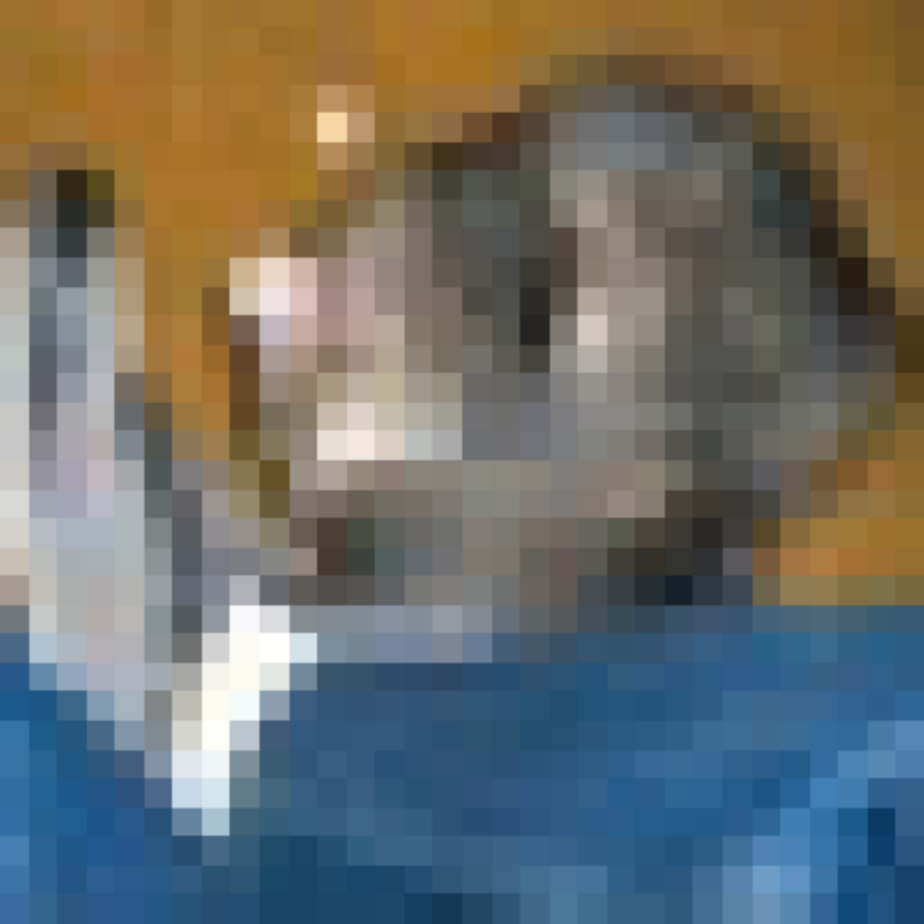}
    \includegraphics[width=0.1\linewidth]{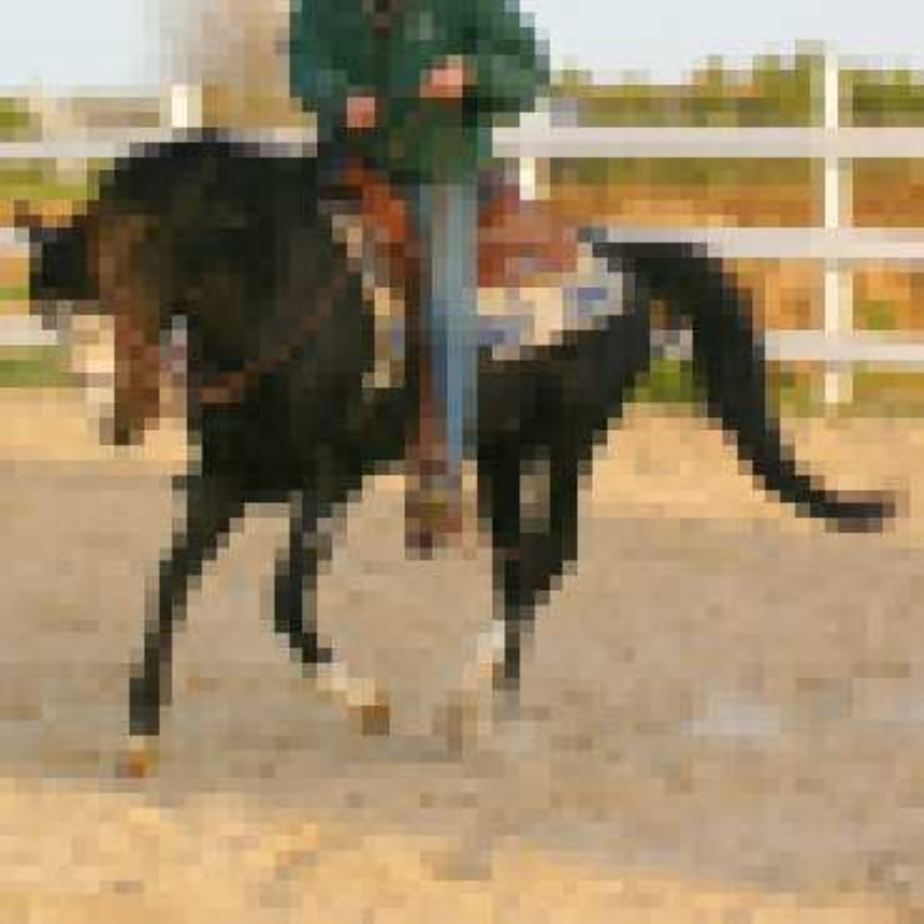}
    \includegraphics[width=0.1\linewidth]{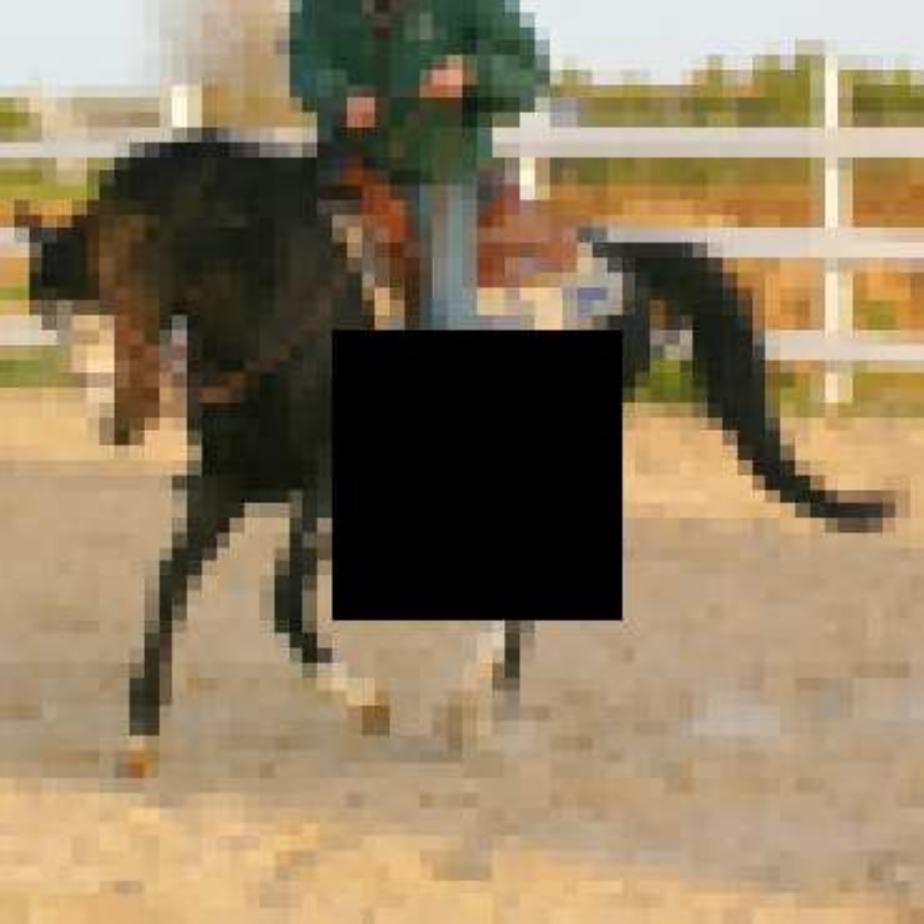}
    \includegraphics[width=0.1\linewidth]{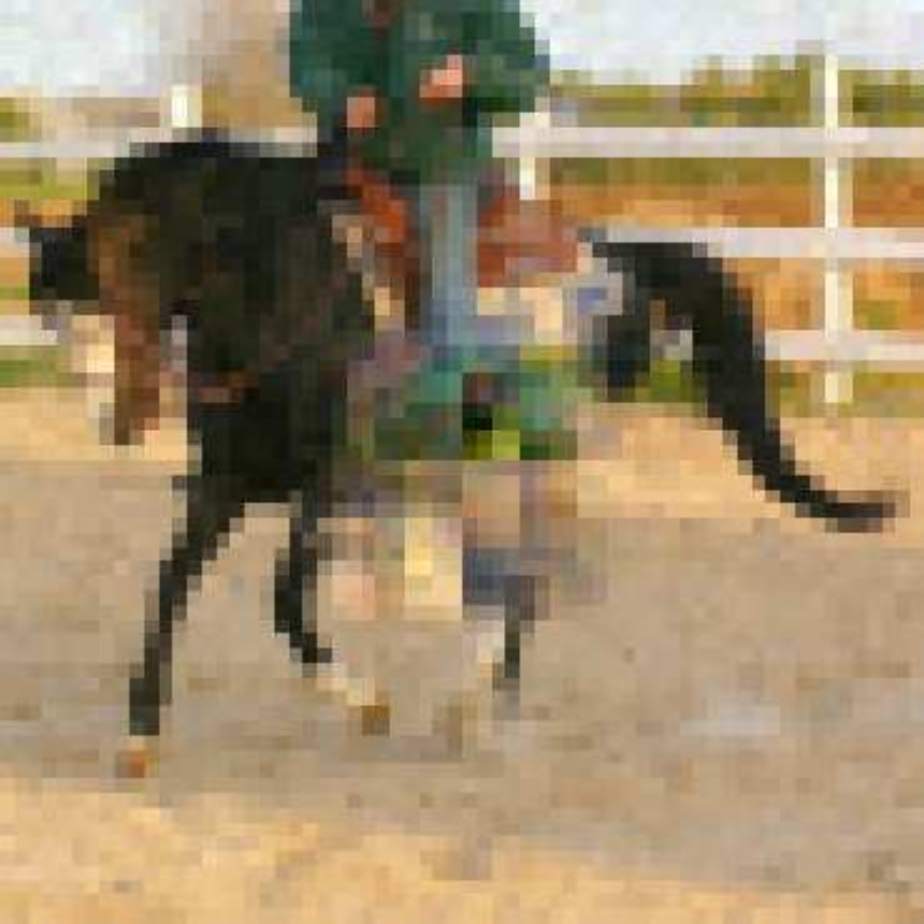}
    \\
    \includegraphics[width=0.1\linewidth]{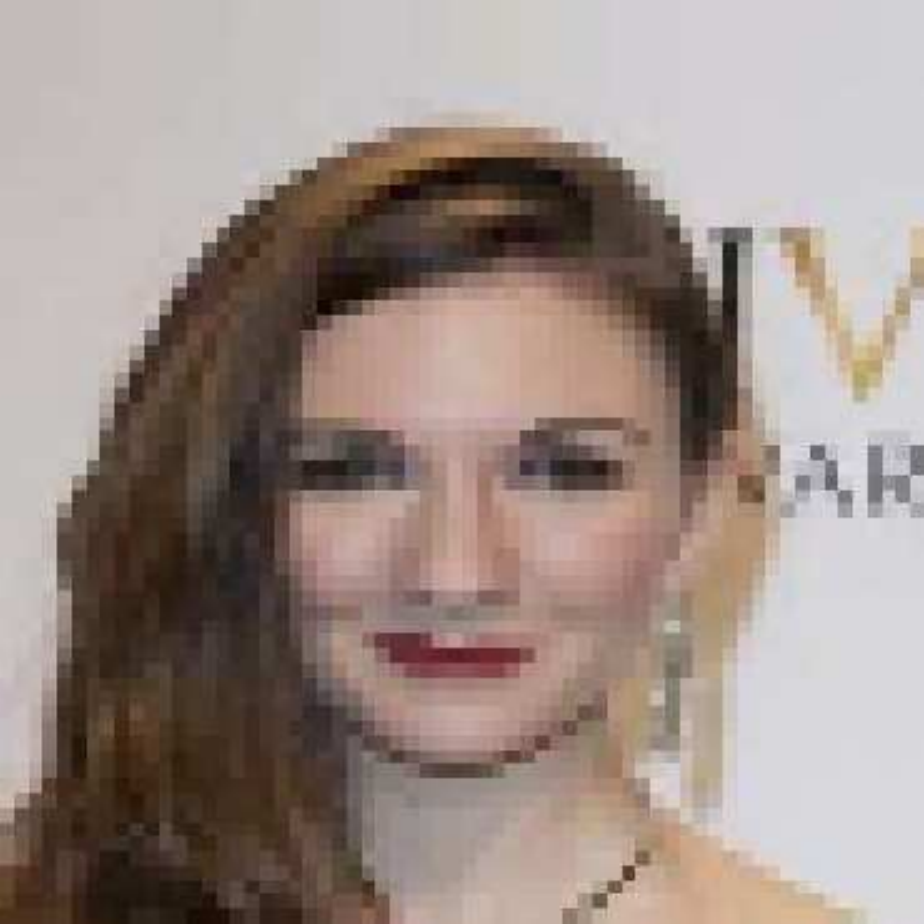}
    \includegraphics[width=0.1\linewidth]{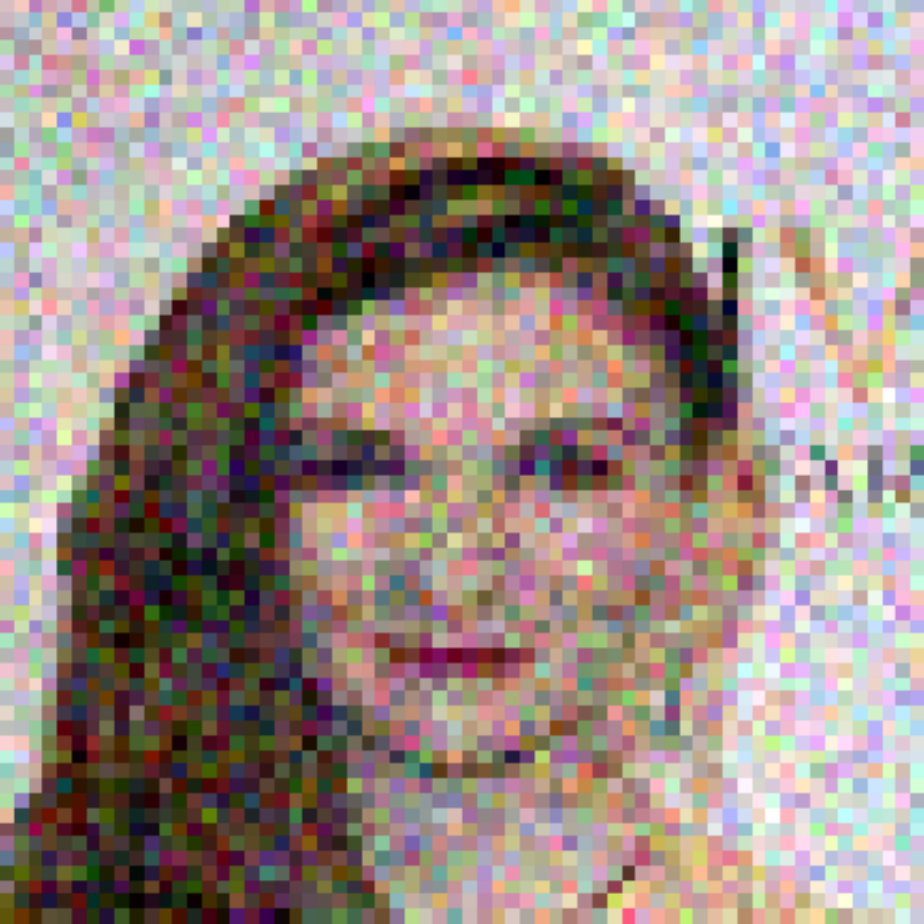}
    \includegraphics[width=0.1\linewidth]{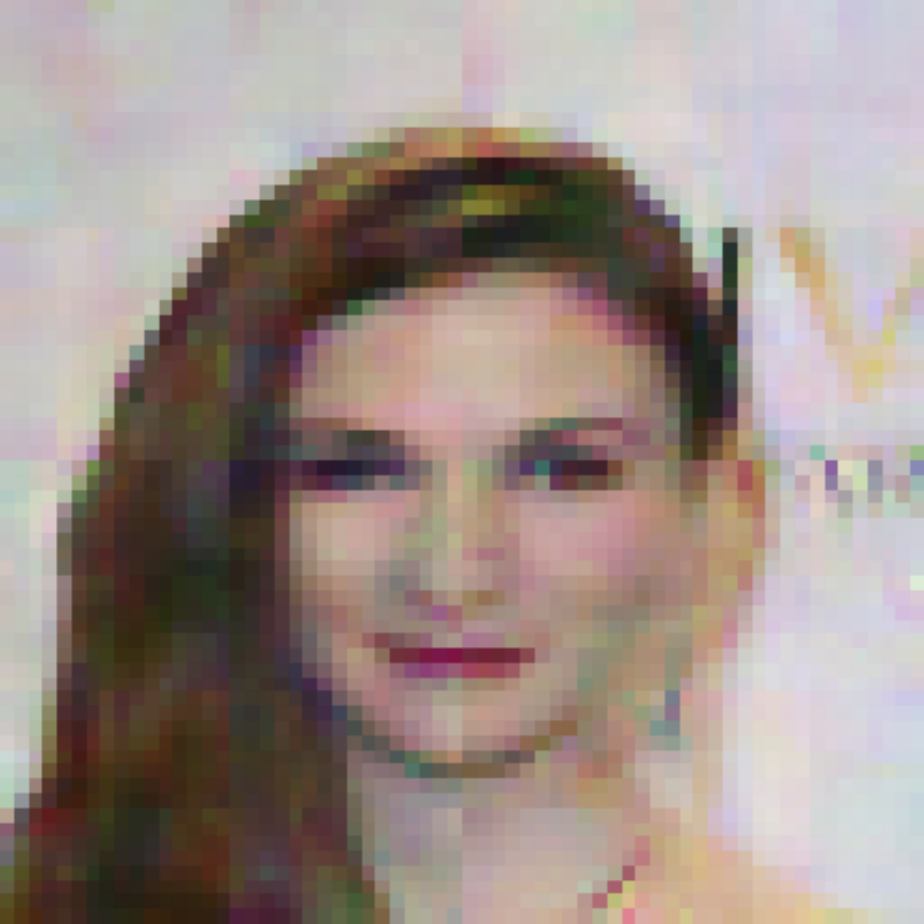}
    \includegraphics[width=0.1\linewidth]{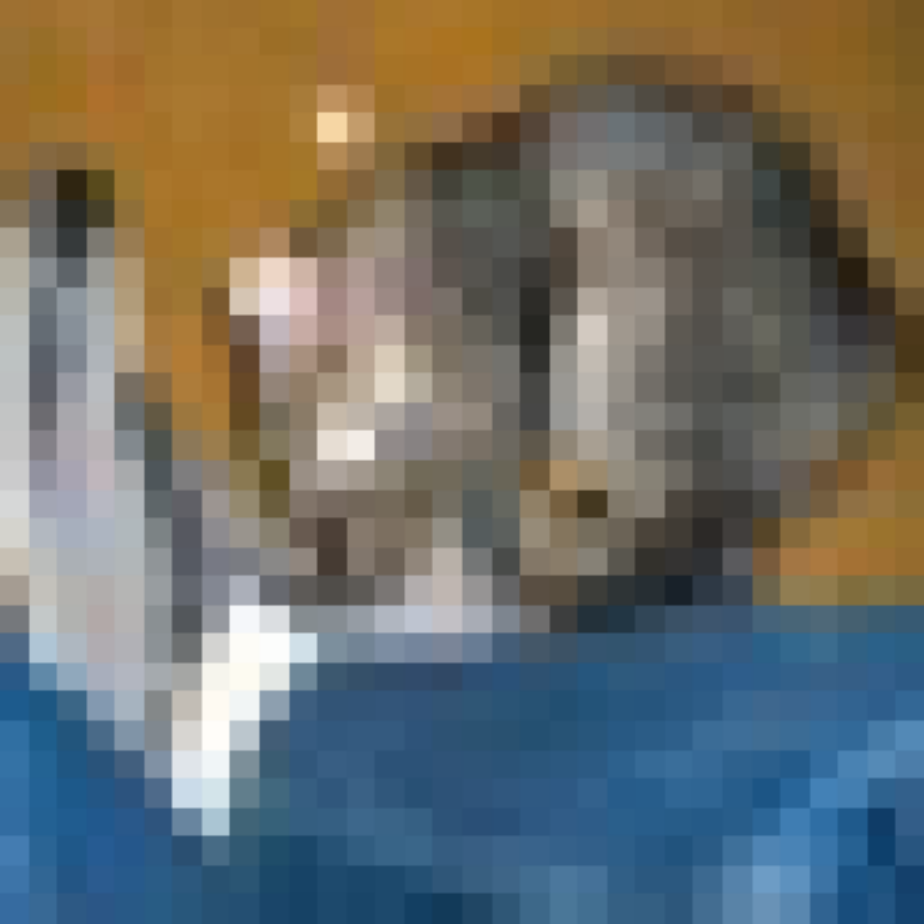}
    \includegraphics[width=0.1\linewidth]{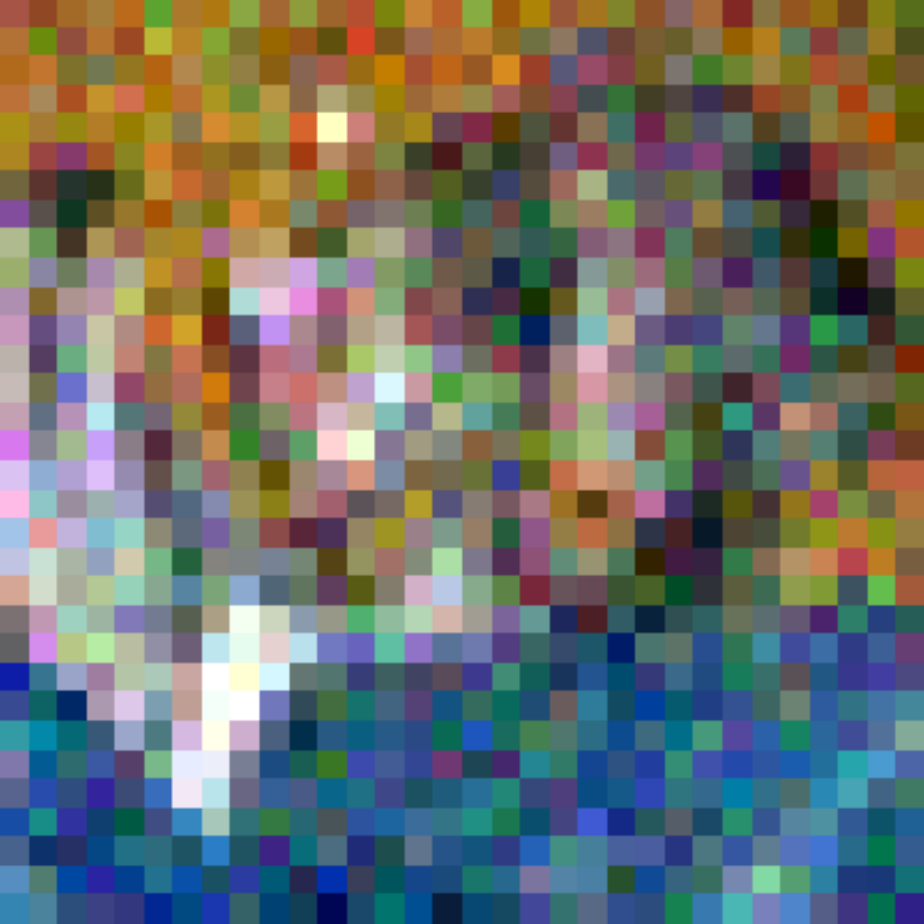}
    \includegraphics[width=0.1\linewidth]{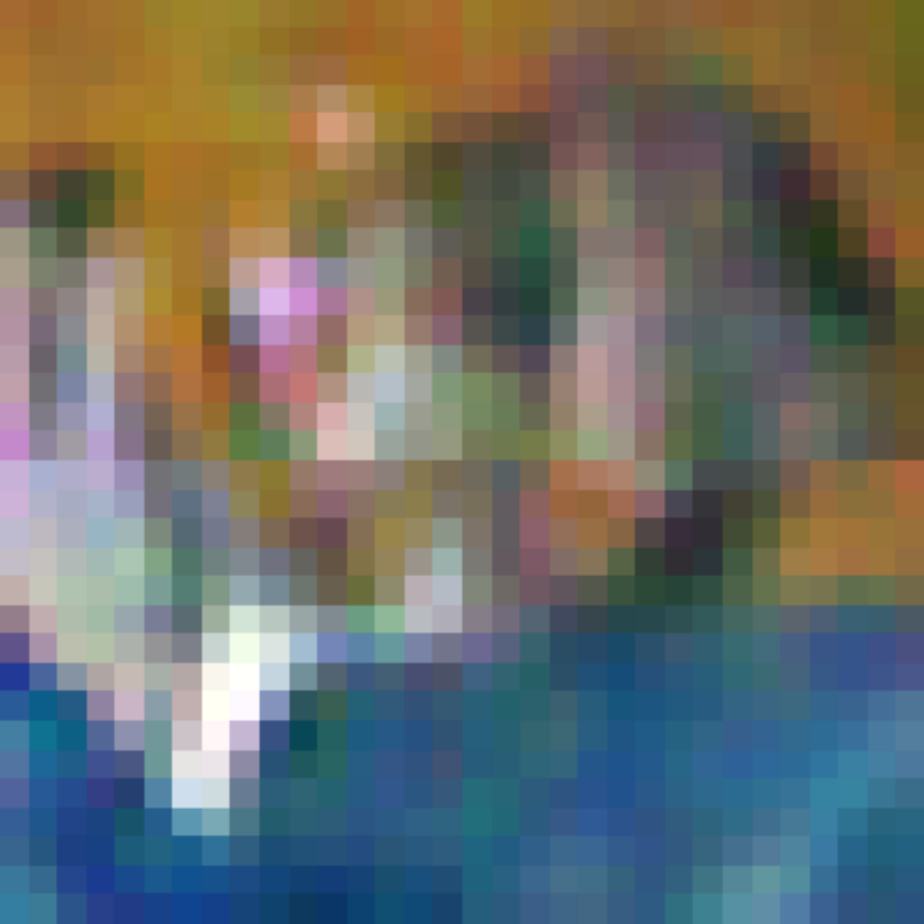}
    \includegraphics[width=0.1\linewidth]{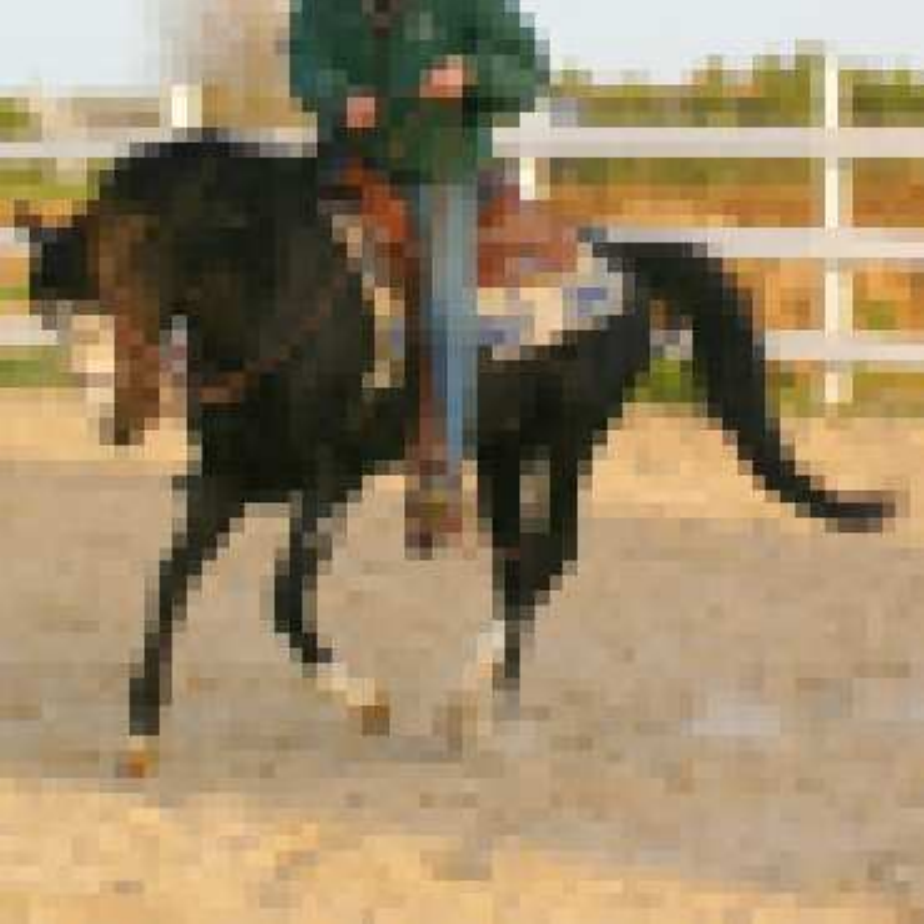}
    \includegraphics[width=0.1\linewidth]{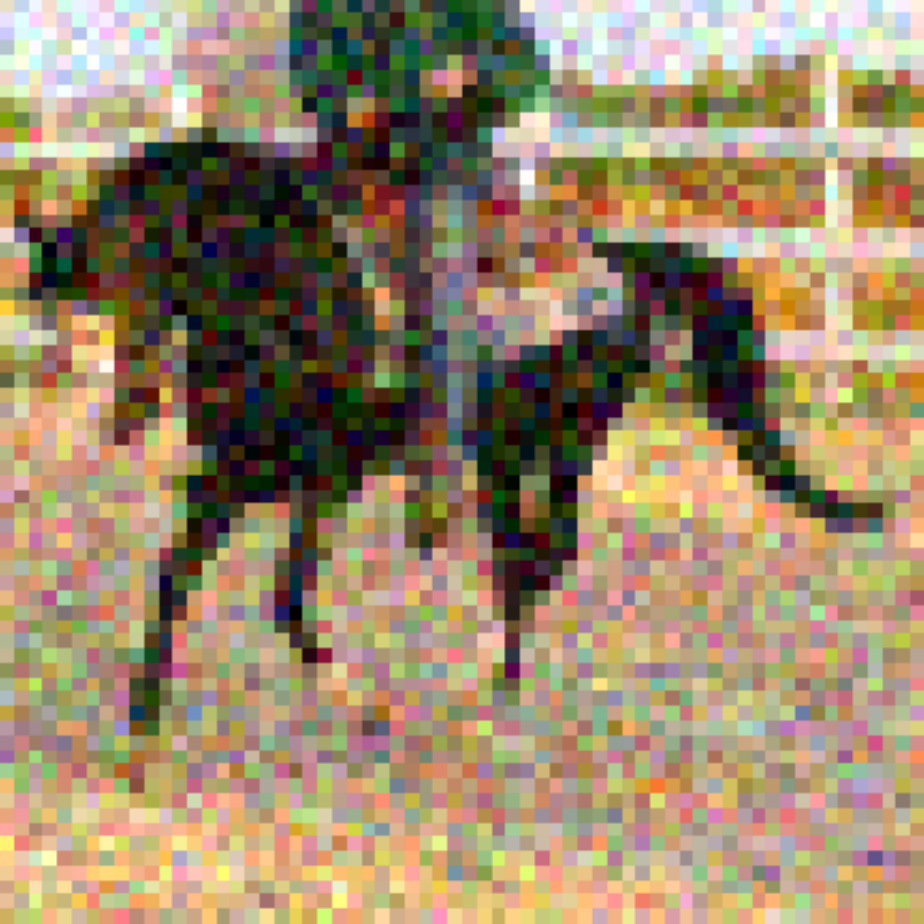}
    \includegraphics[width=0.1\linewidth]{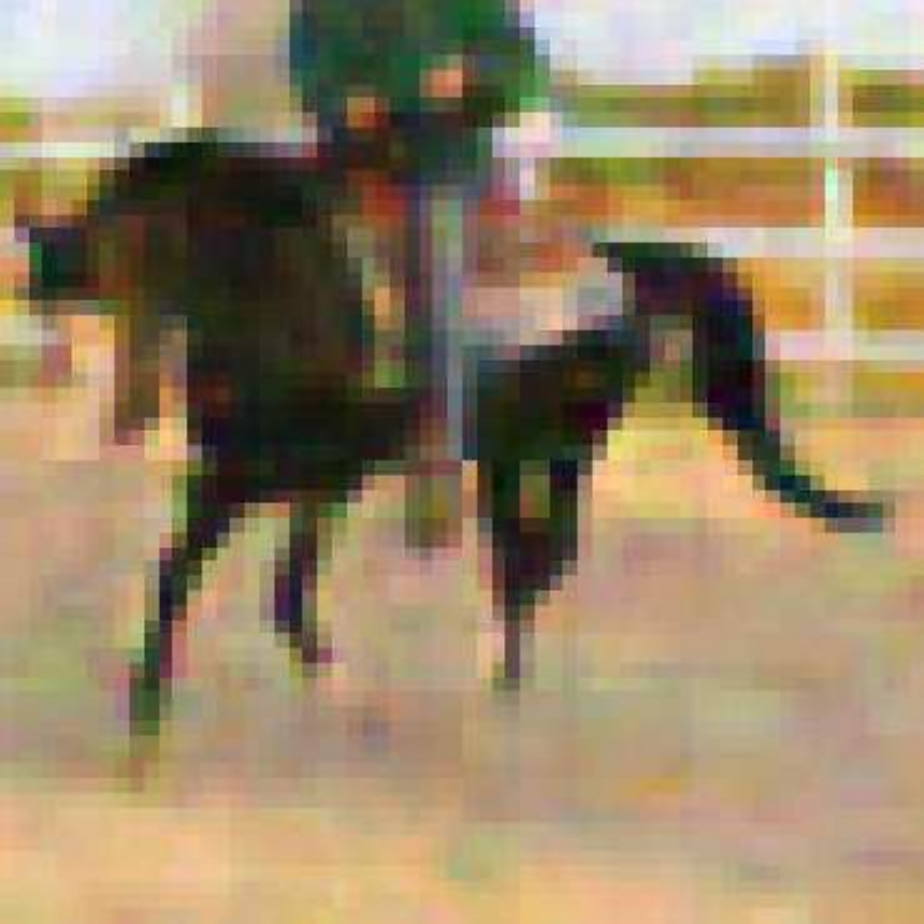}
    \\
    \includegraphics[width=0.1\linewidth]{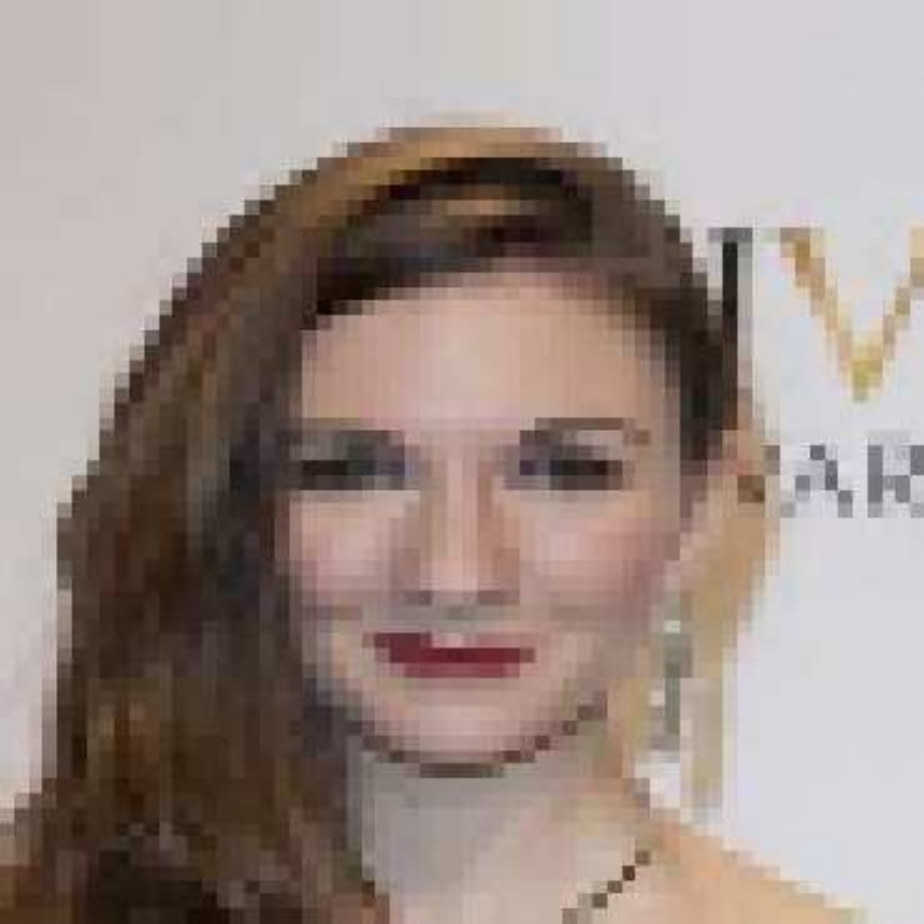}
    \includegraphics[width=0.1\linewidth]{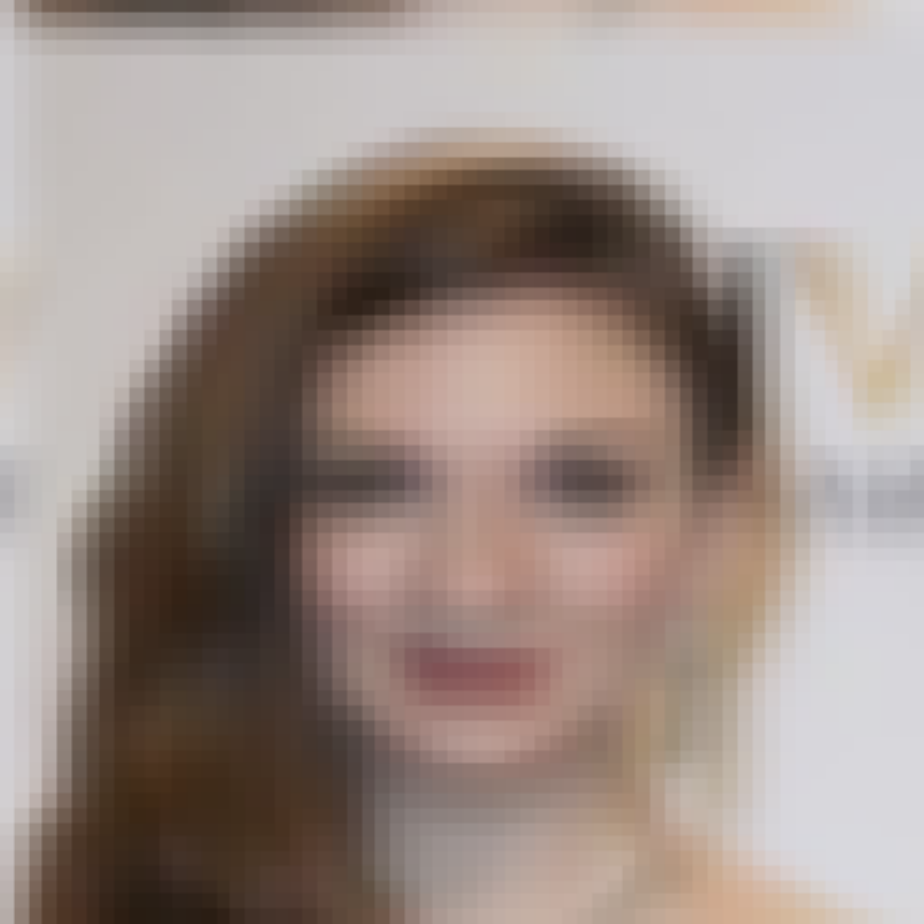}
    \includegraphics[width=0.1\linewidth]{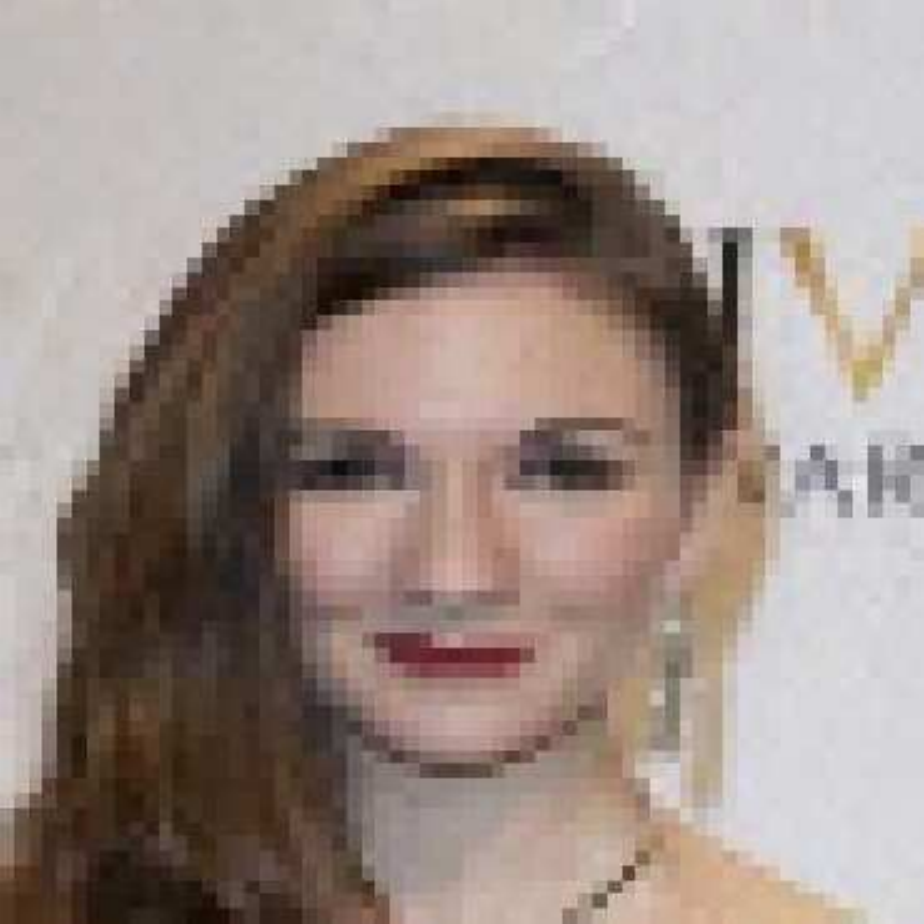}
    \includegraphics[width=0.1\linewidth]{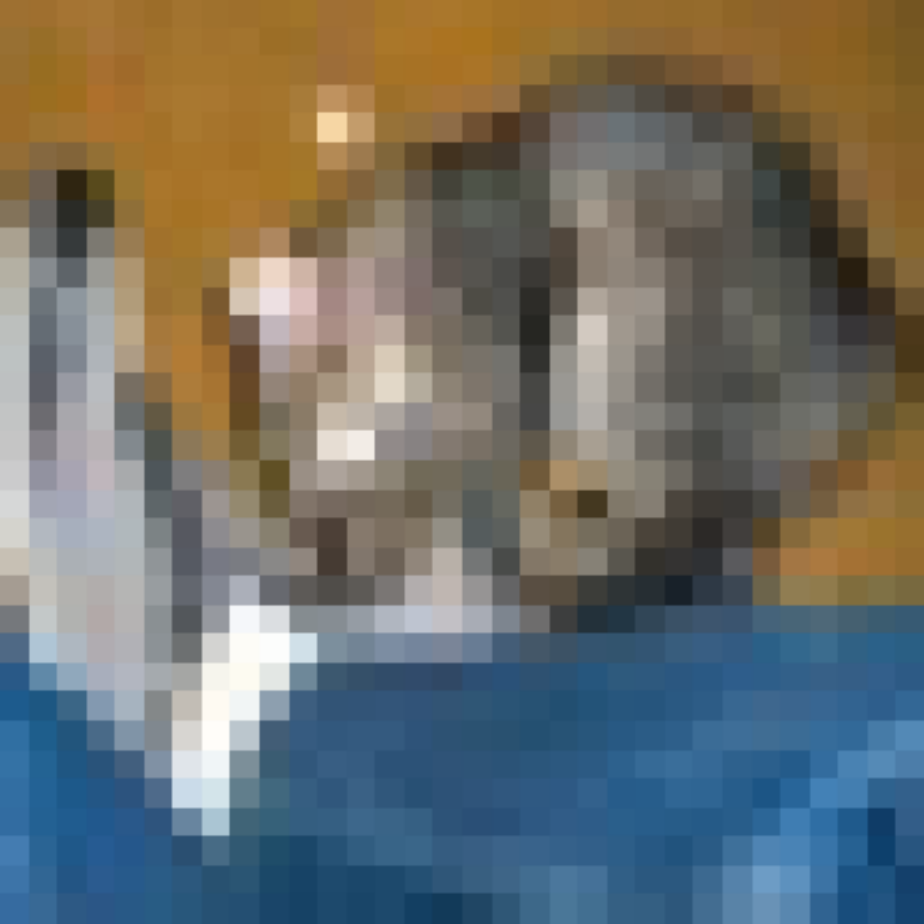}
    \includegraphics[width=0.1\linewidth]{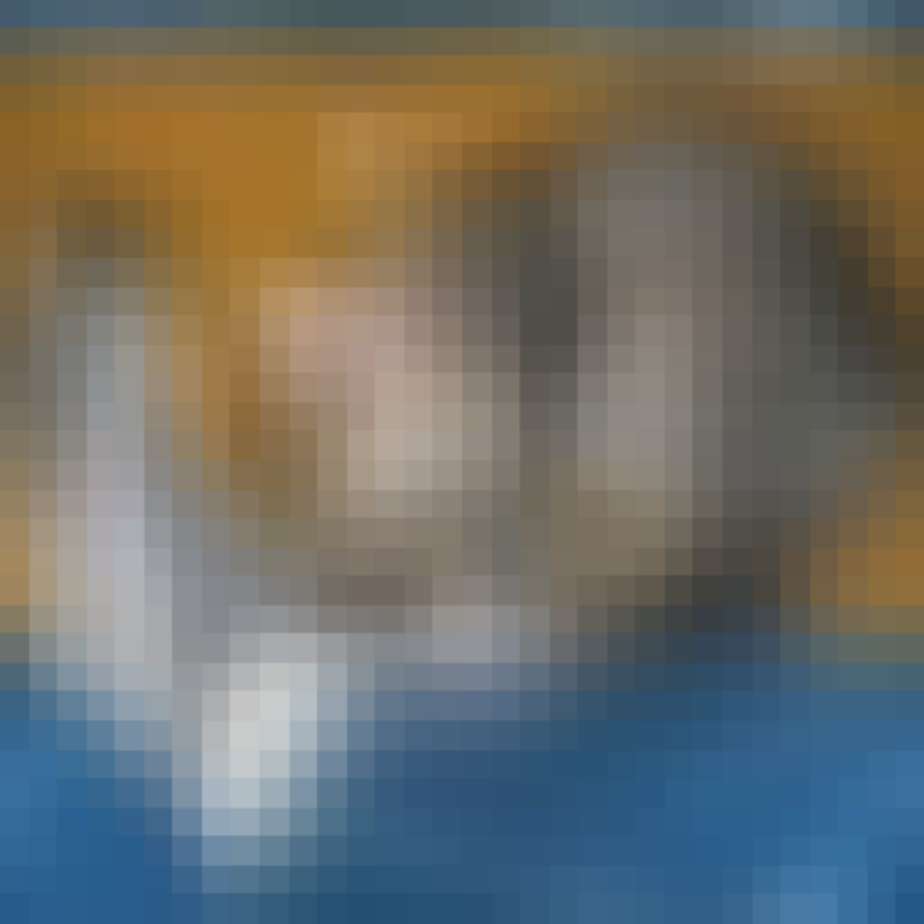}
    \includegraphics[width=0.1\linewidth]{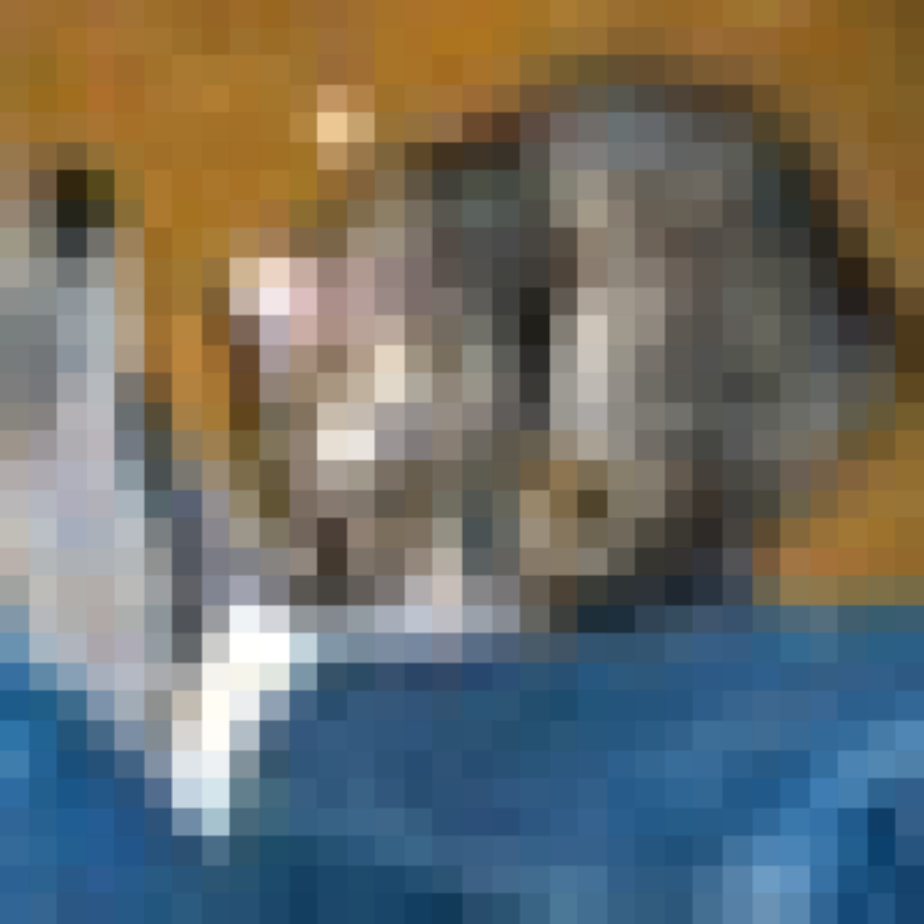}
    \includegraphics[width=0.1\linewidth]{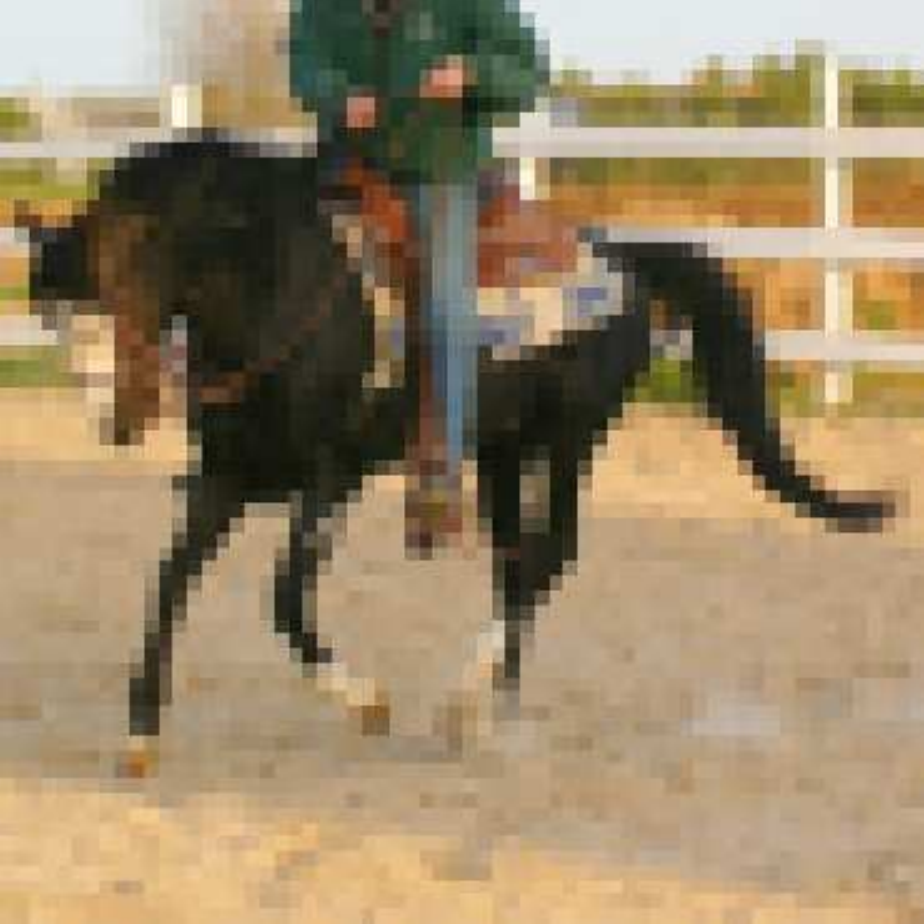}
    \includegraphics[width=0.1\linewidth]{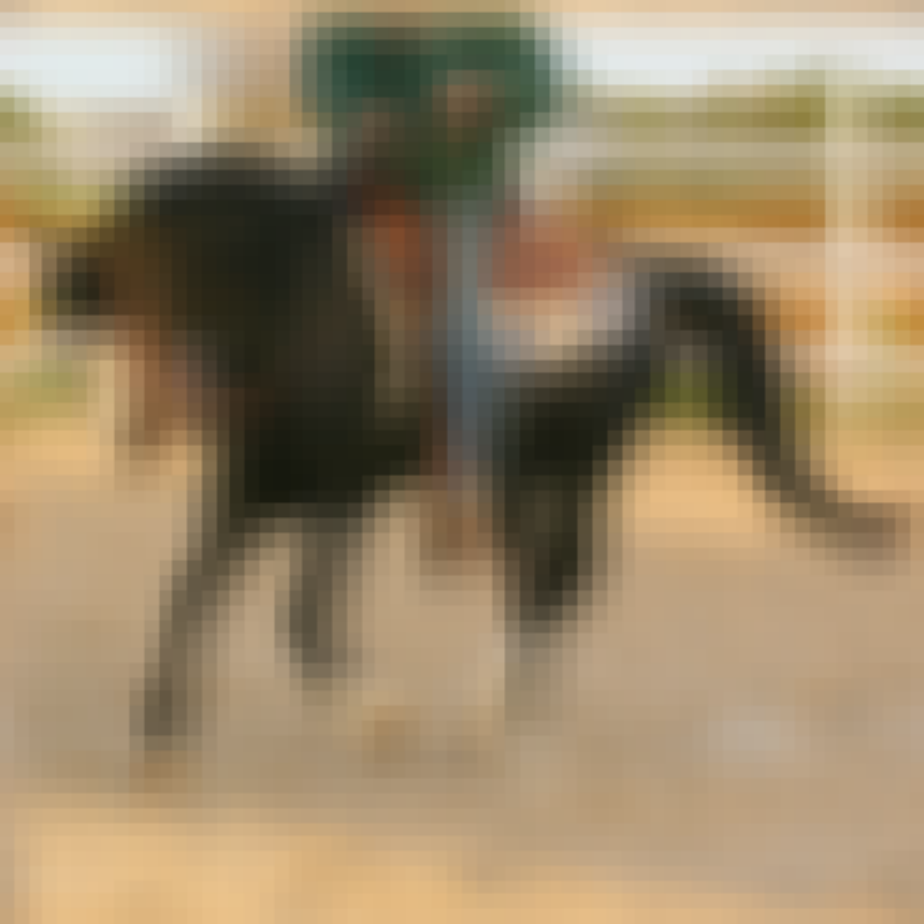}
    \includegraphics[width=0.1\linewidth]{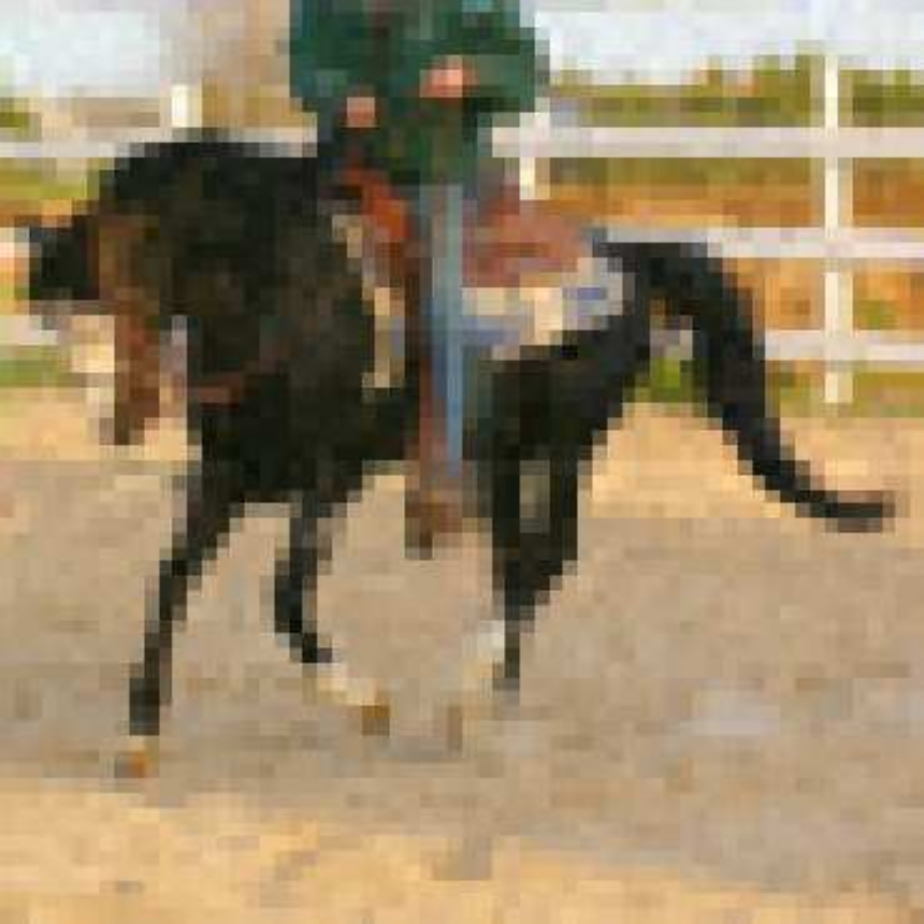}
    \\
    \includegraphics[width=0.1\linewidth]{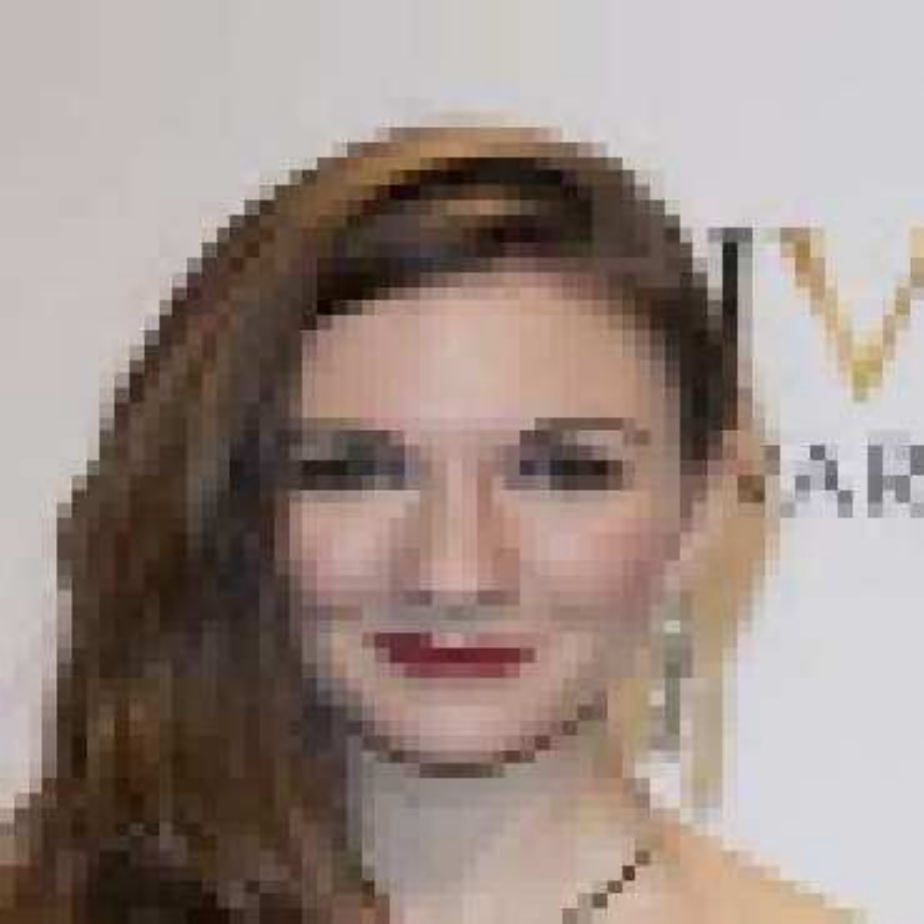}
    \includegraphics[width=0.1\linewidth]{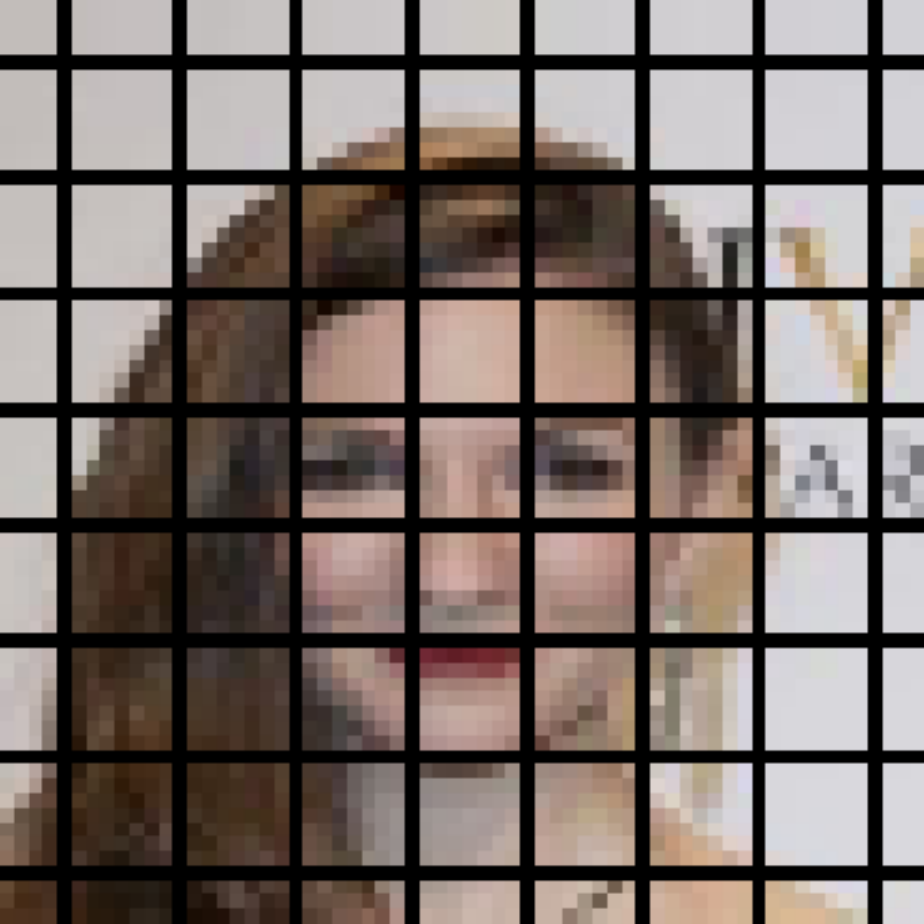}
    \includegraphics[width=0.1\linewidth]{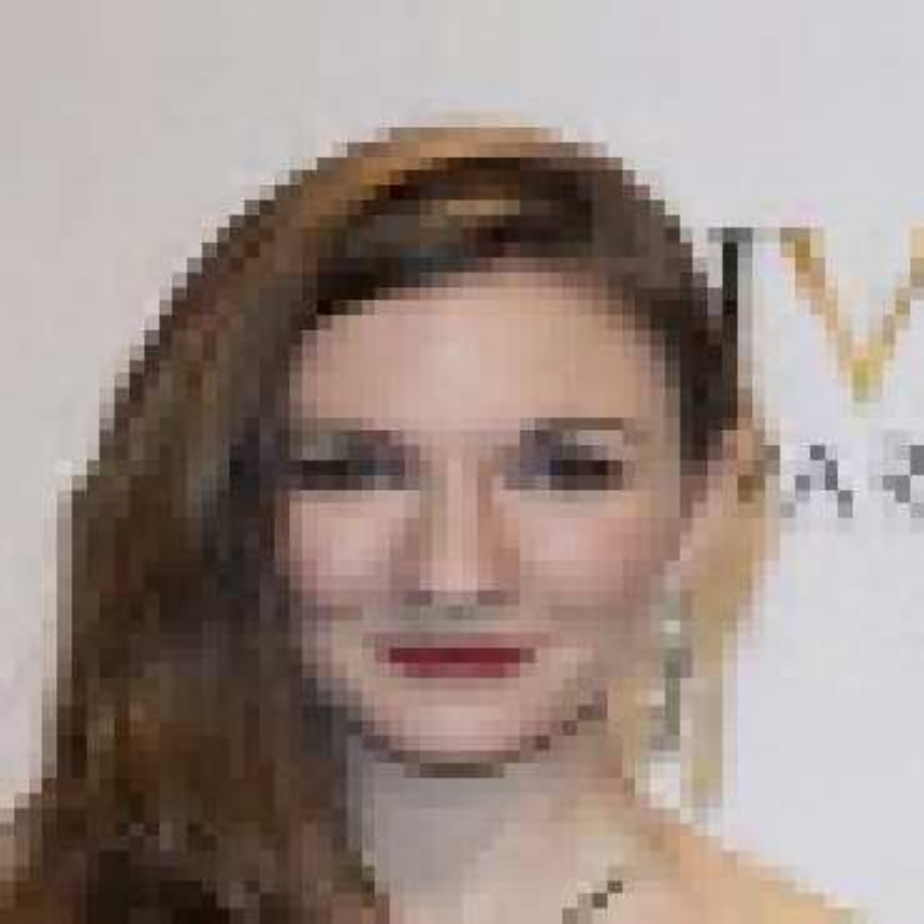}
    \includegraphics[width=0.1\linewidth]{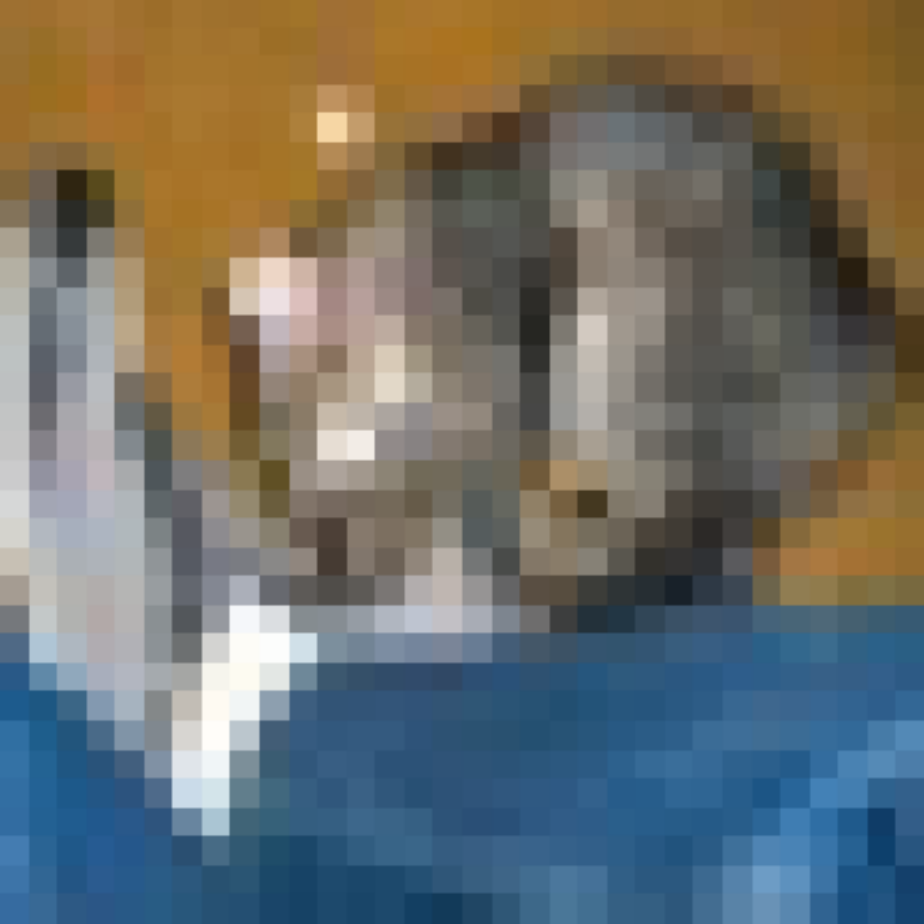}
    \includegraphics[width=0.1\linewidth]{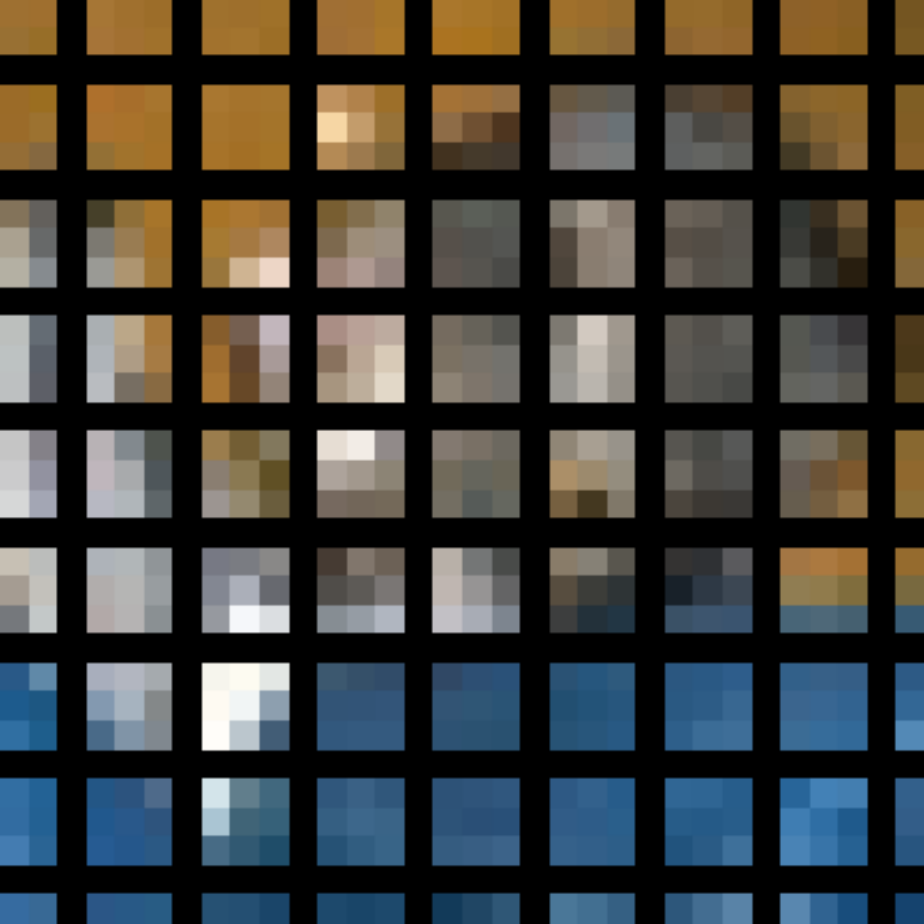}
    \includegraphics[width=0.1\linewidth]{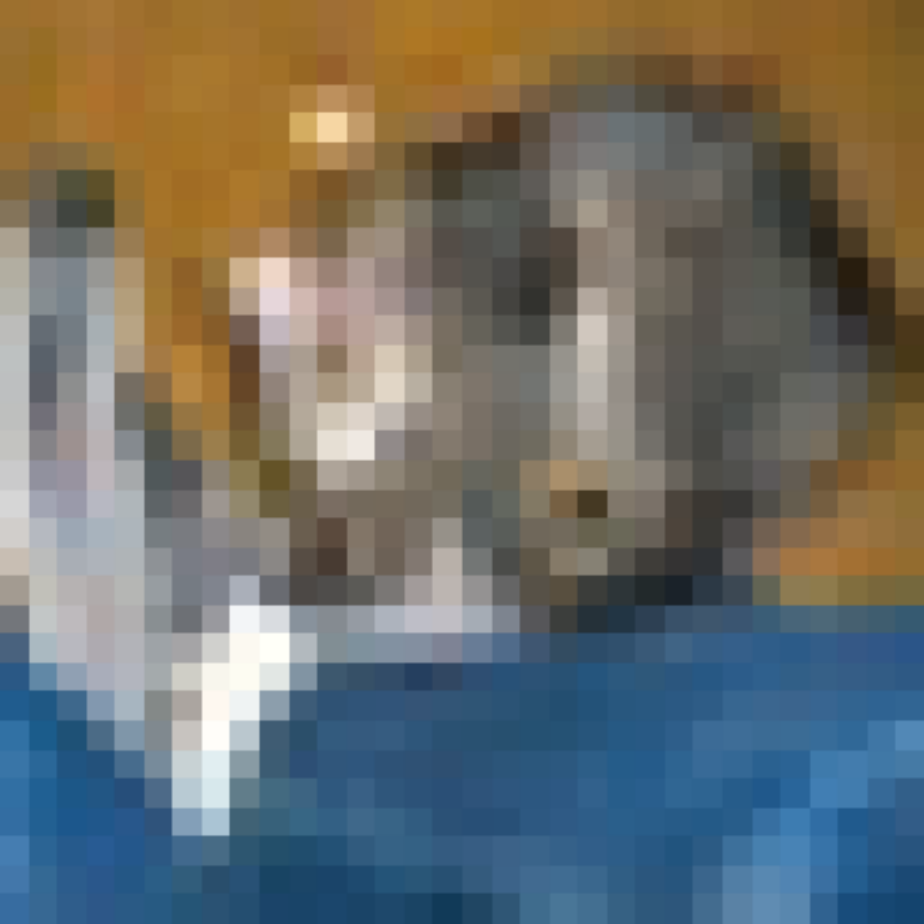}
    \includegraphics[width=0.1\linewidth]{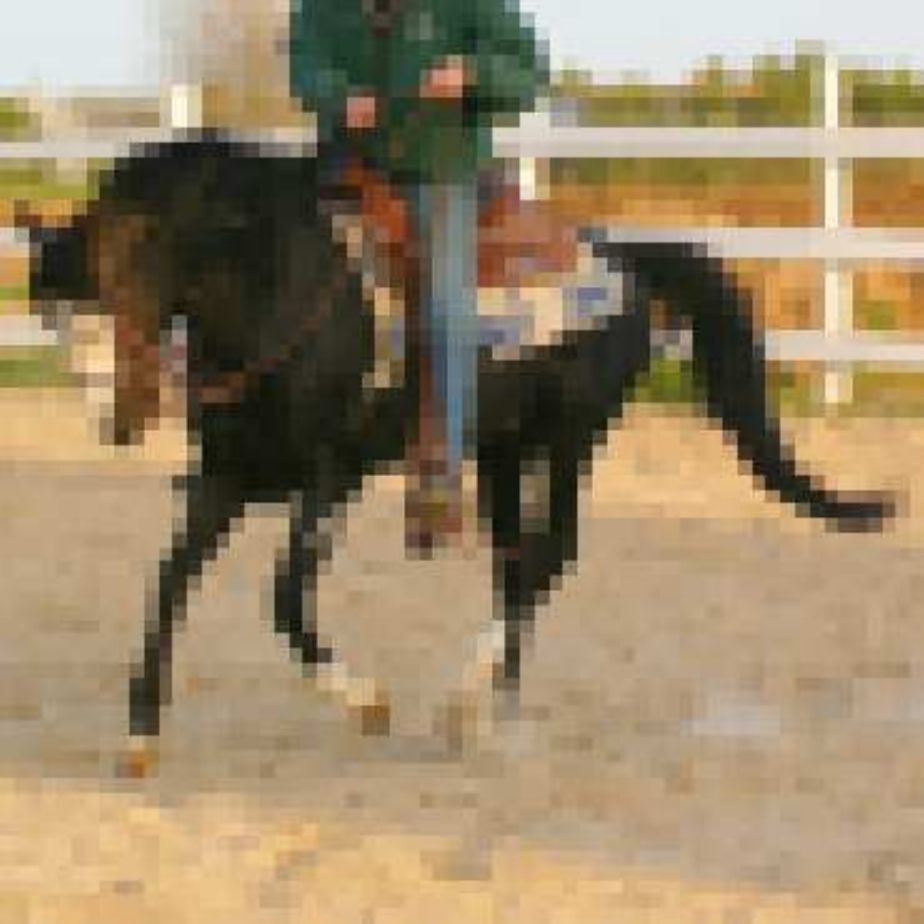}
    \includegraphics[width=0.1\linewidth]{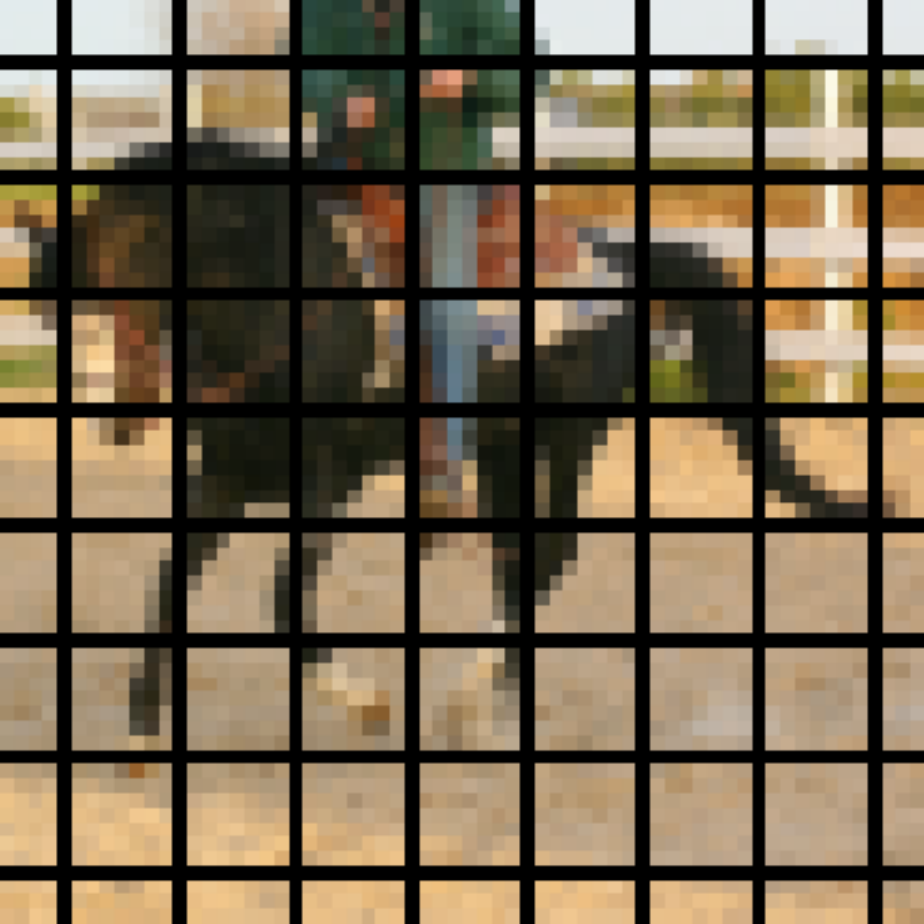}
    \includegraphics[width=0.1\linewidth]{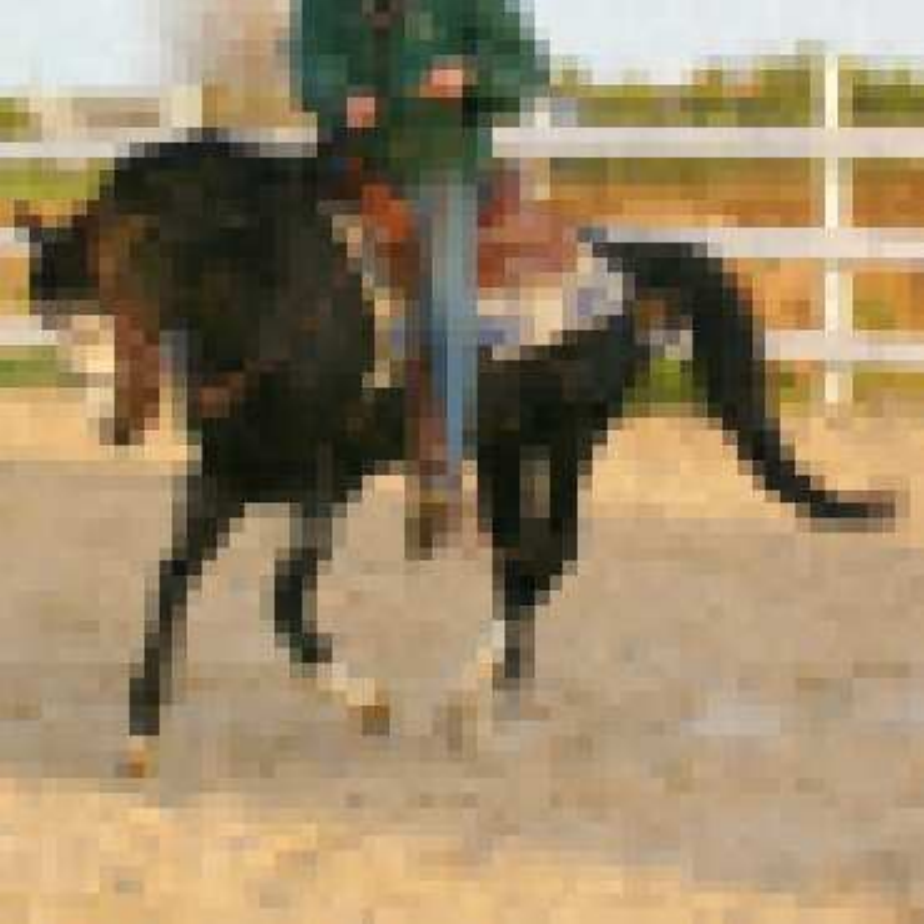}
    \end{minipage}
    } \caption{Original picture, distorted picture, and recovered picture
    by ButterflyNet2D\textsuperscript{2}. From top to bottom, the
    distortions are inpainting, denoising, deblurring, and watermark
    removal. From left to right, the datasets are CelebA, CIFAR10 and
    STL10.}
    \label{fig:image-processing-result}
\end{figure}

\begin{table}[htb]
    \centering
    \begin{tabular}{c c c c}
    \toprule
        Task & Dataset &
        \multicolumn{1}{c}{Preconditioned} &
        \multicolumn{1}{c}{Not Preconditioned} \\
        \toprule
        \multirow{3}*{Inpaint}
        & CelebA($64\times64$) &30.45 &31.06\\
        & CIFAR10($32\times32$) &28.40 &28.20\\
        & STL10($64\times64$) & 28.00&27.47\\
        \toprule
        \multirow{3}*{Deblur}& 
        CelebA($64\times64$) &33.79 &31.01\\
        & CIFAR10($32\times32$) &37.83 &36.55\\
        & STL10($64\times64$) & 30.66&29.43\\
        \bottomrule
    \end{tabular}
    \caption{The numerical results of Neumann network.}
    \label{tab:resNeumann}
\end{table}

Table~\ref{tab:resBF} and Figure~\ref{fig:image-processing-result}
illustrates all numerical results of ButterflyNet2D\textsuperscript{2}
applying to various tasks and datasets. According to
Table~\ref{tab:resBF}, the Fourier initialization outperforms both Kaiming
uniform random initialization and Kaiming normal random initialization. As
the complexity of the neural network increases, the training becomes more
difficult. The Fourier initialization bridges the classical image
processing method with the neural network methods. The network training
starts from the classical method and approaches the benefit of neural
networks. We make another comparison against Neumann network. In the
inpainting tasks, ButterflyNet2D\textsuperscript{2} and Neumann network
perform similarly. While in the deblurring tasks,
ButterflyNet2D\textsuperscript{2} with Fourier initialization outperforms
Neumann Network. This result is not surprising since deblurring is easier
in the frequency domain.

\section{Conclusions and Discussions}
\label{sec:conclusion}

In this paper, we proposed a neural network architecture named
ButterflyNet2D, together with a specially designed Fourier initialization.
The ButterflyNet2D with Fourier initialization approximates discrete
Fourier transforms with $O(N)$ parameters, where $N$ is the input size.
ButterflyNet2D and Fourier initialization allow us to bridge the classical
Fourier transformation method and powerful neural network methods for
image processing tasks. 

Through numerical experiments, we explored the approximation power of
ButterflyNet2D with Fourier initialization before and after training.
Numerical results show that ButterflyNet2D with Fourier initialization
well-approximates the Fourier transform and the training could further
improve the approximation accuracy. Tests of ill-posed image processing
tasks are also conducted. ButterflyNet2D shows its power in these tasks.

The work can be extended in several directions. Firstly, the
initialization method has limited versatility. The Fourier initialization
method heavily relies on the ButterflyNet2D architecture. Many popular
techiniques, e.g., max-pooling, batch normalization, or dropout, currently
cannot be incorporated with the Fourier initialization directly. Hence an
extension of Fourier initialization to incorporating these popular
techiniques is desired. Secondly, the ButterflyNet2D is slow in
backpropagation. The efficiency could be improved if ButterFlyNet2D is
better implemented as a building blocks in neural network frameworks.

\section{Appendix}

\subsection{Proof}
\label{app:proof}

When focusing on some square domain pair $A \subset [0,K)^2,B
\subset [0,1)^2$, the Fourier kernel can be decomposed as 
\begin{equation}
\begin{split}
    \calK(\xi,t) &= e^{-2\pi\imath (\xi \cdot t - \xi_0\cdot t - \xi\cdot t_0 + 
    \xi_0\cdot t_0)}
    e^{-2\pi\imath\xi_0\cdot t} e^{-2\pi\imath\xi\cdot t_0}
    e^{-2\pi\imath (-\xi_0\cdot t_0)}\\
    &= e^{-2\pi\imath R(\xi,t)}e^{-2\pi\imath\xi_0\cdot t}e^{-2\pi\imath\xi\cdot t_0}
    e^{2\pi\imath \xi_0\cdot t_0},
\end{split}
\end{equation}
where $R(\xi,t) = (\xi-\xi_0)\cdot(t-t_0)$, $\xi_0$ is the center of $A$,
and $t_0$ is the center of $B$.

For any fixed $\xi$, we have
\begin{equation}
    e^{-2\pi\imath R(\xi,t)} = \sum_{k = 0}^{\infty}\frac{(-2\pi\imath R(\xi,t))^k}{k!}.
\end{equation}
If we have
$\omega(A)\omega(B) < \displaystyle\frac{r^2}{e\pi}$, the $r^2$-term truncation error
can be bounded as
\begin{equation}
\begin{split}
    & \left| e^{-2\pi \imath R(\xi,t)} - 
    \sum_{k = 0}^{r^2-1}\frac{(-2\pi\imath R(\xi,t))^k}{k!} \right|
    =\left|\sum_{k = r^2}^{\infty}\frac{(-2\pi\imath R(\xi,t))^k}{k!}\right|\\
    &=\left|\sum_{k = r^2}^{\infty}\frac{(-\pi\imath \omega(A)\omega(B))^k}{k!}\right|
    \leq\sum_{k = r^2}^{\infty}\frac{(\pi \omega(A)\omega(B))^k}{k!}\\
    &\leq\sum_{k = r^2}^{\infty}\frac{(e\pi \omega(A)\omega(B))^k}{k^k}
    \leq\sum_{k = r^2}^{\infty}\frac{(e\pi \omega(A)\omega(B))^k}{r^{2k}}
    =\frac{\frac{(e\pi \omega(A)\omega(B))^{r^2}}{r^{2r^2}}}{1-\frac{e\pi \omega(A)\omega(B)}{r^{2}}}\\
    &=\frac{\gamma^{r^2}}{1-\gamma},
\end{split}
\end{equation}
here we use $\gamma$ to denote $\displaystyle\frac{e\pi \omega(A)\omega(B)}{r^{2}}$.

Notice that $\displaystyle\sum_{k = 0}^{r^2-1}\frac{(-2\pi\imath R(\xi,t))^k}{k!}$ is a
polynomial about $t$, which means we have 
\begin{equation}
   \left\|e^{-2\pi\imath R(\xi,t)}-\sum_{k_x=1}^r\sum_{k_y=1}^r e^{-2\pi \imath R(\xi,t_{k_{x}, k_{y}})}
   \calL_{k_x,k_y }(t)\right\|_{\infty} < C\frac{\gamma^{r^2}}{1-\gamma},
\end{equation}
i.e.
\begin{equation}
\begin{split}
   \left\|e^{-2\pi\imath (\xi-\xi_0)\cdot(t-t_0)}-\sum_{k_x=1}^r\sum_{k_y=1}^r e^{-2\pi \imath
   (\xi-\xi_0)\cdot(t_{k_x,k_y }-t_0)}\calL_{k_x, k_y }(t)\right\|_{\infty} \leq C\frac{\gamma^{r^2}}{1-\gamma},\\
   \Longrightarrow 
   \sup_{\xi\in A, t\in B}\left|e^{-2\pi\imath\xi\cdot t}-\sum_{k_x=1}^r\sum_{k_y=1}^r e^{-2\pi\imath \xi\cdot t_{k_x,k_y}}
   e^{-2\pi\imath \xi_0\cdot(t-t_{k_x, k_y })}\calL_{k_x, k_y }(t)\right| \leq 
   C\frac{\gamma^{r^2}}{1-\gamma},
\end{split}
\end{equation}
where $C$ is a constant independent of $t$ and $\xi$, $t_{k_x, k_y}$ are used to
denote the $r^2$ Chebyshev nodes in $B$.

Similarly, for any fixed $t\in B$, we have 
$\displaystyle\sum_{k=0}^{r^2-1}\frac{(-2\pi\imath R(\xi,t))^k}{k!}$ is a polynomial 
about $\xi$. Hence the second conclusion can be obtained through the same procedure.

\subsection{Complex Valued Network}
\label{app:complexNN}

In order to realize complex number multiplication and addition via
nonlinear neural network, we represent a complex number as four real
numbers, i.e., a complex number $x = \Re x + \imath \Im x$ $\in \bbC$ is
represented as 
\begin{equation}
    \begin{bmatrix}
    (\Re x)_{+} & (\Im x)_{+} & (\Re x)_{-} & (\Im x)_{-}
    \end{bmatrix}^{\top},
\end{equation}
where $(z)_+ = \max(z, 0)$, $(z)_- = \max(-z, 0)$ for any $z \in \bbR$.
Then a complex-scalar multiplication 
\begin{equation}
    ax = y.
\end{equation}
can be represented as
\begin{equation}
    \sigma\left(
    \begin{bmatrix}
        \Re a & -\Im a & -\Re a & \Im a \\
        \Im a & \Re a & -\Im a & -\Re a \\
        -\Re a & \Im a & \Re a & -\Im a \\
        -\Im a & -\Re a & \Im a & \Re a
    \end{bmatrix}
    \begin{bmatrix}
        (\Re x)_+ \\
        (\Im x)_+ \\
        (\Re x)_- \\
        (\Im x)_-
    \end{bmatrix}
    \right)
    =
    \begin{bmatrix}
        (\Re y)_+ \\
        (\Im y)_+ \\
        (\Re y)_- \\
        (\Im y)_-
    \end{bmatrix}.
\end{equation}
Here $\sigma$ is the activation function called ReLU.

\subsection{Approximation to Fourier Transform with Random Initialization}
\label{app:ftapp-rand-train}

The training accuracy of ButterflyNet2D with random initialization to
approximate the Fourier and invese Fourier transform are included in
Table~\ref{tab:ftapp-rand-train} and Table~\ref{tab:iftapp-rand-train}
respectively.

\begin{table}[htb]
\centering
\begin{tabular}{c c c c c}
    \multicolumn{5}{c}{$N = $ $64\times64$(FT)} \\
    \toprule
    & \multicolumn{2}{c}{layer $L$ (with $r=2$)} & \multicolumn{2}{c}{Cheb (with $L = 6$)} \\ 
    \cmidrule(lr){2-3}
    \cmidrule(lr){4-5}
    & $3$ & $4$ & $2^2$ & $3^2$ \\ 
    \toprule
    $\epsilon_1$&
    $9.71\times10^{-1}$ & 
    $9.71\times10^{-1}$ & 
    $9.92\times10^{-1}$ & 
    $9.71\times10^{-1}$ \\ 
    $\epsilon_2$&
    $5.06\times10^{-1}$ & 
    $5.06\times10^{-1}$ & 
    $7.95\times10^{-1}$ & 
    $5.06\times10^{-1}$ \\ 
    $\epsilon_{\infty}$&
    $2.61\times10^{-1}$ & 
    $2.61\times10^{-1}$ & 
    $7.11\times10^{-1}$ & 
    $2.62\times10^{-2}$ \\
    \bottomrule
\end{tabular}

\caption{Relative errors of the network approximating the Fourier
Fourier operator after training.}
\label{tab:ftapp-rand-train}
\end{table}

\begin{table}[htb]
\centering
\begin{tabular}{c c c c c}
    \multicolumn{5}{c}{$N = $ $64\times64$(IFT)} \\
    \toprule
    & \multicolumn{2}{c}{layer $L$ (with $r=2$)} & \multicolumn{2}{c}{Cheb (with $L = 6$)} \\ 
    \cmidrule(lr){2-3}
    \cmidrule(lr){4-5}
    & $3$ & $4$ & $2^2$ & $3^2$ \\ 
    \toprule
    $\epsilon_1$&
    $6.27\times10^{-1}$ & 
    $6.30\times10^{-1}$ & 
    $6.38\times10^{-1}$ & 
    $6.43\times10^{-1}$ \\ 
    $\epsilon_2$&
    $6.52\times10^{-1}$ & 
    $6.53\times10^{-1}$ & 
    $6.64\times10^{-1}$ & 
    $6.66\times10^{-1}$ \\
    $\epsilon_{\infty}$&
    $9.75\times10^{-1}$ & 
    $9.78\times10^{-1}$ & 
    $9.87\times10^{-1}$ & 
    $9.94\times10^{-1}$ \\
    \bottomrule
\end{tabular}

\caption{Relative errors of the network approximating the inverse
Fourier operator after training.}
\label{tab:iftapp-rand-train}
\end{table}

\subsection{Loss Curve}
In this section, we offer some images that describe how the loss drops.

\begin{figure}[htb]
    \centering
    \subfigure{
    \begin{minipage}{0.5\linewidth}
    \centering
    \includegraphics[width=0.9\linewidth]{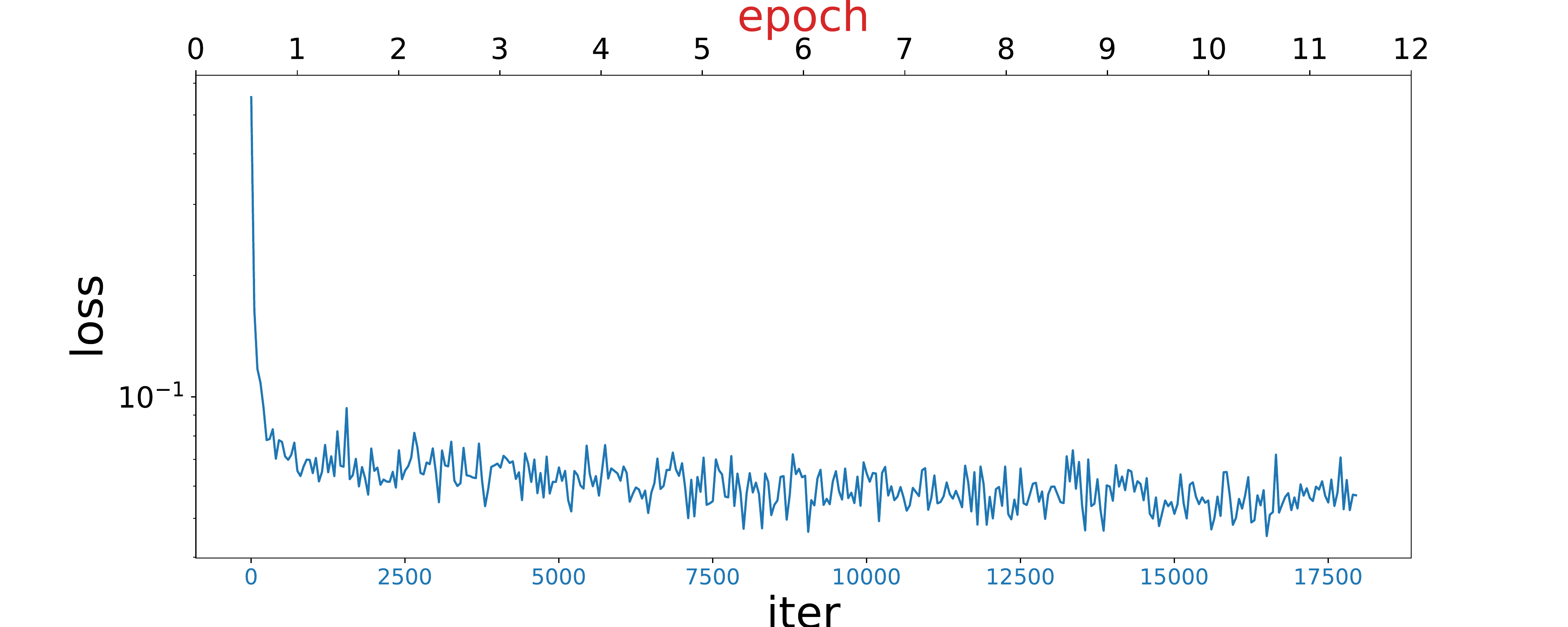}
    \end{minipage}
    \begin{minipage}{0.5\linewidth}
    \centering
    \includegraphics[width=0.9\linewidth]{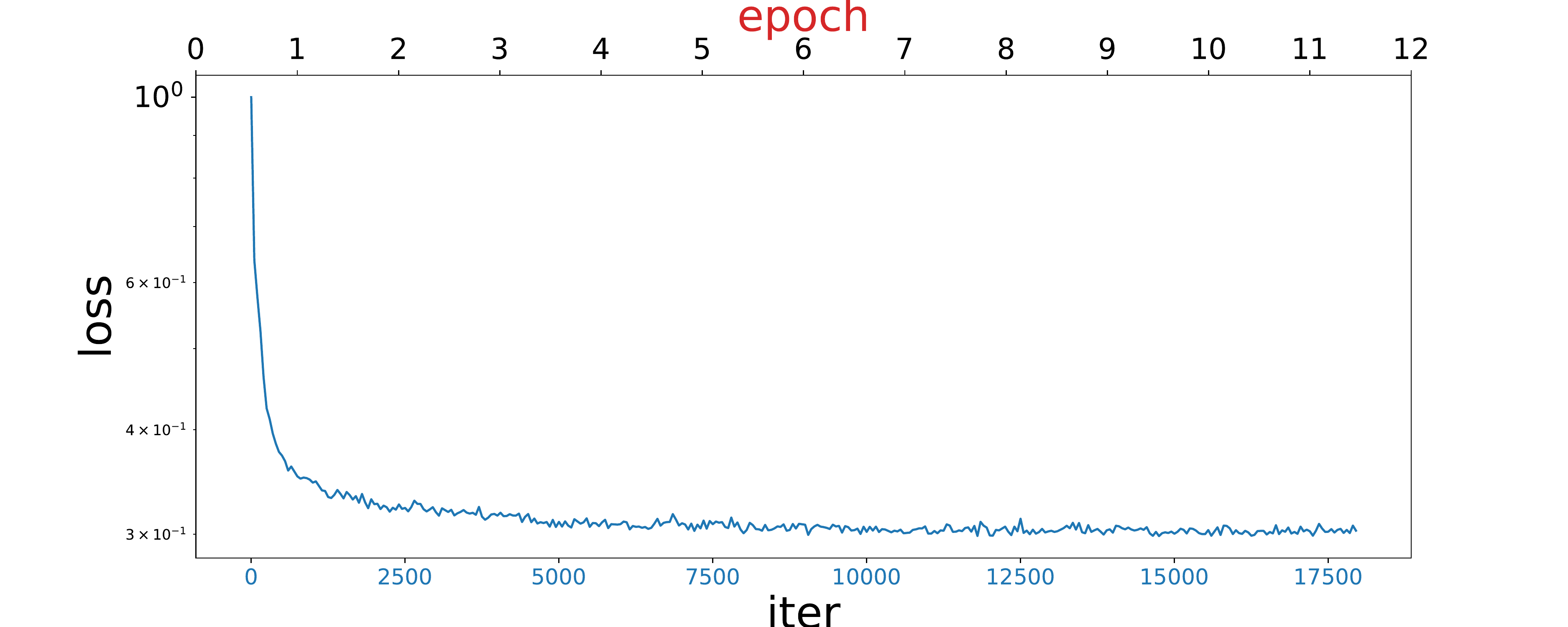}
    \end{minipage}
    }
    \subfigure{
    \begin{minipage}{0.5\linewidth}
    \centering
    \includegraphics[width=0.9\linewidth]{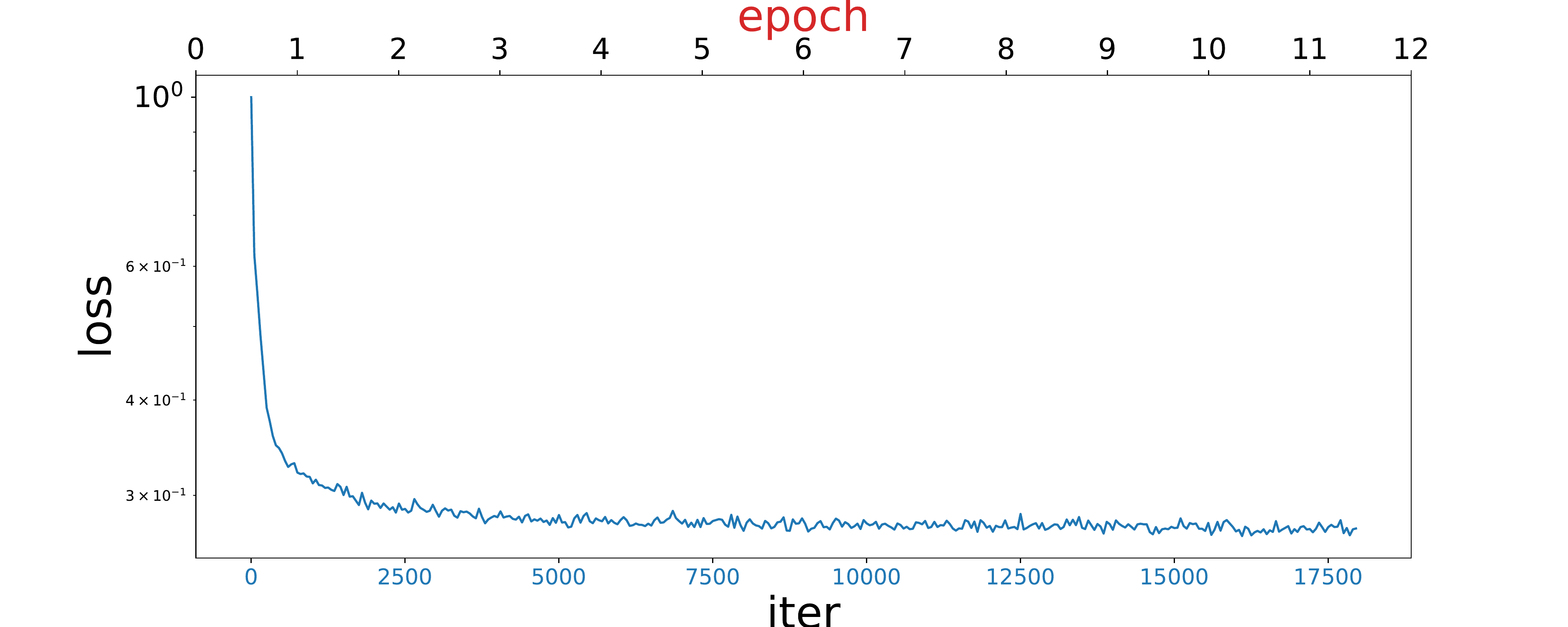}
    \end{minipage}
    \begin{minipage}{0.5\linewidth}
    \centering
    \includegraphics[width=0.9\linewidth]{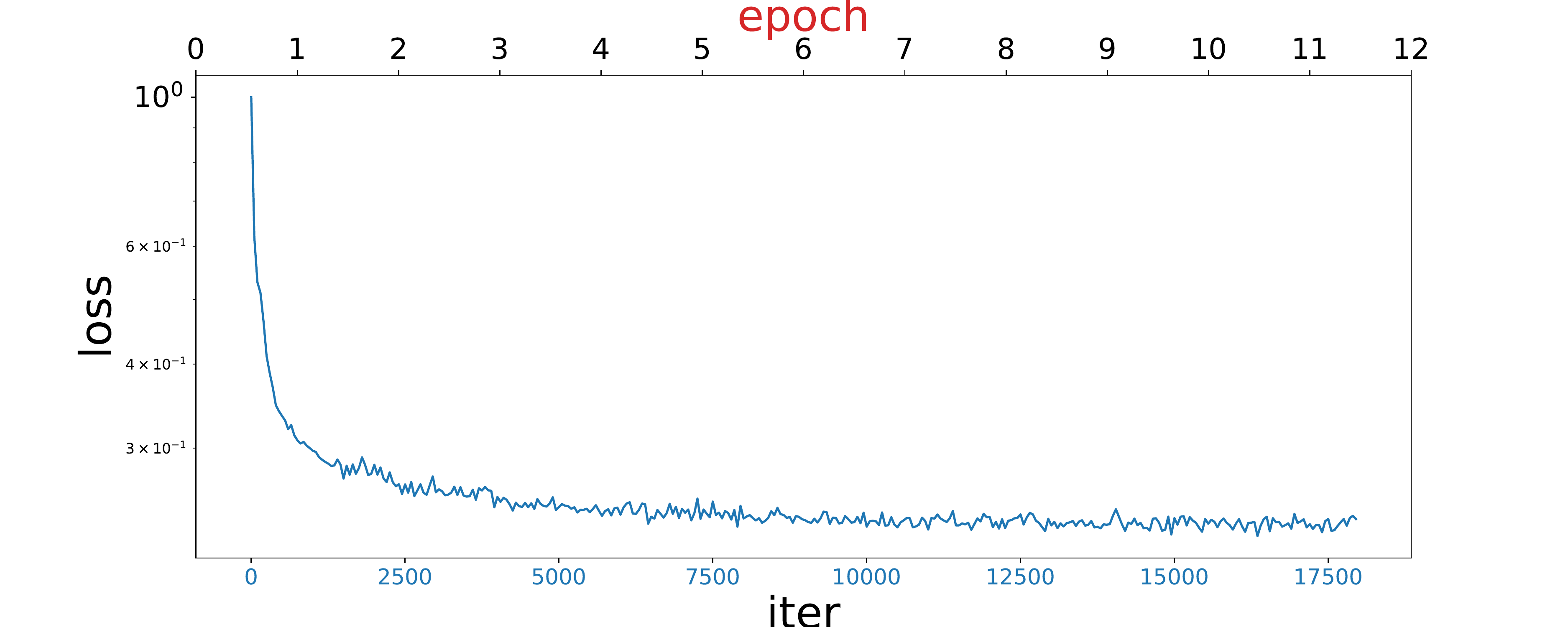}
    \end{minipage}
    }
    \caption{Loss drop in task Inpainting, dataset CelebA. These corresponds to four
    different initializations used in our experiments, namely Fourier,
    Kaiming Uniform, Kaiming Normal and Orthogonal.}
\end{figure}

\begin{figure}[htb]
    \centering
    \subfigure{
    \begin{minipage}{0.5\linewidth}
    \centering
    \includegraphics[width=0.9\linewidth]{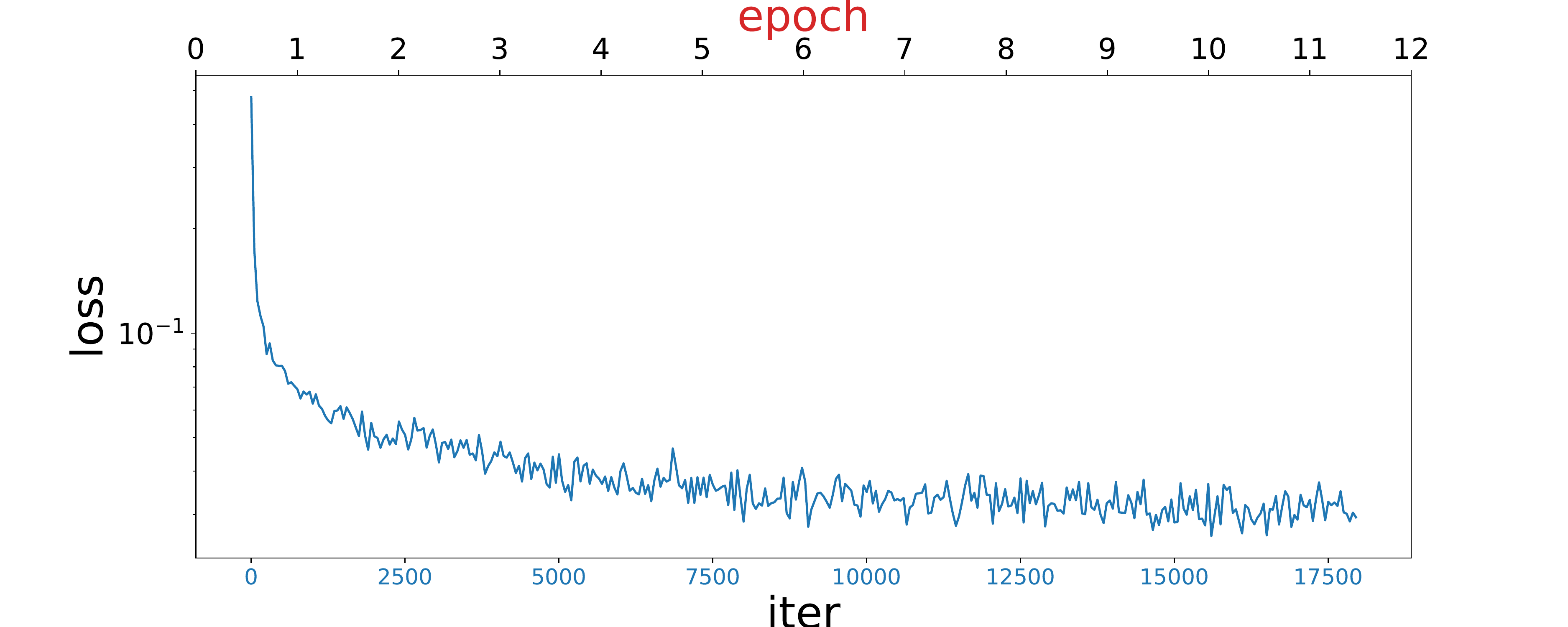}
    \end{minipage}
    \begin{minipage}{0.5\linewidth}
    \centering
    \includegraphics[width=0.9\linewidth]{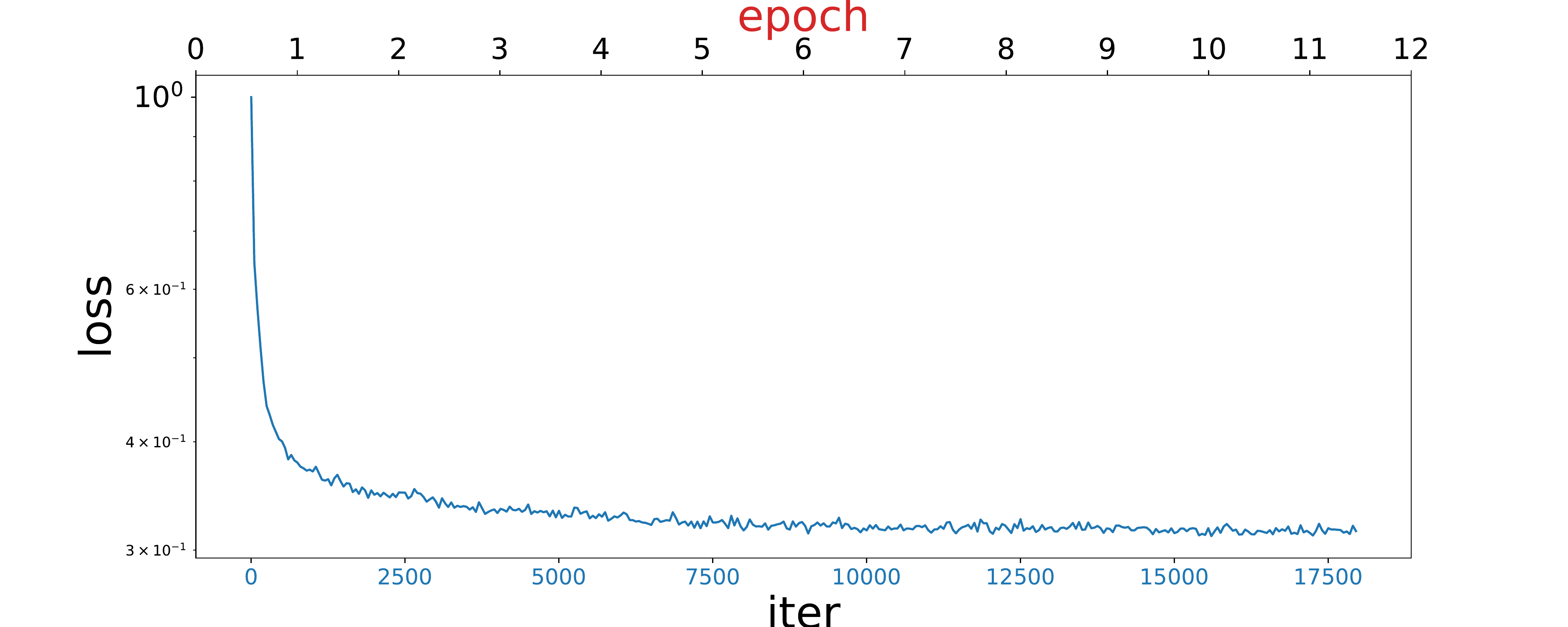}
    \end{minipage}
    }
    \subfigure{
    \begin{minipage}{0.5\linewidth}
    \centering
    \includegraphics[width=0.9\linewidth]{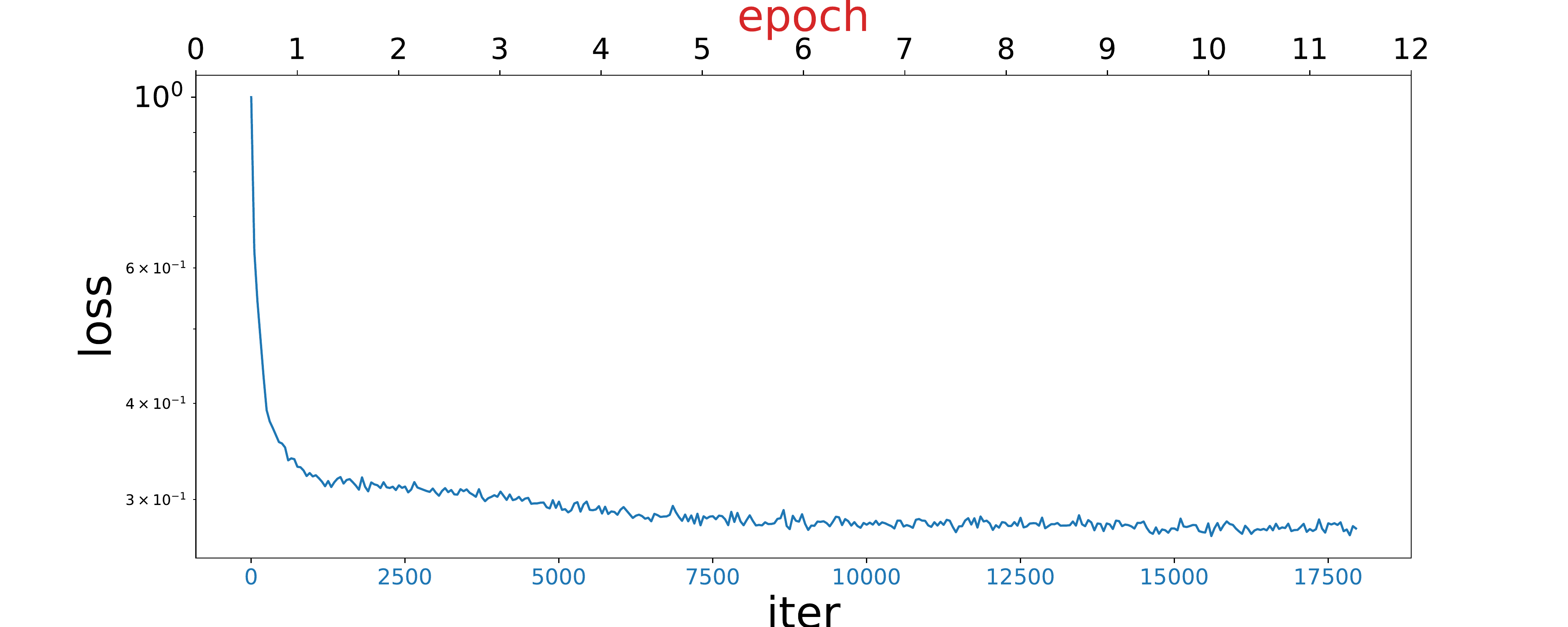}
    \end{minipage}
    \begin{minipage}{0.5\linewidth}
    \centering
    \includegraphics[width=0.9\linewidth]{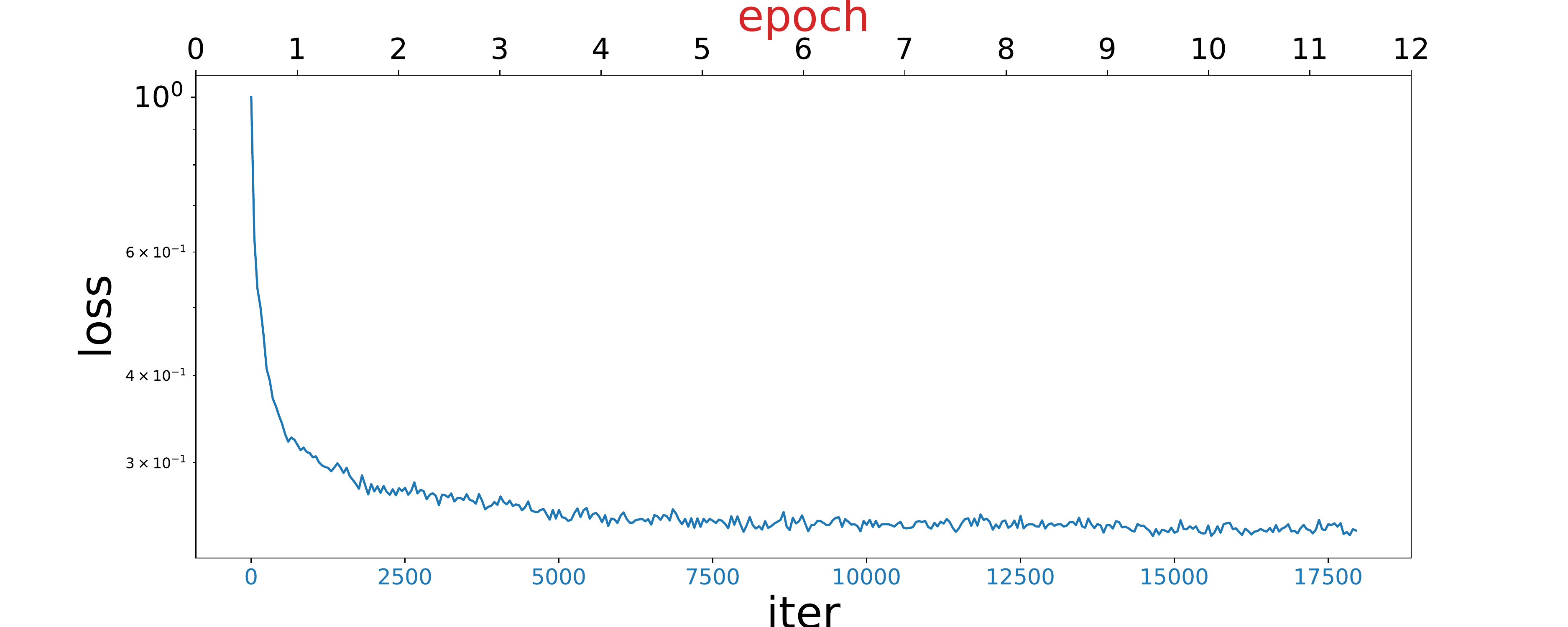}
    \end{minipage}
    }
    \caption{Loss drop in task Deblurring, dataset CelebA. These corresponds to four
    different initializations used in our experiments, namely Fourier,
    Kaiming Uniform, Kaiming Normal and Orthogonal.}
\end{figure}

\begin{figure}[htb]
    \centering
    \subfigure{
    \begin{minipage}{0.5\linewidth}
    \centering
    \includegraphics[width=0.9\linewidth]{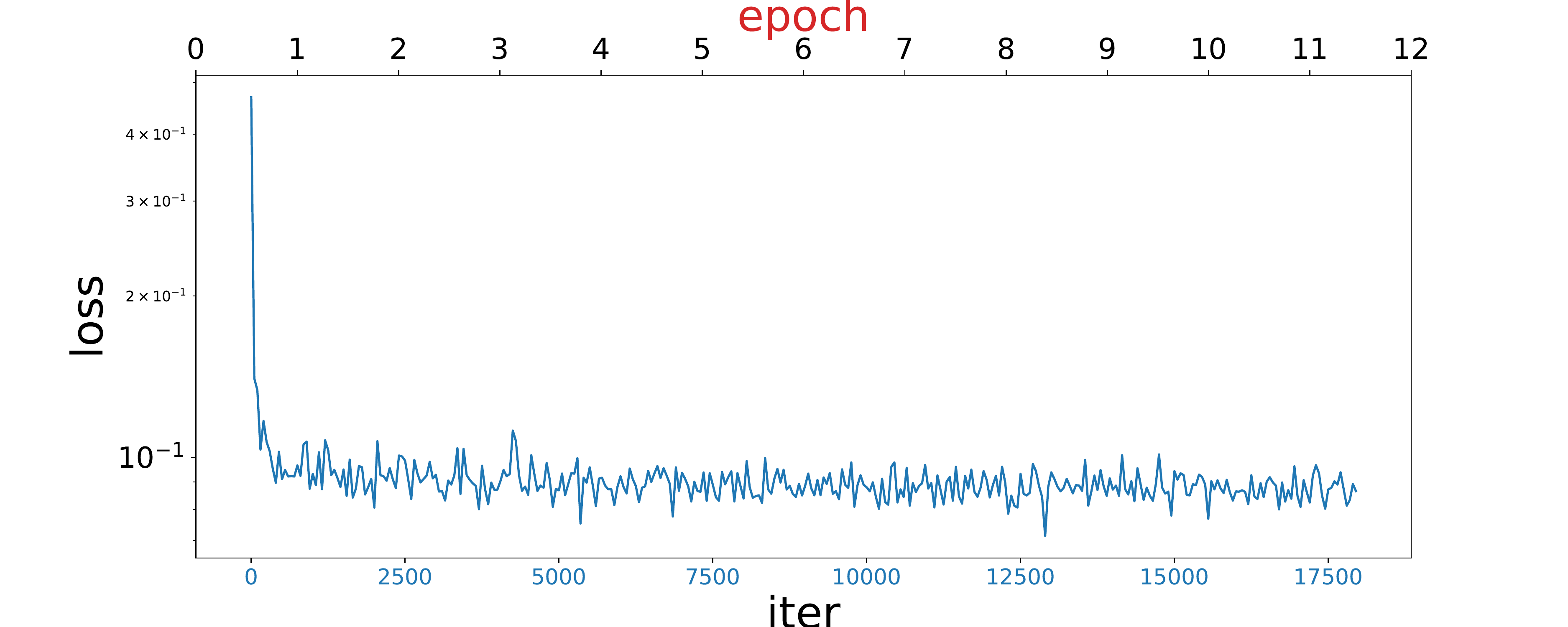}
    \end{minipage}
    \begin{minipage}{0.5\linewidth}
    \centering
    \includegraphics[width=0.9\linewidth]{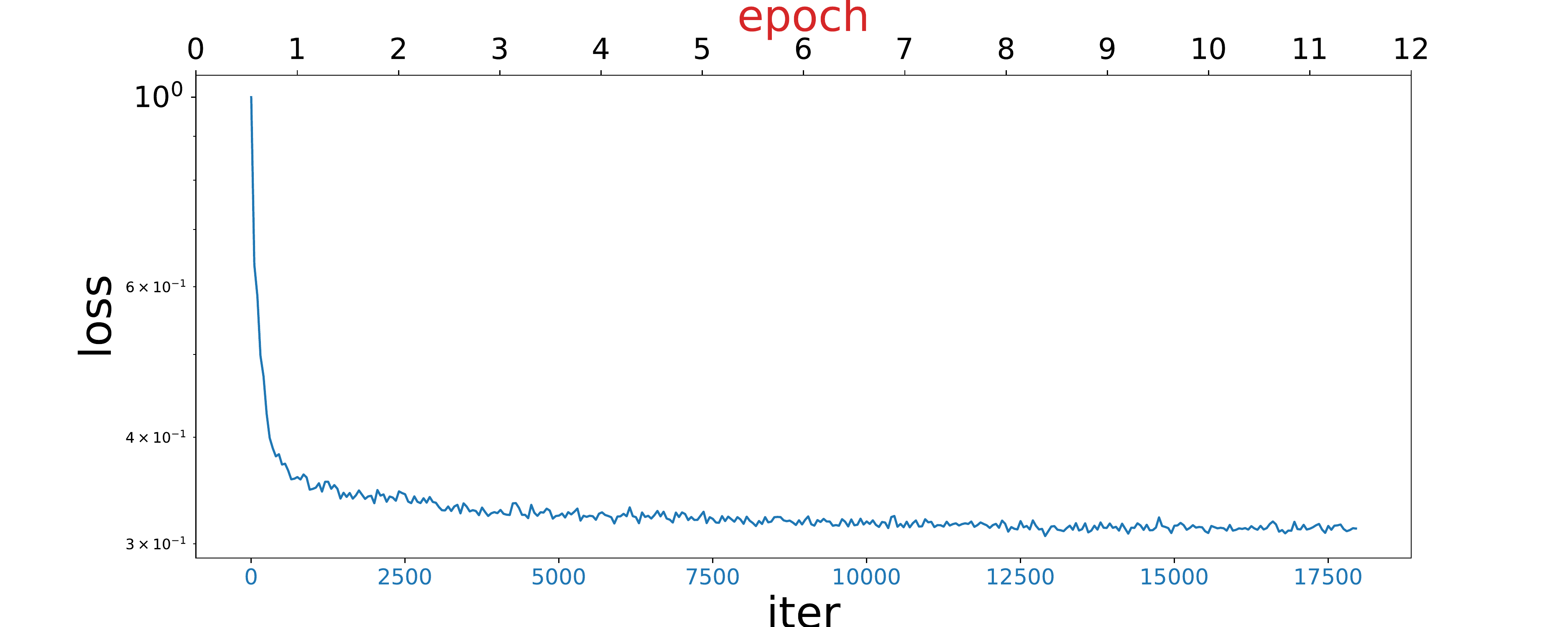}
    \end{minipage}
    }
    \subfigure{
    \begin{minipage}{0.5\linewidth}
    \centering
    \includegraphics[width=0.9\linewidth]{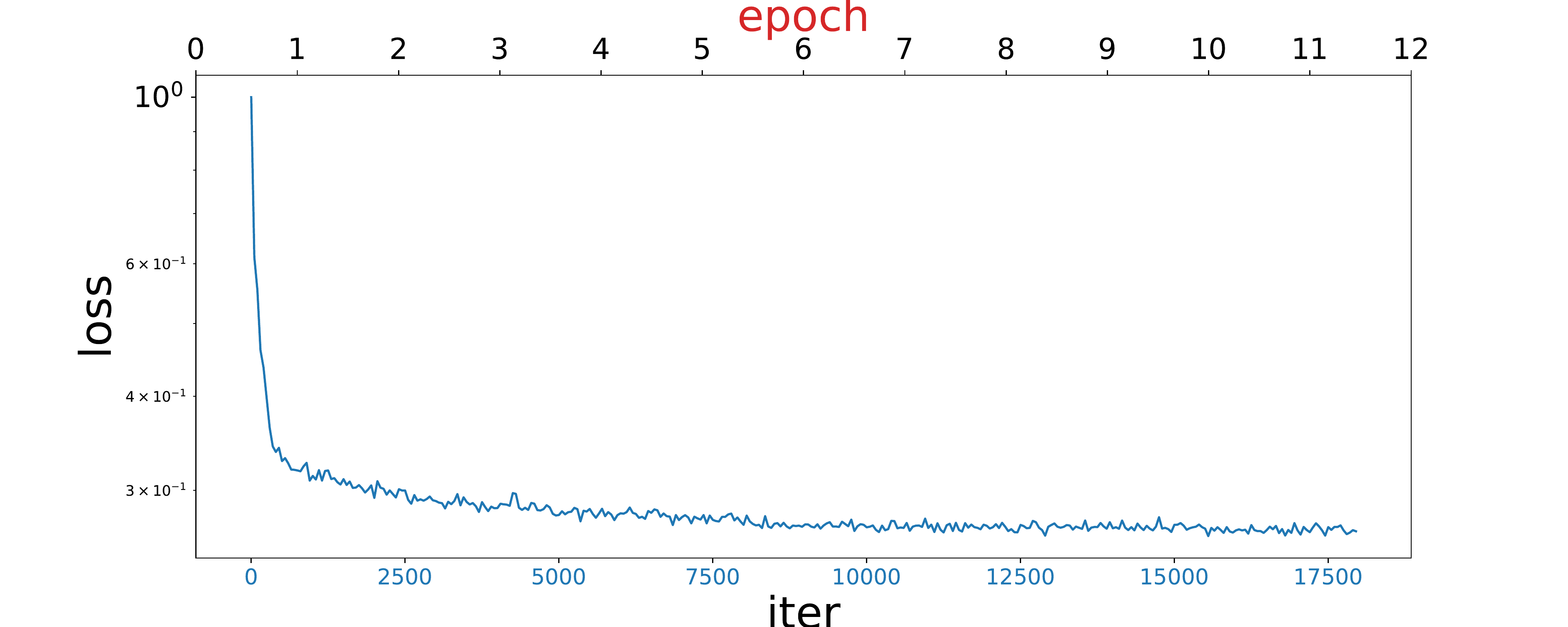}
    \end{minipage}
    \begin{minipage}{0.5\linewidth}
    \centering
    \includegraphics[width=0.9\linewidth]{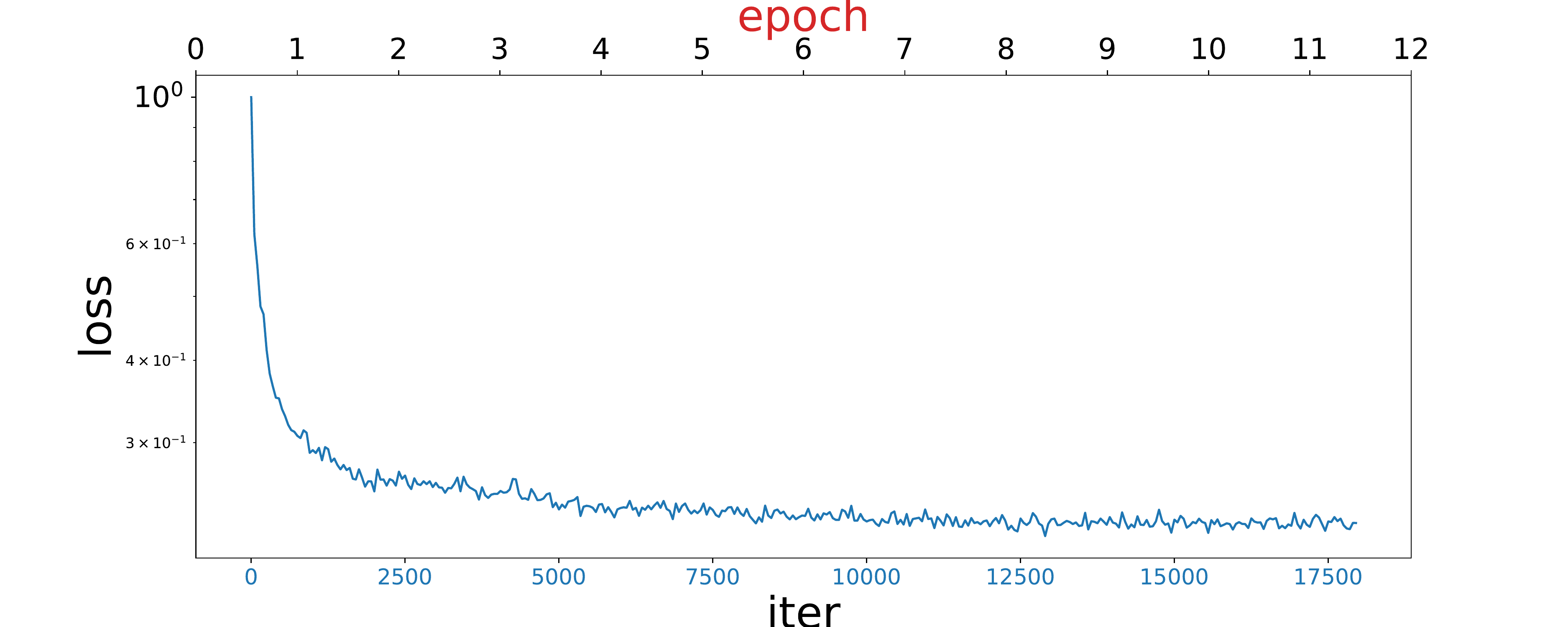}
    \end{minipage}
    }
    \caption{Loss drop in task Denoising, dataset CelebA. These corresponds to four
    different initializations used in our experiments, namely Fourier,
    Kaiming Uniform, Kaiming Normal and Orthogonal.}
\end{figure}

\begin{figure}[htb]
    \centering
    \subfigure{
    \begin{minipage}{0.5\linewidth}
    \centering
    \includegraphics[width=0.9\linewidth]{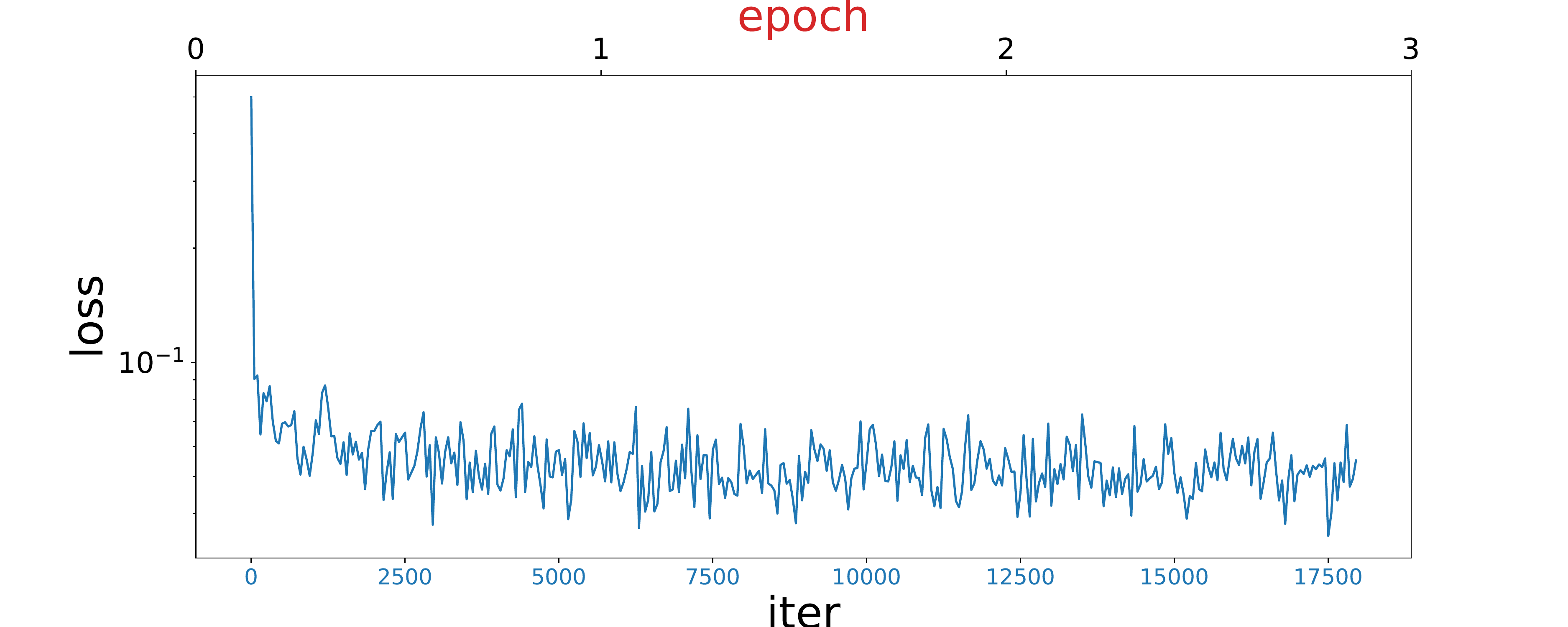}
    \end{minipage}
    \begin{minipage}{0.5\linewidth}
    \centering
    \includegraphics[width=0.9\linewidth]{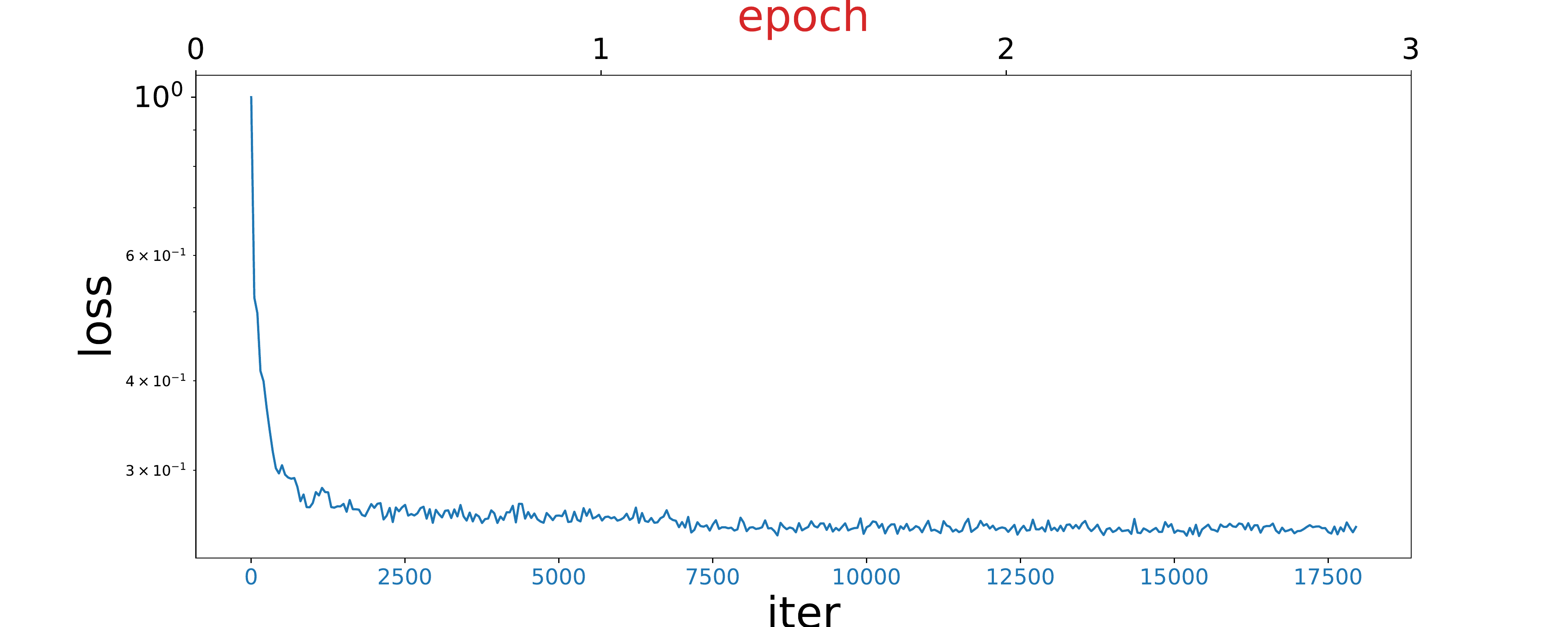}
    \end{minipage}
    }
    \subfigure{
    \begin{minipage}{0.5\linewidth}
    \centering
    \includegraphics[width=0.9\linewidth]{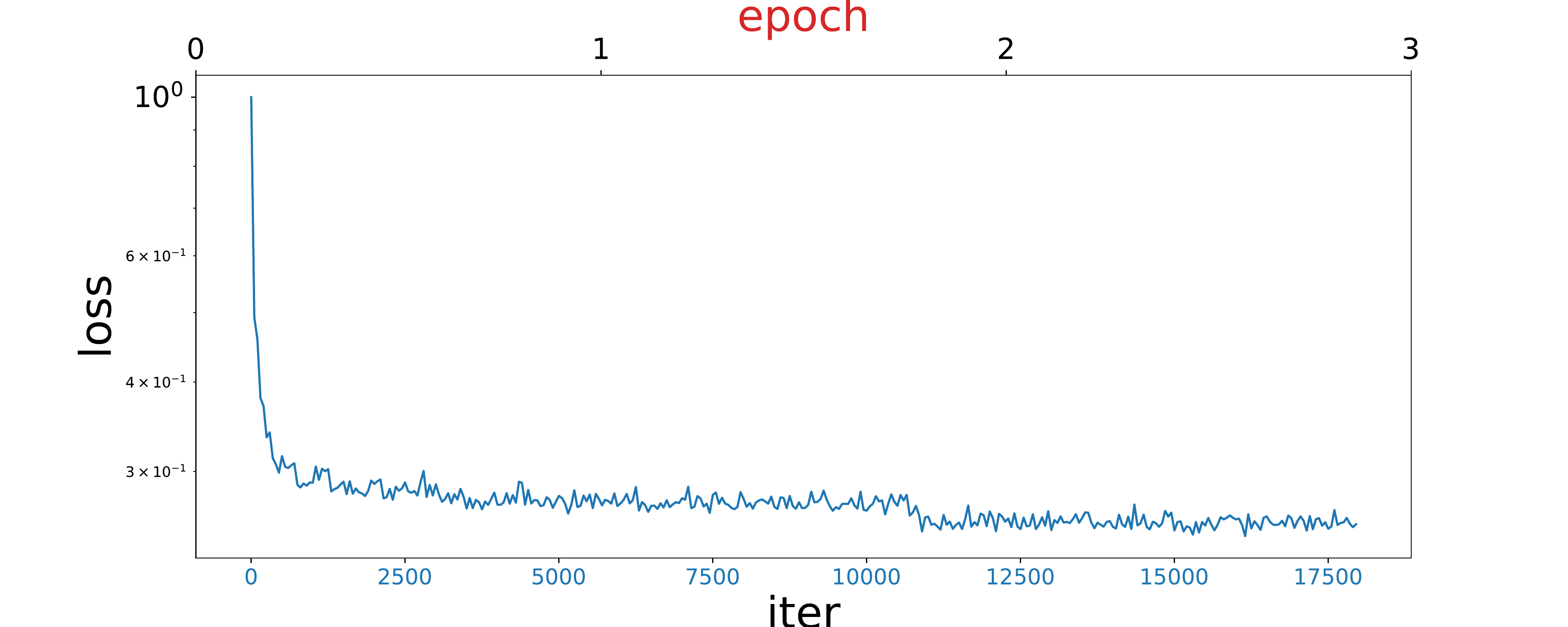}
    \end{minipage}
    \begin{minipage}{0.5\linewidth}
    \centering
    \includegraphics[width=0.9\linewidth]{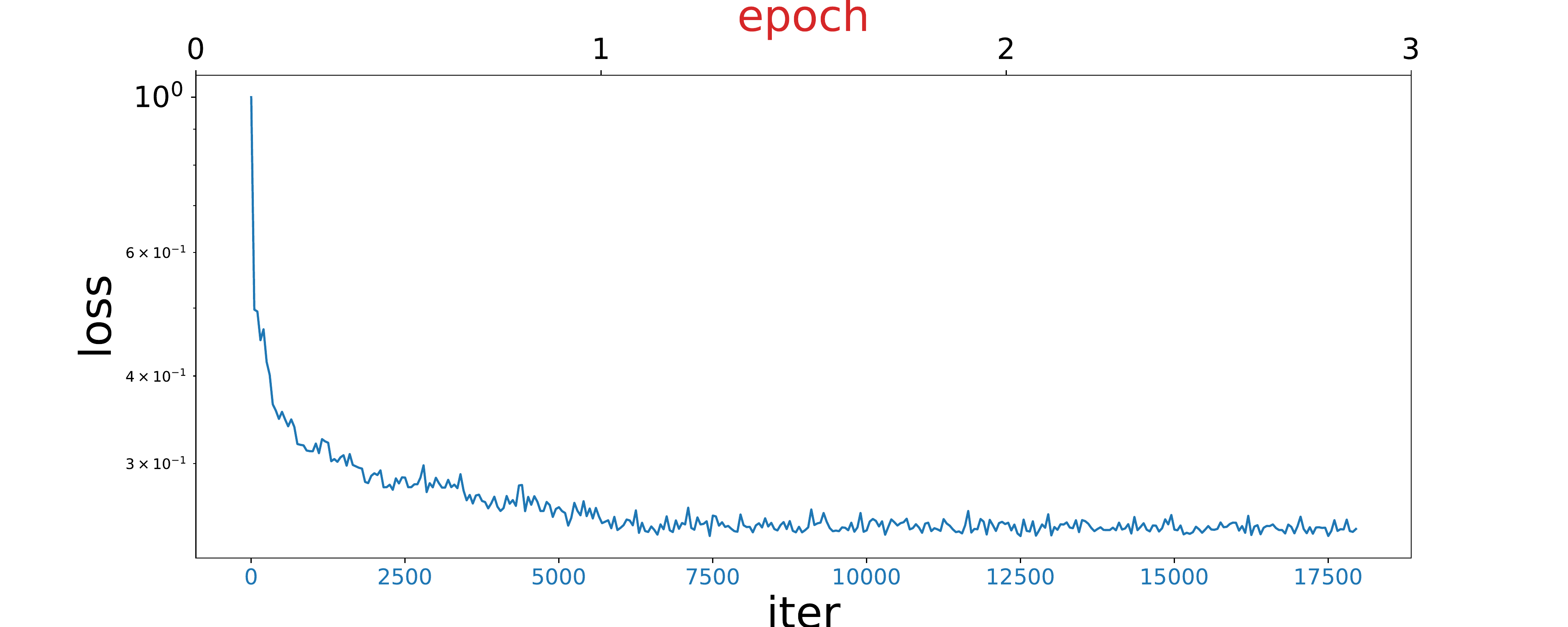}
    \end{minipage}
    }
    \caption{Loss drop in task watermark removal, dataset CelebA. These corresponds to
    four different initializations used in our experiments, namely
    Fourier, Kaiming Uniform, Kaiming Normal and Orthogonal.}
\end{figure}
\clearpage

\bibliography{main}

\begin{thebibliography}{10}

\bibitem{bhatt2021cnn}
Dulari Bhatt, Chirag Patel, Hardik Talsania, Jigar Patel, Rasmika Vaghela,
  Sharnil Pandya, Kirit Modi, and Hemant Ghayvat.
\newblock Cnn variants for computer vision: History, architecture, application,
  challenges and future scope.
\newblock {\em Electronics}, 10(20):2470, 2021.

\bibitem{lenet}
Yann LeCun, L{\'e}on Bottou, Yoshua Bengio, and Patrick Haffner.
\newblock Gradient-based learning applied to document recognition.
\newblock {\em Proceedings of the IEEE}, 86(11):2278--2324, 1998.

\bibitem{alexnet}
Alex Krizhevsky, Ilya Sutskever, and Geoffrey~E Hinton.
\newblock Imagenet classification with deep convolutional neural networks.
\newblock {\em Communications of the ACM}, 60(6):84--90, 2017.

\bibitem{He_2017_ICCV}
Kaiming He, Georgia Gkioxari, Piotr Dollar, and Ross Girshick.
\newblock Mask r-cnn.
\newblock In {\em Proceedings of the IEEE International Conference on Computer
  Vision (ICCV)}, Oct 2017.

\bibitem{ren2019}
Yinhao Ren, Zhe Zhu, Yingzhou Li, Dehan Kong, Rui Hou, Lars~J Grimm, Jeffery~R
  Marks, and Joseph~Y Lo.
\newblock Mask embedding for realistic high-resolution medical image synthesis.
\newblock In {\em International Conference on Medical Image Computing and
  Computer-Assisted Intervention}, pages 422--430. Springer, 2019.

\bibitem{unet}
Olaf Ronneberger, Philipp Fischer, and Thomas Brox.
\newblock U-net: Convolutional networks for biomedical image segmentation.
\newblock In {\em International Conference on Medical image computing and
  computer-assisted intervention}, pages 234--241. Springer, 2015.

\bibitem{fractionalFT_SIGNAL}
Rajiv Saxena and Kulbir Singh.
\newblock Fractional fourier transform: A novel tool for signal processing.
\newblock {\em Journal of the Indian Institute of Science}, 85(1):11, 2005.

\bibitem{FTsignal_processing}
Lawrence~R Rabiner and Bernard Gold.
\newblock Theory and application of digital signal processing.
\newblock {\em Englewood Cliffs: Prentice-Hall}, 1975.

\bibitem{ghani2020review}
Hadhrami~Ab Ghani, Mohamad Razwan~Abdul Malek, Muhammad Fadzli~Kamarul Azmi,
  Muhammad~Jefri Muril, and Azizul Azizan.
\newblock A review on sparse fast fourier transform applications in image
  processing.
\newblock {\em International Journal of Electrical \& Computer Engineering
  (2088-8708)}, 10(2), 2020.

\bibitem{uzun2005fpga}
Isa~Servan Uzun, Abbes Amira, and Ahmed Bouridane.
\newblock Fpga implementations of fast fourier transforms for real-time signal
  and image processing.
\newblock {\em IEE Proceedings-Vision, Image and Signal Processing},
  152(3):283--296, 2005.

\bibitem{xu2013unnatural}
Li~Xu, Shicheng Zheng, and Jiaya Jia.
\newblock Unnatural l0 sparse representation for natural image deblurring.
\newblock In {\em Proceedings of the IEEE conference on computer vision and
  pattern recognition}, pages 1107--1114, 2013.

\bibitem{cooley1967historical}
James~W Cooley, Peter~AW Lewis, and Peter~D Welch.
\newblock Historical notes on the fast fourier transform.
\newblock {\em Proceedings of the IEEE}, 55(10):1675--1677, 1967.

\bibitem{cooley1969fast}
James~W Cooley, Peter~AW Lewis, and Peter~D Welch.
\newblock The fast fourier transform and its applications.
\newblock {\em IEEE Transactions on Education}, 12(1):27--34, 1969.

\bibitem{candes2009fast}
Emmanuel Candes, Laurent Demanet, and Lexing Ying.
\newblock A fast butterfly algorithm for the computation of fourier integral
  operators.
\newblock {\em Multiscale Modeling \& Simulation}, 7(4):1727--1750, 2009.

\bibitem{li2017interpolative}
Yingzhou Li and Haizhao Yang.
\newblock Interpolative butterfly factorization.
\newblock {\em SIAM Journal on Scientific Computing}, 39(2):A503--A531, 2017.

\bibitem{li2015}
Yingzhou Li, Haizhao Yang, Eileen~R. Martin, Kenneth~L. Ho, and Lexing Ying.
\newblock Butterfly factorization.
\newblock {\em Multiscale Modeling \& Simulation}, 13(2):714--732, 2015.

\bibitem{li2015multiscale}
Yingzhou Li, Haizhao Yang, and Lexing Ying.
\newblock A multiscale butterfly algorithm for multidimensional fourier
  integral operators.
\newblock {\em Multiscale Modeling \& Simulation}, 13(2):614--631, 2015.

\bibitem{li2018}
Yingzhou Li, Haizhao Yang, and Lexing Ying.
\newblock Multidimensional butterfly factorization.
\newblock {\em Applied and Computational Harmonic Analysis}, 44(3):737--758,
  2018.

\bibitem{ying2009sparse}
Lexing Ying.
\newblock Sparse fourier transform via butterfly algorithm.
\newblock {\em SIAM Journal on Scientific Computing}, 31(3):1678--1694, 2009.

\bibitem{cnnzipcode}
John Denker, W~Gardner, Hans Graf, Donnie Henderson, R~Howard, W~Hubbard,
  Lawrence~D Jackel, Henry Baird, and Isabelle Guyon.
\newblock Neural network recognizer for hand-written zip code digits.
\newblock {\em Advances in neural information processing systems}, 1, 1988.

\bibitem{cnnTrans}
Yutong Xie, Jianpeng Zhang, Chunhua Shen, and Yong Xia.
\newblock Cotr: Efficiently bridging cnn and transformer for 3d medical image
  segmentation.
\newblock In {\em International conference on medical image computing and
  computer-assisted intervention}, pages 171--180. Springer, 2021.

\bibitem{zhang2017learning}
Kai Zhang, Wangmeng Zuo, Shuhang Gu, and Lei Zhang.
\newblock Learning deep cnn denoiser prior for image restoration.
\newblock In {\em Proceedings of the IEEE conference on computer vision and
  pattern recognition}, pages 3929--3938, 2017.

\bibitem{cnnsucess}
Yanting Pei, Yaping Huang, Qi~Zou, Hao Zang, Xingyuan Zhang, and Song Wang.
\newblock Effects of image degradations to cnn-based image classification.
\newblock {\em arXiv preprint arXiv:1810.05552}, 2018.

\bibitem{DingxuanZhou}
Felipe Cucker and Ding~Xuan Zhou.
\newblock {\em Learning theory: an approximation theory viewpoint}, volume~24.
\newblock Cambridge University Press, 2007.

\bibitem{universalapproximation}
Allan Pinkus.
\newblock Approximation theory of the mlp model in neural networks.
\newblock {\em Acta numerica}, 8:143--195, 1999.

\bibitem{li2018butterfly}
Yingzhou Li, Xiuyuan Cheng, and Jianfeng Lu.
\newblock Butterfly-net: Optimal function representation based on convolutional
  neural networks.
\newblock {\em arXiv preprint arXiv:1805.07451}, 2018.

\bibitem{fcnn}
Harry Pratt, Bryan Williams, Frans Coenen, and Yalin Zheng.
\newblock Fcnn: Fourier convolutional neural networks.
\newblock In {\em Joint European Conference on Machine Learning and Knowledge
  Discovery in Databases}, pages 786--798. Springer, 2017.

\bibitem{fourierOperator}
Zongyi Li, Nikola Kovachki, Kamyar Azizzadenesheli, Burigede Liu, Kaushik
  Bhattacharya, Andrew Stuart, and Anima Anandkumar.
\newblock Fourier neural operator for parametric partial differential
  equations.
\newblock {\em arXiv preprint arXiv:2010.08895}, 2020.

\bibitem{Pbutterfly}
Beidi Chen, Tri Dao, Kaizhao Liang, Jiaming Yang, Zhao Song, Atri Rudra, and
  Christopher Re.
\newblock Pixelated butterfly: Simple and efficient sparse training for neural
  network models.
\newblock {\em arXiv preprint arXiv:2112.00029}, 2021.

\bibitem{learningBF}
Tri Dao, Albert Gu, Matthew Eichhorn, Atri Rudra, and Christopher R{\'e}.
\newblock Learning fast algorithms for linear transforms using butterfly
  factorizations.
\newblock In {\em International conference on machine learning}, pages
  1517--1527. PMLR, 2019.

\bibitem{xu2020butterfly}
Zhongshu Xu, Yingzhou Li, and Xiuyuan Cheng.
\newblock Butterfly-net2: Simplified butterfly-net and fourier transform
  initialization.
\newblock In {\em Mathematical and Scientific Machine Learning}, pages
  431--450. PMLR, 2020.

\bibitem{gilton2019neumann}
Davis Gilton, Greg Ongie, and Rebecca Willett.
\newblock Neumann networks for linear inverse problems in imaging.
\newblock {\em IEEE Transactions on Computational Imaging}, 6:328--343, 2019.

\end{thebibliography}

\end{document}